%% file: main.tex
\documentclass[11pt]{article}
\usepackage{arxiv}
\usepackage{amsmath,amssymb,mathtools,amsthm}
\usepackage{booktabs,graphicx,multirow,subcaption}
\usepackage{placeins}
\usepackage{algorithm,algorithmic}
\usepackage[numbers,sort&compress]{natbib}
\newcommand{\yrcite}[1]{\citep{#1}}
\renewcommand{\cite}[1]{\citep{#1}}
\newcommand{\BibStyle}{unsrtnat}

\theoremstyle{plain}
\newtheorem{proposition}{Proposition}
\theoremstyle{definition}

\newcommand{\AppendixLayout}{}
\newcommand{\ReproCodeNote}{Full source code, experiment scripts, and verification logs are publicly released at \url{https://github.com/synsis/weight_space_programming}.}

\shorttitle{Do We Really Need an LLM for Weight-Space Programming?}
\title{Do We Really Need an LLM for Weight-Space Programming?\\A Unified Forward-Only Optimization Framework and Its Capability Boundaries}
\author{
  \textbf{Lin Yao}\\[3pt]
  School of Computer Science, Shanghai Jiao Tong University, Shanghai, China\\
  Zhongguancun Academy, Beijing, China\\[2pt]
  \texttt{lin.yao@sjtu.edu.cn}
}

\begin{document}

\maketitle

\input{body}

\end{document}

%% file: body.tex
\begin{abstract}
\emph{Weight-Space Programming} (WSP) edits weights without backpropagation via LLM-authored scripts, causing multi-turn retries and zero-score failures. How much of WSP requires an LLM, and how much reduces to linear algebra? We formalize \emph{Forward Weight-Space Programming (FWSP)}, a deterministic compilation layer mapping forward statistics into permanent, hook-free updates via four closed-form operators and optional retention-orthogonal subspace discovery. FWSP runs standalone on forward-observable edits and serves as a compiler backend for LLM planners. Under unmodified evaluators on November 2026 benchmarks, optimizer-LLM-free FWSP outperforms published frontier agents on matched subsets: $\mathbf{81.07}$ overall and $\mathbf{86.49}$ on \texttt{W4/W5} ($0/14$ zero-score failures, $5/25$ passes) on \texttt{OpenWSP} v8 public \texttt{Edit} (vs.\ \texttt{GPT-6.1 Sol} $75.73$, trailing on \texttt{W2}), and $\mathbf{0.373}$ on \texttt{WeightBench} \texttt{TR02-LOOP} (vs.\ \texttt{Claude Opus 5.5} $0.284$). Across those $15$ \texttt{OpenWSP} tasks ($2$ agents $\times$ $3$ arms, $90$ episodes), FWSP's compiler beats scratch scripting and standard Ridge/SVD tools on \texttt{Kimi-K3.1} ($10.33 \to 29.88 \to \mathbf{65.66}$) and \texttt{DeepSeek-V4.1-Flash} ($0.37 \to 0.74 \to \mathbf{39.56}$). The same operators scale across seven modalities and $33$ models, while geometric bounds and forward predictors ($87.5\%$ cross-architecture accuracy for solver selection and protection gating) separate analytic compilation from gradient handoff.
\end{abstract}

\begin{figure*}[t]
    \centering
    \includegraphics[width=0.94\textwidth]{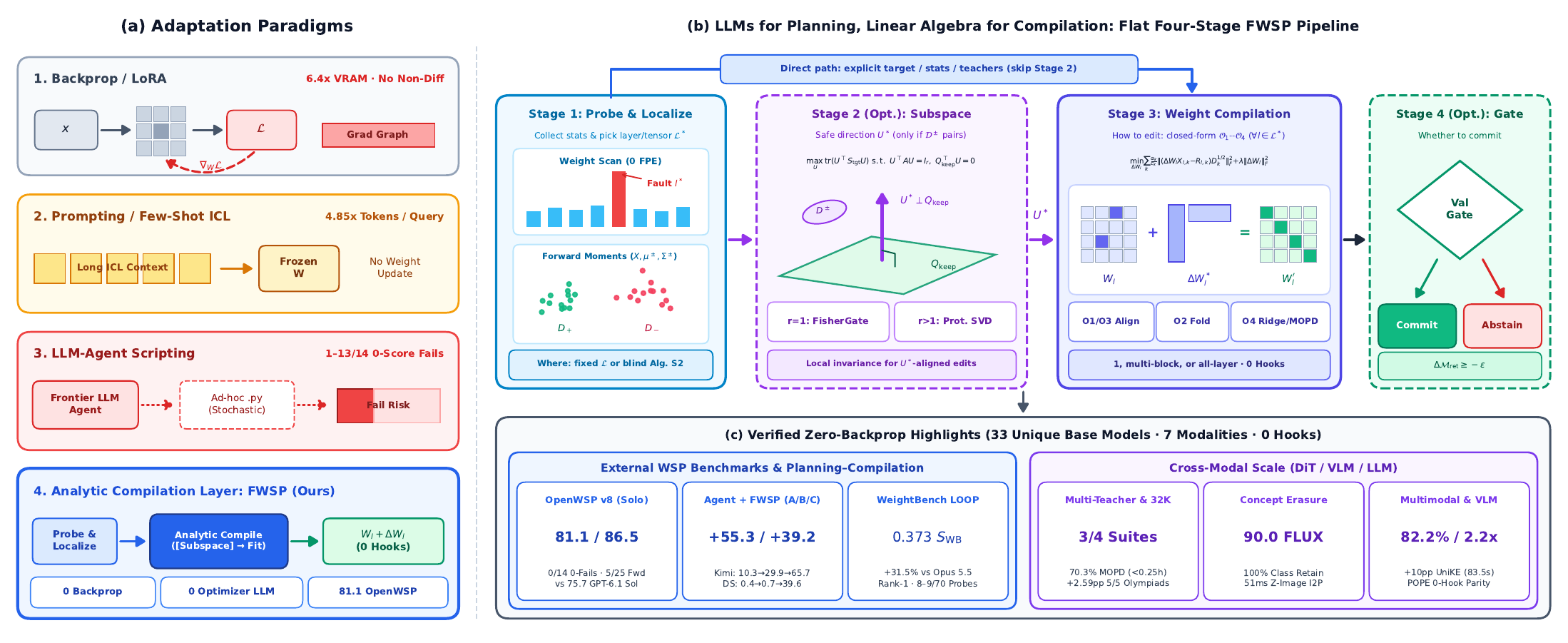}
    \caption{Overview of FWSP: (a) adaptation regimes comparing backpropagation, prompting, LLM-agent scripting, and FWSP's deterministic compilation layer; (b) the four-stage forward pipeline (localization, optional protected subspace discovery, weight compilation, and verification); and (c) external \texttt{OpenWSP} v8 and \texttt{WeightBench} \texttt{TR02-LOOP} highlights.}
    \label{fig:overview}
\end{figure*}

\section{Introduction}
Adapting a trained neural network traditionally splits into two regimes: gradient-based fine-tuning, which requires backpropagation graphs, activation memory, and differentiable losses, and inference-time prompting or external harnesses~\cite{brown2020gpt3}, which leave weights unchanged and add per-query overhead. \emph{Weight-Space Programming} (WSP) introduces a third path. Takehi et al.~\yrcite{takehi2026wsp} showed that a frontier LLM can inspect and edit weights without gradients, Fan et al.~\yrcite{fan2026selfrewriting} demonstrated \emph{self-rewriting} by a 27B model in a Python sandbox, and two November 2026 benchmarks (\texttt{OpenWSP}~\cite{moreira2026openwsp} and \texttt{WeightBench}~\cite{marasini2026weightbench}) standardized gradient-free weight editing against frontier LLM agents.

Existing WSP workflows conflate two distinct functions: high-level task planning and low-level numerical weight compilation. Because current agents generate ad-hoc editing scripts from scratch, they frequently spend multiple turns debugging code or trigger collateral collapse and zero-score failures~\cite{fan2026selfrewriting,moreira2026openwsp,marasini2026weightbench}. At the same time, WSP has lacked compute-matched comparisons against zeroth-order optimization (\texttt{MeZO}~\cite{malladi2023mezo}, ES~\cite{salimans2017es}), test-time adaptation (\texttt{BN-Adapt}~\cite{schneider2020bnadapt}, \texttt{Tent}~\cite{wang2021tent}), in-context learning (ICL), and gradient updates (\texttt{LoRA}~\cite{hu2022lora}, \texttt{Full-FT}), leaving its exact capability boundaries unmapped.

We separate planning from compilation by formalizing \emph{Forward Weight-Space Programming (FWSP)}, a deterministic, architecture-agnostic compilation layer built from forward activation statistics and closed-form linear algebra (Figure~\ref{fig:overview}). On tasks with forward-observable diagnostics or contrastive pairs, FWSP compiles permanent weight updates directly with zero optimizer LLM tokens. On open-ended semantic or symbolic tasks (such as physics policy synthesis in \texttt{WeightBench} \texttt{CTRL}, circuit construction in \texttt{WRITE}, or example-free skill injection in \texttt{OpenWSP Skill}), LLM agents retain a clear role at the planning layer and can invoke FWSP operators as reliable compilation primitives.

Our empirical study addresses five research questions:
\begin{itemize}
    \item \emph{RQ1 (External WSP Generalization \& Scale):} How does frozen LLM-free FWSP compare against published 2026 frontier LLM agents on \texttt{OpenWSP}~\cite{moreira2026openwsp} and \texttt{WeightBench}~\cite{marasini2026weightbench}, where do LLM planners remain complementary, and does FWSP scale across CNNs~\cite{he2016resnet}, ViTs~\cite{dosovitskiy2021vit,oquab2024dinov2}, 12B DiTs, RL, PDE surrogates~\cite{li2021fno}, and 8B autoregressive and diffusion LMs~\cite{yang2025qwen3,nie2025llada}?
    \item \emph{RQ2 (Compute \& Memory Frontier):} Under matched Forward-Pass Equivalents (FPE), peak VRAM, and wall-clock time, where does FWSP beat zeroth-order optimization, continual SGD, and 4-shot ICL, and where is the crossover budget $B^*$?
    \item \emph{RQ3 (Protected Reasoning Control):} Can deterministic forward-only edits improve held-out reasoning on \texttt{GSM8K}, \texttt{MATH-500}, \texttt{HumanEval}, \texttt{MBPP}, \texttt{AIME 2024--2026}, and \texttt{LiveCodeBench}, transfer zero-shot across tasks and serving engines, and abstain on tokenization boundaries?
    \item \emph{RQ4 (Geometric Capability Boundaries):} Where do forward-only subspace projection, moment calibration, and ridge compilation hit entanglement or representation-span walls relative to \texttt{MEMIT/ROME} and backpropagation?
    \item \emph{RQ5 (Scope Prediction \& Hybrid Handoff):} Can forward statistics predict solver choice, parameter scope, and protection needs prior to editing, and how should FWSP hand off to localized \texttt{LoRA}?
\end{itemize}

\noindent\textbf{Contributions.}
\begin{enumerate}
    \item \emph{Unified Analytic Weight Compilation (FWSP):} A four-stage, gradient-free, hook-free framework combining optional retention-orthogonal subspace discovery with four closed-form weight-compilation operators across arbitrary target layer sets.
    \item \emph{External Benchmark Evidence and Planning--Compilation Validation:} Optimizer-LLM-free FWSP outperforms published 2026 frontier agents on exact matched subsets of \texttt{OpenWSP} public \texttt{Edit} and \texttt{WeightBench} \texttt{TR02-LOOP}; a controlled $3$-arm evaluation across all $15$ \texttt{OpenWSP} tasks ($2$ live agents $\times$ $3$ conditions, $90$ episodes) further separates generic tool availability from FWSP's full analytic compiler design (retention-covariance protection, non-trigger contrast, and analytical margin calibration), lifting \texttt{Kimi-K3.1} and \texttt{DeepSeek-V4.1-Flash} by $+35.8$ to $+38.8$ points over standard Ridge/SVD tools ($+39.2$ to $+55.3$ over scripting from scratch), while the same operators scale across seven modalities and $33$ unique base models.
    \item \emph{Capability Boundaries and Forward Optimization Routing:} Theoretical bounds on feature entanglement and frozen representation span, paired with grid-verified forward predictors for Analytic-vs.-Gradient solver selection and independent retention-protection gating.
\end{enumerate}

\section{Related Work}
\label{sec:related}
\paragraph{Weight-Space Programming and Zeroth-Order Optimization.}
Recent studies treat neural weights as editable programs without gradients, including frontier WSP~\cite{takehi2026wsp}, 27B self-rewriting~\cite{fan2026selfrewriting}, \texttt{WeightCode}~\cite{li2026weightcode}, architecture-splicing agents~\cite{doubleai2026redesign}, lookup-based WSP~\cite{ok2026wsp}, reading-side audits~\cite{ahmed2026wsp}, and two November 2026 benchmarks (\texttt{OpenWSP}~\cite{moreira2026openwsp} and \texttt{WeightBench}~\cite{marasini2026weightbench}). All rely on LLM agents to write task-specific editing scripts. Classical backprop-free learners instead use isotropic zeroth-order perturbations (\texttt{SPSA}~\cite{spall1992spsa}, \texttt{MeZO}~\cite{malladi2023mezo}, ES~\cite{salimans2017es}, forward gradients~\cite{baydin2022forwardgrad}) with $O(d)$ directional variance, or iterative local surrogate-gradient rules~\cite{nokland2016dfa,lillicrap2016fa,hinton2022forward}. FWSP replaces both ad-hoc script generation and random perturbations with closed-form spectral and covariance updates (Table~\ref{tab:llm_vs_fwsp}), running standalone on forward-observable tasks or acting as deterministic primitives for LLM planners.

\paragraph{Model Editing, Concept Erasure, and Multi-Teacher Distillation.}
Locate-and-edit methods (\texttt{ROME}~\cite{meng2022rome}, \texttt{MEMIT}~\cite{meng2023memit}, \texttt{MEND}~\cite{mitchell2022mend}), representation engineering~\cite{zou2023repe,arditi2024refusal,ilharco2023taskarithmetic}, and closed-form diffusion erasers (\texttt{UCE}~\cite{gandikota2024uce}, \texttt{MACE}~\cite{lu2024mace}, \texttt{RECE}~\cite{gong2024rece}, \texttt{SPEED}~\cite{speed2026nullspace}) edit weights or directions for single tasks. In model consolidation, \texttt{RegMean}~\cite{jin2023regmean} merges fine-tuned models via input Gram matrices, whereas multi-teacher on-policy distillation (\texttt{MOPD}~\cite{ma2026mopd}, \texttt{Open-MOPD}~\cite{gao2026openmopd}, \texttt{IM-MOPD}~\cite{kim2026immopd}, \texttt{P$^2$-MOPD}~\cite{wang2026p2mopd}) relies on multi-domain RL rollouts. FWSP unifies retention-orthogonal subspace excision and curated, base-centered layer-wise ridge synthesis under one gradient-free, hook-free framework.

\paragraph{Reasoning Control, Test-Time Adaptation, and Continual Learning.}
Existing reasoning-length controls rely on decoding wrappers (\texttt{s1}~\cite{muennighoff2025s1}, \texttt{Budget Guidance}~\cite{li2025budgetguidance}), runtime activation hooks (\texttt{ActAdd}~\cite{turner2023actadd}, \texttt{CAA}~\cite{rimsky2024caa}, \texttt{SEAL}~\cite{chen2025seal}), or unwhitened attention-head edits (\texttt{ThinkEdit}~\cite{sun2025thinkedit}). Across distribution shifts, continual learning, and security, prior methods span forward \texttt{BN-Adapt}~\cite{schneider2020bnadapt}, backprop TTA (\texttt{Tent}~\cite{wang2021tent}, \texttt{EATA}~\cite{niu2022eata}, \texttt{SAR}~\cite{niu2023sar}, \texttt{DeYO}~\cite{lee2024deyo}), analytic continual readouts (\texttt{ACIL}~\cite{zhuang2022acil}, \texttt{DFR}~\cite{kirichenko2023dfr,sagawa2020groupdro}), backdoor defenses (\texttt{Fine-Pruning}~\cite{liu2018finepruning}, \texttt{Neural Cleanse}~\cite{wang2019neuralcleanse,gu2019badnets}), and bit-flip reliability~\cite{rakin2019bitflip}. FWSP compiles these regimes permanently into static checkpoints and characterizes their exact geometric boundaries (extended discussion in Appendix~\ref{app:method_details}).

\section{Method: Forward Weight-Space Programming (FWSP)}
\label{sec:method}

\subsection{Problem Formulation, Diagnostic Data, and Flat Four-Stage Pipeline}
\label{sec:pipeline}
Given a trained $L$-layer network $f_\theta$ with layer weights $\{W_l\}_{l=1}^L$ and a small diagnostic batch $\mathcal{D}_{\text{diag}}$ ($0$--$64$ examples disjoint from test splits; Appendix Table~\ref{tab:solver_taxonomy}), FWSP computes permanent in-place weight updates $W_l' = W_l + \Delta W_l^*$ using forward passes (\texttt{no\_grad}) and closed-form linear algebra, with zero backpropagation and zero runtime hooks. Diagnostic inputs scale with the task: zero data for static weight faults, unlabeled shifted or contrastive batches for TTA, concept erasure, and reasoning control, and few-shot labeled or teacher-rollout batches for weighted ridge compilation.

Every task follows one four-stage pipeline separating \emph{where} to edit, \emph{what} safe direction to use, \emph{how} to update weights, and \emph{whether} to commit (Table~\ref{tab:pipeline_stages}, Figure~\ref{fig:overview}b; Algorithms~\ref{alg:fwsp}--\ref{alg:blind}). \textbf{Stage~1 (Probe \& Localize)} records per-layer summary statistics $\mathcal{S}_l \triangleq ((\rho_l,\eta_l), (\mu_l,\Sigma_l,X_l X_l^\top))$, where $(\rho_l, \eta_l)$ are static weight anomaly ratios (per-layer maximum-to-median entry magnitude $\rho_l = \max|W_l|/\mathrm{med}|W_l|$ and maximum-to-median row-norm ratio $\eta_l = \max_i\|W_{l,i:}\|_2/\mathrm{med}_j\|W_{l,j:}\|_2$, detecting bit-flips or row spikes without data) and $(\mu_l, \Sigma_l, X_l X_l^\top)$ are the mean, covariance, and uncentered second moment of layer-$l$ forward activations $X_l$. Stage~1 then selects the target layer set $\mathcal{L}^*$ and compilation operator $\mathcal{O}^* \in \{\mathcal{O}_1, \dots, \mathcal{O}_4\}$ (fixed by the regime or localized automatically in blind repair via Algorithm~\ref{alg:blind}). Tasks with explicit targets (normal weight statistics, clean moments, labels, neutral anchors, or teacher weights) proceed directly to \textbf{Stage~3 (Weight Compilation)}. Tasks given only unlabeled contrastive pairs $(\mathcal{D}^+, \mathcal{D}^-)$ first run \textbf{Stage~2 (Protected Subspace Discovery)} to extract a rank-$r$ basis $U_l^* \in \mathbb{R}^{D \times r}$ whitened under metric $A$ and orthogonal to a protected reference basis $Q_{\text{keep}}$ (Section~\ref{sec:pfg_method}). Finally, optional \textbf{Stage~4 (Verify \& Rollback)} evaluates the edited candidate $\theta'$ on a disjoint validation slice $\mathcal{D}_{\text{val}}$ and commits $\theta_{\text{out}} = \theta'$ only if the mean paired gain $\bar{d} \ge \tau_{\text{gain}}$, paired validation statistic $z(\mathcal{D}_{\text{val}}) \ge z_{\min}$, and retention change $\Delta\mathcal{M}_{\text{ret}} \ge -\epsilon$ all pass; otherwise it rolls back to $\theta$.

\begin{table*}[t]
\centering
\caption{Flat four-stage FWSP pipeline separating \emph{where} to edit (Stage~1: layer/tensor set $\mathcal{L}^*$), \emph{what} safe direction (Stage~2, optional), \emph{how} to compile weights (Stage~3), and \emph{whether} to commit (Stage~4, optional).}
\label{tab:pipeline_stages}
\footnotesize
\resizebox{\textwidth}{!}{%
\begin{tabular}{lll}
\toprule
\textbf{Stage (Role)} & \textbf{Formal Output / Update} & \textbf{Responsibility \& Key Symbols} \\
\midrule
\begin{tabular}[c]{@{}l@{}}\textbf{1. Probe \& Localize}\\ (\emph{Core; blind opt.})\end{tabular} & $\mathcal{S}_l \triangleq \big((\rho_l,\eta_l),\, (\mu_l,\Sigma_l,X_l X_l^\top)\big),\; (\mathcal{L}^*,\mathcal{O}^*)$ & \begin{tabular}[c]{@{}l@{}}\textbf{Where to edit:} one-pass \texttt{no\_grad} stats + target layer/tensor\\ set $\mathcal{L}^*$ (known scope, or blind localization via Alg.~\ref{alg:blind})\end{tabular} \\
\begin{tabular}[c]{@{}l@{}}\textbf{2. Subspace Discovery}\\ (\emph{Optional; $\mathcal{D}^\pm$ only})\end{tabular} & $U_l^* \in \mathbb{R}^{D \times r}$ s.t.\ $U_l^{*\top} A U_l^* {=} I_r,\; Q_{\text{keep}}^\top U_l^* {=} 0$ & \begin{tabular}[c]{@{}l@{}}\textbf{Safe direction:} skipped if explicit targets/stats/teachers exist;\\ else extracts $U_l^*$ (Sec.~\ref{sec:pfg_method}, Table~\ref{tab:solver2_ops}) for Stage~3\end{tabular} \\
\begin{tabular}[c]{@{}l@{}}\textbf{3. Weight Compilation}\\ (\emph{Core; $\forall l \in \mathcal{L}^*$})\end{tabular} & $W_l' = W_l + \Delta W_l^*(W_l,\mathcal{S}_l,U_l^*;\mathcal{O}^*)$ & \begin{tabular}[c]{@{}l@{}}\textbf{How to update:} closed-form $\mathcal{O}_1$--$\mathcal{O}_4$ (Sec.~\ref{sec:solver1}, Table~\ref{tab:solver1_ops})\\ folding explicit targets or $U_l^*$ in-place into $W_l'$ (0 hooks)\end{tabular} \\
\begin{tabular}[c]{@{}l@{}}\textbf{4. Verify \& Rollback}\\ (\emph{Optional guard})\end{tabular} & $\theta_{\text{out}} = \theta'$ iff $\bar{d} \ge \tau_{\text{gain}} \wedge z(\mathcal{D}_{\text{val}}) \ge z_{\min} \wedge \Delta\mathcal{M}_{\text{ret}} \ge -\epsilon$ (else $\theta$) & \begin{tabular}[c]{@{}l@{}}\textbf{Whether to commit:} held-out empirical validation gate on $\mathcal{D}_{\text{val}}$\\ (never used to solve $\theta'$); rolls back to $\theta$ on failure, else commits\end{tabular} \\
\bottomrule
\end{tabular}}
\end{table*}

\subsection{Stage 2 (Optional): Protected Subspace Discovery}
\label{sec:pfg_method}
When no explicit target output $Y^*$ or teacher weight is given and only unlabeled contrastive activation pairs $(\mathcal{D}^+, \mathcal{D}^-)$ are observed (for instance, hidden states at \texttt{</think>} versus ongoing reasoning tokens in an LLM, faithful versus hallucinated token states in a VLM, or safe versus unsafe prompt cross-attention activations in a diffusion DiT), Stage~2 extracts a rank-$r$ edit basis $U^* \in \mathbb{R}^{D \times r}$ before modifying any weight. Let $\mu^+, \mu^- \in \mathbb{R}^D$ denote the activation means on $(\mathcal{D}^+, \mathcal{D}^-)$ with mean shift $\Delta\mu \triangleq \mu^- - \mu^+$, and let $\Delta X$ denote the matrix of paired contrastive activation differences. Stage~2 constructs a target scatter matrix $S_{\text{target}}$ ($\Delta\mu\Delta\mu^\top$ for rank-$1$ gating or $\Delta X^\top \Delta X$ for rank-$r$ excision), a positive-definite whitening metric $A \succ 0$ (either pooled within-class covariance $\Sigma_w + \lambda I$ or protected feature covariance $\Sigma_{\text{keep}} + \lambda I$ with ridge $\lambda > 0$), and an orthonormal protected basis $Q_{\text{keep}}$ spanning the top singular vectors of clean reference activations (e.g., WikiText passages, MS-COCO prompts, or clean ImageNet/encyclopedic probes). It then solves (for any feasible rank $r \le D - \mathrm{rank}(Q_{\text{keep}})$):
\begin{equation}
\begin{aligned}
    U^* &= \arg\max_{U \in \mathbb{R}^{D \times r}} \; \mathrm{tr}\!\left(U^\top S_{\text{target}} U\right) \\
    &\text{s.t.} \quad U^\top A U = I_r, \quad Q_{\text{keep}}^\top U = 0.
\end{aligned}
\label{eq:pfg_unified}
\end{equation}
Specializing $(S_{\text{target}}, A, r)$ yields the two canonical extractors in Table~\ref{tab:solver2_ops} (derivations in Appendix~\ref{app:method_details}): rank-$1$ \texttt{Protected FisherGate} produces a unit direction $\hat{d}_{\text{PFG}}^*$ folded with gain $\gamma$ into a single readout row $e_{\text{tgt}}$ (e.g., the \texttt{</think>} or \texttt{Yes}/\texttt{No} logit row), while rank-$r$ \texttt{Protected Excision} takes the top-$r$ generalized eigenvectors $U_A^*$ and orthonormalizes them via $\bar{U}_A = \mathrm{QR}(U_A^*)$ for damped projection with strength $\alpha \in (0,1]$. Because $Q_{\text{keep}}^\top U^* = 0$, every protected activation $h \in \mathrm{span}(Q_{\text{keep}})$ satisfies exact local invariance ($\Delta z(h){=}0$ or $\Delta h{=}0$) under Stage~3 edits aligned with $U^*$ (projective $\mathcal{O}_2$ excision and single-row additive $\mathcal{O}_2$ gating; unconstrained Stage~3 updates instead enforce retention via explicit keep-data regularization). Stage~2 passes $U^*$ to Stage~3 for permanent weight compilation and is skipped whenever explicit targets or teachers are available.

\begin{table*}[t]
\centering
\caption{Two canonical Stage~2 protected subspace extractors specializing Eq.~\eqref{eq:pfg_unified} ($\Delta\mu{=}\mu^-{-}\mu^+$; $U_A^*$ top-$r$ eigenvectors).}
\label{tab:solver2_ops}
\footnotesize
\resizebox{\textwidth}{!}{%
\begin{tabular}{llll}
\toprule
\textbf{Subspace Extractor} & \textbf{Eq.~\eqref{eq:pfg_unified} Specialization} & \textbf{Closed-Form Basis $U^*$ \& Stage~3 Compilation} & \textbf{Target Scenarios ($\mathcal{D}^+, \mathcal{D}^-$)} \\
\midrule
\begin{tabular}[c]{@{}l@{}}Rank-1 \texttt{Protected}\\ \texttt{FisherGate}\end{tabular} & \begin{tabular}[c]{@{}l@{}}$S_{\text{target}}{=}\Delta\mu\Delta\mu^\top$\\ $A{=}\Sigma_w{+}\lambda I,\; r{=}1$\end{tabular} & \begin{tabular}[c]{@{}l@{}}$d_{\text{PFG}}^* = A^{-1}\Delta\mu - A^{-1} Q_{\text{keep}}(Q_{\text{keep}}^\top A^{-1} Q_{\text{keep}})^{-1} Q_{\text{keep}}^\top A^{-1}\Delta\mu$\\ $\hat{d}_{\text{PFG}}^* = d_{\text{PFG}}^*/\|d_{\text{PFG}}^*\|_2$ $\;\to\;$ Stage~3 $\mathcal{O}_2$ additive row fold ($W'{=}W{+}\gamma e_{\text{tgt}}\hat{d}_{\text{PFG}}^{*\top}$)\end{tabular} & \begin{tabular}[c]{@{}l@{}}CoT length (\texttt{</think>}),\\ VLM hallucination control\end{tabular} \\
\begin{tabular}[c]{@{}l@{}}Rank-$r$ Protected\\ Excision\end{tabular} & \begin{tabular}[c]{@{}l@{}}$S_{\text{target}}{=}\Delta X^\top \Delta X$\\ $A \in \{I,\, \Sigma_{\text{keep}}{+}\lambda I\}$\end{tabular} & \begin{tabular}[c]{@{}l@{}}$\bar{U}_A = \mathrm{QR}(U_A^*)$\\ $\;\to\;$ Stage~3 $\mathcal{O}_2$ projective fold: $W_l' = (I - \alpha \bar{U}_A \bar{U}_A^\top) W_l$\end{tabular} & \begin{tabular}[c]{@{}l@{}}Diffusion/DiT erasure,\\ unlabeled backdoor removal\end{tabular} \\
\bottomrule
\end{tabular}}
\end{table*}

\subsection{Stage 3 (Core): Analytic Weight Compilation ($\mathcal{O}_1$--$\mathcal{O}_4$)}
\label{sec:solver1}
Given the layer set $\mathcal{L}^*$ from Stage~1 and either explicit targets or the Stage~2 basis $U^*$, Stage~3 updates each linear layer $l \in \mathcal{L}^*$ ($h_l = W_l x_l + b_l$, with bias-augmented input $x_l$) by minimizing regularized multi-source output mismatch over structural constraint set $\mathcal{C}$:
\begin{equation}
\begin{aligned}
    \Delta W_l^* = \arg\min_{\Delta W_l \in \mathcal{C}}
    &\sum_{k=1}^{K} \frac{\alpha_k}{c_k} \big\|(\Delta W_l X_{l,k} - R_{l,k}) D_k^{1/2}\big\|_F^2 \\[-2pt]
    &+ \lambda \|\Delta W_l\|_F^2, \qquad \forall l \in \mathcal{L}^*,
\end{aligned}
\label{eq:solver1_template}
\end{equation}
where $X_{l,k}$ stacks the forward input activations to layer $l$ for source $k$ (for example, MLP down-projection inputs from factual prompts, CLIP/T5 text-encoder token embeddings in a DiT, or teacher rollout activations in multi-teacher consolidation), $R_{l,k}$ is the corresponding desired output residual (such as target-minus-current outputs $Y_{l,k}^* - W_l X_{l,k}$ or teacher weight deltas $\Delta W_{l,k} X_{l,k}$), $D_k \succeq 0$ weights tokens or samples ($D_k{=}I$ by default), $c_k > 0$ normalizes source scale ($c_k{=}1$ for standard $\mathcal{O}_4$; $c_k{=}\mathrm{tr}(D_k)$ for trace-normalized multi-teacher synthesis), $\alpha_k > 0$ sets source priority, and $\lambda \ge 0$ is the ridge penalty.

Specializing $(X_{l,k}, R_{l,k}, D_k, c_k, \lambda, \mathcal{C})$ recovers all four closed-form operators in Table~\ref{tab:solver1_ops} (derivations in Appendix~\ref{app:method_details}): $\mathcal{O}_1$ clips anomalous entries to $\pm \tau\,\mathrm{med}|W_l|$ over the normal-magnitude box $\mathcal{C}_{\text{normal}}$; $\mathcal{O}_2$ compiles the Stage~2 basis $U_l^*$ via projective excision or single-row additive gating; $\mathcal{O}_3$ rescales shifted channels using diagonal gain $G_l = \mathrm{diag}(\sigma_{l,\text{clean}} / \sigma_{l,\text{shift}})$ and centering vector $c_l$; and $\mathcal{O}_4$ solves the weighted normal equations from cross-covariance $M_l$ and Gram matrix $\Sigma_l$. Notably, \texttt{Protected FisherGate}'s additive readout update $\Delta W = \gamma e_{\text{tgt}} \hat{d}_{\text{PFG}}^{*\top}$ is an exact $\mathcal{O}_2$ instance ($K{=}1, X{=}I, R{=}\gamma e_{\text{tgt}} \hat{d}_{\text{PFG}}^{*\top}, \lambda{=}0$) over the rank-$1$ row family $\mathcal{C}_{\text{row}}(\hat{d}_{\text{PFG}}^*) = \{c\,e_{\text{tgt}}\hat{d}_{\text{PFG}}^{*\top} : c \in \mathbb{R}\}$, unifying projective excision and additive gating without adding a fifth operator. Optional rank truncation $\mathrm{SVD}_r[\cdot]$ in $\mathcal{O}_4$ provides a post-hoc Frobenius approximation to the exact ridge solution $M_l(\Sigma_l + \lambda I)^{-1}$.

\begin{table*}[t]
\centering
\caption{Four closed-form Stage~3 weight-compilation operators specializing Eq.~\eqref{eq:solver1_template} for any target layer/tensor set $l \in \mathcal{L}^*$ ($|\mathcal{L}^*| \ge 1$).}
\label{tab:solver1_ops}
\footnotesize
\resizebox{\textwidth}{!}{%
\begin{tabular}{llll}
\toprule
\textbf{Operator} & \textbf{Eq.~\eqref{eq:solver1_template} Specialization ($l \in \mathcal{L}^*$)} & \textbf{Closed-Form Weight Update ($\forall l \in \mathcal{L}^*$)} & \textbf{Target Scenarios \& Layer Scope $|\mathcal{L}^*|$} \\
\midrule
$\mathcal{O}_1$ (Clip/Rescale) & $K{=}1,\; X_l{=}I,\; R_l{=}0,\; \mathcal{C}_{\text{normal}}$ & $W_l' = \mathrm{Clip}(W_l,\, -\tau\,\mathrm{med}|W_l|,\, \tau\,\mathrm{med}|W_l|)$ & Bit-flip / outlier repair (0 data; 1 or multi-tensor) \\
\begin{tabular}[c]{@{}l@{}}$\mathcal{O}_2$ (Subspace\\ Projection / Add.)\end{tabular} & \begin{tabular}[c]{@{}l@{}}\emph{Proj.:} $X_l{=}I,\; R_l{=}0,\; U_l^\top W_l'{=}0$\\ \emph{Add.:} $X_l{=}I,\; R_l{=}\gamma e_{\text{tgt}} u_l^\top,\; \mathcal{C}_{\text{row}}(u_l)$\end{tabular} & \begin{tabular}[c]{@{}l@{}}$W_l' = (I - \alpha U_l U_l^\top) W_l$ (+ mean bias $\mu_l^+$)\\ $W_l' = W_l + \gamma e_{\text{tgt}} u_l^\top$ ($u_l{=}\hat{d}_{\text{PFG}}^* \perp Q_{\text{keep}}$)\end{tabular} & \begin{tabular}[c]{@{}l@{}}Compiles Stage~2 $U_l^*$: backdoor / erasure (1--30 blocks)\\ or single-row readout gating (\texttt{PFG}, \texttt{</think>}, \texttt{POPE})\end{tabular} \\
$\mathcal{O}_3$ (Moment Align) & $K{=}1,\; R_l = W_l[(G_l{-}I)X_l + c_l\mathbf{1}^\top]$ & $W_l' = W_l G_l,\; G_l = \mathrm{diag}(\sigma_{l,\text{clean}} / \sigma_{l,\text{shift}})$ & Vision TTA (\texttt{L6--11}) \& CoT decay (48-layer \texttt{dt\_bias}) \\
\begin{tabular}[c]{@{}l@{}}$\mathcal{O}_4$ (Weighted\\ Ridge)\end{tabular} & \begin{tabular}[c]{@{}l@{}}$R_{l,k} = Y_{l,k}^*{-}W_l X_{l,k}$ (or $\Delta W_{l,k} X_{l,k}$),\\ $\lambda{>}0$; $c_k{=}1$ (std.) or $\mathrm{tr}(D_k)$ (MOPD)\end{tabular} & \begin{tabular}[c]{@{}l@{}}$W_l' = W_l + \mathrm{SVD}_r\!\big(M_l\,(\Sigma_l + \lambda I)^{-1}\big)^{\dagger}$\\ $M_l{=}\sum_k \frac{\alpha_k}{c_k} R_{l,k} D_k X_{l,k}^\top,\; \Sigma_l{=}\sum_k \frac{\alpha_k}{c_k} X_{l,k} D_k X_{l,k}^\top$\end{tabular} & \begin{tabular}[c]{@{}l@{}}1-layer ($K{=}1$): prune, CIL, NLP, \texttt{UniKE};\\ dual/all-layer: \texttt{FLUX} (T5+CLIP), \texttt{MOPD}\end{tabular} \\
\bottomrule
\end{tabular}}
\vspace{1pt}
\par\noindent{\footnotesize $^{\dagger}$When rank truncation $r < \mathrm{rank}(\Delta W_{\text{ridge}}^*)$ is applied, $\mathrm{SVD}_r[\cdot]$ is a post-hoc Frobenius low-rank approximation to the exact unconstrained ridge minimizer $M_l(\Sigma_l+\lambda I)^{-1}$ (exact whenever no truncation is applied). For projective $\mathcal{O}_2$, $\alpha{=}1$ is the exact constrained minimizer of Eq.~\eqref{eq:solver1_template}, while $\alpha \in (0,1)$ acts as a damped relaxation toward $W_l$; for $\mathcal{O}_3$, the Gated-DeltaNet \texttt{dt\_bias} offset is an architecture-specific scalar bias-channel extension rather than an unconstrained quadratic matrix optimum.}
\end{table*}

\paragraph{Why No Backpropagation Is Required.}
FWSP replaces end-to-end nonlinear optimization with structured local quadratic and spectral optimization problems constructed from fixed forward observations. Contrasting iterative gradient descent with analytic $\mathcal{O}_4$ compilation makes the distinction explicit:
\begin{equation*}
\begin{aligned}
    \text{GD:}\; & W_{t+1} = W_t - \eta \nabla_W \mathcal{L}_{\mathrm{task}}, \\
    \text{FWSP:}\; & W' = W + \underbrace{R X^\top}_{\text{Residual stats}}\underbrace{(X X^\top + \lambda I)^{-1}}_{\text{Feature geometry}}.
\end{aligned}
\end{equation*}
Although the closed-form stationary conditions of Eq.~\eqref{eq:pfg_unified} and Eq.~\eqref{eq:solver1_template} are derived analytically once on paper, FWSP executes zero gradient evaluations and zero backward passes at edit time. Weight updates are solved directly from forward sufficient statistics, spectral decompositions, and linear systems without building a backward computation graph or maintaining optimizer states.

\paragraph{Layer-Set Generality and Minimal-Scope Principle.}
Every operator $\mathcal{O}_1$--$\mathcal{O}_4$ applies uniformly across any layer set $\mathcal{L}^*$. Single-layer edits ($|\mathcal{L}^*|{=}1$, $K{=}1$) are simply the minimal case, while multi-layer TTA, concept erasure, and all-layer $K$-teacher consolidation ($R_{l,k}{=}\Delta W_{l,k} X_{l,k}$, $c_k{=}\mathrm{tr}(D_k)$; Eq.~\eqref{eq:fwsp_mopd}) follow from the same template. Following a minimal-scope design principle, FWSP targets the smallest causal bottleneck matching task rank and expands $|\mathcal{L}^*|$ only when distributed capacity is required.

\subsection{Theoretical Characterization of Capability Boundaries}
\label{sec:theory}
Two geometric bounds explain why subspace protection is required and where frozen-upstream single-layer linear edits hit an exact representational floor (proofs in Appendix~\ref{app:proofs}):
\begin{proposition}[Subspace Excision vs.\ Feature Entanglement Bound for $\mathcal{O}_2$]
For clean activations $h^+ = z_{\mathrm{core}} + \xi$ with mean $\mu^+ \triangleq \mathbb{E}[h^+]$ and covariance $\Sigma_+ = \mathrm{Cov}(h^+)$, fault activations $h^- = z_{\mathrm{core}} + v_{\mathrm{fault}}$ ($v_{\mathrm{fault}} \in \mathrm{span}(U), U^\top U = I_r$), orthogonal projector $P_U \triangleq I - U U^\top$, and any downstream loss $\mathcal{L}$ that is $L$-smooth on a convex activation region containing $\{h^+, \mu^+, P_U h^+ {+} U U^\top \mu^+\}$ (and the interpolating segments between them), applying $\mathcal{O}_2$ ($\alpha{=}1$, mean alignment) eliminates $v_{\mathrm{fault}}$ identically while bounding expected clean loss increase by:
\begin{equation}
\begin{aligned}
    &\mathbb{E}\!\left[\mathcal{L}(P_U h^+ {+} U U^\top \mu^+) - \mathcal{L}(h^+)\right] \\
    &\quad \le \tfrac{L}{2}\,\mathrm{tr}\!\left(U^\top \Sigma_+ U\right)
    + L\sqrt{\mathrm{tr}(\Sigma_+)\,\mathrm{tr}\!\left(U^\top \Sigma_+ U\right)},
\end{aligned}
\end{equation}
which vanishes identically whenever $U^\top \Sigma_+ U = 0$.
\end{proposition}
\begin{proposition}[Frozen-Span Barrier for Frozen-Upstream Linear Edits]
For frozen-upstream feature matrix $\tilde{\Phi} \triangleq [\Phi_\theta(X), \mathbf{1}]$ with column-space projector $\Pi_{\tilde{\Phi}} = \tilde{\Phi}(\tilde{\Phi}^\top \tilde{\Phi})^\dagger \tilde{\Phi}^\top$ and any target $Y^*$, every frozen-upstream affine edit ($\mathcal{O}_1$--$\mathcal{O}_4$) obeys the exact lower bound:
\begin{equation}
\begin{aligned}
    \min_{\Delta W, \Delta b} \| \Phi_\theta(X)(W {+} \Delta W)^\top &+ \mathbf{1}(b {+} \Delta b)^\top - Y^* \|_F^2 \\
    &= \| (I - \Pi_{\tilde{\Phi}}) Y^* \|_F^2.
\end{aligned}
\end{equation}
\end{proposition}
Proposition~1 motivates both the $Q_{\text{keep}}$-orthogonalization constraint in Stage~2 and the forward entanglement risk gate formalized in Section~\ref{sec:exp_boundaries}. Proposition~2 characterizes the exact representational limit of \emph{frozen-upstream} single-layer affine edits and motivates the frozen-span residual diagnostic in Section~\ref{sec:exp_boundaries}; exceeding this floor requires modifying upstream representations (either via multi-layer forward edits or localized gradient updates such as \texttt{LoRA}), without implying that gradient descent is uniquely mandatory whenever multi-layer forward compilation can reshape upstream features.

\section{Experiments and Results}
\label{sec:experiments}

\begin{table*}[t]
\centering
\caption{Independent November 2026 WSP benchmarks under official unmodified evaluators: (a) \texttt{OpenWSP} v8 public \texttt{Edit} ($15$ tasks)~\cite{moreira2026openwsp}; (b) \texttt{WeightBench} \texttt{TR02-LOOP} ($4$ tasks)~\cite{marasini2026weightbench}; and (c) controlled $3$-arm Planning--Compilation ablation on all $15$ \texttt{OpenWSP} v8 tasks ($90$ episodes; per-task details in Appendix Table~\ref{tab:app_external_per_task}).}
\label{tab:external_wsp_benchmarks}
\scriptsize
\setlength{\tabcolsep}{3.1pt}
\resizebox{\textwidth}{!}{%
\begin{tabular}{lcccccccccc}
\toprule
\multicolumn{11}{l}{\textbf{(a) \texttt{OpenWSP} v8 Public \texttt{Edit} Suite (15 Self-Contained Tasks, Disguised/Permuted Weights, Official \texttt{score\_write.py}, Budget $\le 25$ Fwd Passes)}} \\
\textbf{Optimizer (LLM-Free vs.\ Published Agents)} & \textbf{\texttt{W4} (5)} & \textbf{\texttt{W4c} (2)} & \textbf{\texttt{W4l} (2)} & \textbf{\texttt{W4m} (2)} & \textbf{\texttt{W5} (3)} & \textbf{\texttt{W4/W5} (14)} & \textbf{\texttt{W2} (1)} & \textbf{All-15 ($\uparrow$)} & \textbf{0-Score Fails (\texttt{W4/5}, $\downarrow$)} & \textbf{Fwd / LLM Tok} \\
\midrule
\texttt{DeepSeek-V4-Flash} (Published Agent) & $19.02$ & $0.00$ & $0.00$ & $0.00$ & $0.00$ & $6.79$ & $9.00$ & $6.94$ & $13/14$ & $\le 25$ / Multi-turn \\
\texttt{Qwen3.8-27B} (Published Open Agent) & $31.00$ & $0.00$ & $25.35$ & $9.05$ & $6.63$ & $17.41$ & $0.00$ & $16.25$ & $8/14$ & $\le 25$ / Multi-turn \\
\texttt{Claude Opus 5.5} (Published Agent) & $59.44$ & $\underline{80.35}$ & $\underline{90.85}$ & $\mathbf{96.05}$ & $12.80$ & $62.15$ & $\underline{42.30}$ & $60.83$ & $4/14$ & $\le 25$ / Multi-turn \\
\texttt{Claude Sonnet 5.5} (Published Agent) & $72.10$ & $56.50$ & $81.75$ & $76.65$ & $\mathbf{85.23}$ & $74.71$ & $0.00$ & $69.73$ & $\underline{1/14}$ & $\le 25$ / Multi-turn \\
\texttt{GPT-6.1 Sol} (Published Frontier Agent) & $\mathbf{99.72}$ & $31.70$ & $\mathbf{94.75}$ & $43.45$ & $\underline{84.20}$ & $\underline{77.93}$ & $\mathbf{44.90}$ & $\underline{75.73}$ & $2/14$ & $\le 25$ / Multi-turn \\
\textbf{FWSP (Ours, Analytic $\mathcal{O}_4$, LLM-free)} & $\underline{88.39}$ & $\mathbf{86.39}$ & $90.47$ & $\underline{90.30}$ & $78.21$ & $\mathbf{86.49}$ & $5.13$ & $\mathbf{81.07}$ & $\mathbf{0/14}$ & $\mathbf{5}$ / $\mathbf{0}$ \\
\midrule
\multicolumn{11}{l}{\textbf{(b) \texttt{WeightBench} \texttt{TR02-LOOP-001--004} (\texttt{Qwen3.5-0.8B}, Seed 0, Official Metered Sandbox \& Hash-Chained Ledger, $S_{\text{WB}} \in [0, 1]$)}} \\
\textbf{Optimizer (LLM-Free vs.\ Published Agents)} & \multicolumn{2}{c}{\textbf{\texttt{LOOP-001} (GSM8K)}} & \multicolumn{2}{c}{\textbf{\texttt{LOOP-002} (MATH 1--3)}} & \textbf{\texttt{LOOP-003} (IFEval)} & \textbf{\texttt{LOOP-004} (MATH 4--5)} & \textbf{4-Task Macro ($\uparrow$)} & \textbf{Mean $D_{\text{col}}$ ($\downarrow$)} & \textbf{Max Rank} & \textbf{Probes / Ledger} \\
\midrule
\texttt{Claude Haiku 5.5} (Published Agent) & \multicolumn{2}{c}{$0.000$} & \multicolumn{2}{c}{$0.000$} & $0.000$ & $0.000$ & $0.000$ & --- & --- & $\le 70$ / Official \\
\texttt{DeepSeek-V4-Flash} (Published Agent) & \multicolumn{2}{c}{$0.000$} & \multicolumn{2}{c}{$0.000$} & $0.000$ & $0.000$ & $0.000$ & --- & --- & $\le 70$ / Official \\
\texttt{Claude Sonnet 5.5} (Published Agent) & \multicolumn{2}{c}{$0.432$} & \multicolumn{2}{c}{$0.000$} & $0.000$ & $0.000$ & $0.108$ & --- & --- & $\le 70$ / Official \\
\texttt{Claude Opus 5.5} (Published Agent) & \multicolumn{2}{c}{$\mathbf{0.552}$} & \multicolumn{2}{c}{$\underline{0.348}$} & $\underline{0.235}$ & $0.000$ & $\underline{0.284}$ & --- & --- & $\le 70$ / Official \\
\textbf{FWSP (Ours, Rank-1 \texttt{PFG}, LLM-free)} & \multicolumn{2}{c}{$\underline{0.485}$} & \multicolumn{2}{c}{$\mathbf{0.540}$} & $\mathbf{0.439}$ & $\mathbf{0.030}$ & $\mathbf{0.373}$ & $\mathbf{0.0075}$ & $\mathbf{1}$ & $\mathbf{8\text{--}9/70}$ / \textbf{Verified} \\
\midrule
\multicolumn{11}{l}{\textbf{(c) Controlled $3$-Arm Planning--Compilation Ablation on All $15$ \texttt{OpenWSP} v8 Tasks ($2$ Live B200 Agents $\times$ $3$ Arms $= 90$ Episodes, $25$-Pass Cap, $16$ Turns)}} \\
\textbf{Live Agent + Condition} & \textbf{All-15 ($\uparrow$)} & \textbf{\texttt{W4/W5} (14)} & \textbf{\texttt{W2} (1)} & \textbf{0-Score (\texttt{W4/5})} & \textbf{Paired $\Delta$ vs.\ \texttt{Scratch} [95\% CI]} & \textbf{Paired $\Delta$ vs.\ \texttt{Std} [95\% CI]} & \textbf{W:T:L (vs.\ \texttt{Std})} & \textbf{Mean Fwd} & \textbf{Wall Time ($\downarrow$)} & \textbf{Mean LLM Tok} \\
\midrule
\texttt{DeepSeek-V4.1-Flash} (\texttt{Scratch}) & $0.37$ & $0.40$ & $0.00$ & $12/14$ & Ref. & --- & --- & $1.5/25$ & $297.4\text{s}$ & $107.5\text{K}$ \\
\texttt{DeepSeek-V4.1-Flash} (\texttt{+Standard Tools}) & $0.74$ & $0.80$ & $0.00$ & $12/14$ & $+0.37\;[-0.6, +1.7]$ & Ref. & --- & $3.7/25$ & $216.4\text{s}$ & $123.2\text{K}$ \\
\textbf{\texttt{DeepSeek-V4.1-Flash} (\texttt{+FWSP Tools})} & $\mathbf{39.56}$ & $\mathbf{42.38}$ & $0.00$ & $\mathbf{6/14}$ & $\mathbf{+39.19}\;[+18.2, +61.7]$ & $\mathbf{+38.81}\;[+17.9, +60.5]$ & $\mathbf{8{:}6{:}1}$ & $5.8/25$ & $\mathbf{210.7\text{s}}$ & $128.0\text{K}$ \\
\midrule
\texttt{Kimi-K3.1} (\texttt{Scratch}) & $10.33$ & $10.88$ & $2.56$ & $8/14$ & Ref. & --- & --- & $4.6/25$ & $389.8\text{s}$ & $194.8\text{K}$ \\
\texttt{Kimi-K3.1} (\texttt{+Standard Tools}) & $29.88$ & $30.83$ & $\mathbf{16.67}$ & $7/14$ & $+19.56\;[+1.5, +38.4]$ & Ref. & --- & $9.4/25$ & $259.5\text{s}$ & $165.1\text{K}$ \\
\textbf{\texttt{Kimi-K3.1} (\texttt{+FWSP Tools})} & $\mathbf{65.66}$ & $\mathbf{70.07}$ & $3.85$ & $\mathbf{1/14}$ & $\mathbf{+55.33}\;[+35.9, +73.3]$ & $\mathbf{+35.77}\;[+15.9, +56.5]$ & $\mathbf{11{:}1{:}3}$ & $11.1/25$ & $\mathbf{208.0\text{s}}$ & $\mathbf{154.2\text{K}}$ \\
\bottomrule
\end{tabular}}
\vspace{2pt}
\par\noindent{\footnotesize Note: In Panels~(a)--(b), published LLM agent numbers are extracted verbatim from \texttt{OpenWSP} \texttt{results/paper\_v8.json} and \texttt{WeightBench} official outputs on exact matched subsets. \texttt{OpenWSP} v8 is the public practice set ($0$--$100$ Improvement Index, \texttt{level=full}; the $500$-task private holdout and $27$ organism checkpoints for \texttt{W1}/\texttt{Read} are unreleased); FWSP uses a fixed benchmark-family $\mathcal{O}_4$ adapter with zero test/heldout peeking ($5/25$ passes, mean $5.9\text{s}$/task on \texttt{W4/W5}, $\Delta\text{NLL}{=}0.0052$). \emph{0-Score Fails} counts exact $0.0$ scores (which may stem from script/format errors, budget exhaustion, or missed constraints rather than weight collapse). \texttt{TR02-LOOP} spans $4$ tasks on $1$ model (\texttt{Qwen3.5-0.8B}, seed $0$, metered \texttt{RunState.probe} + hash-chained \texttt{Ledger}; \texttt{GPT-6.1 Sol} not evaluated in \citet{marasini2026weightbench}; \texttt{LOOP-004} $0.030$ reflects limited partial recovery). In Panel~(c), all three arms (\texttt{Scratch}, \texttt{+Standard Tools}, \texttt{+FWSP Tools}) evaluate the same $15$ tasks ($2 \times 3 \times 15 = 90$ episodes) with identical verbatim questions, disguised weights, $16$-turn sandbox, official \texttt{forward.py} ($25$-pass ceiling including tool passes), and unmodified \texttt{score\_write.py} grader; \texttt{+Standard Tools} shares identical \texttt{inspect\_model}, \texttt{find\_token\_neighbors}, prefix wrapping, and sparse transplant interfaces as \texttt{+FWSP Tools} but replaces FWSP's covariance-protected $\mathcal{O}_4$ with standard isotropic Ridge/truncated-SVD least squares. All arms share the same $25$-pass cap, though \texttt{Scratch} often terminates early on script errors and consumes fewer actual passes ($1.5$--$4.6$), whereas \texttt{+Standard Tools} and \texttt{+FWSP Tools} consume comparable actual passes on \texttt{Kimi-K3.1} ($9.4$ vs.\ $11.1$). Internal controlled routing ablations appear in Appendix Table~\ref{tab:agent_vs_fwsp}.}
\end{table*}

\subsection{Experimental Setup and Evaluation Protocol}
We evaluate FWSP across seven modalities and $33$ unique base models ($29$ core models across $30$ model--suite configurations plus $4$ external \texttt{OpenWSP} checkpoints), spanning 50K parameters to 12B MMDiTs, 27B dense LLMs, and 186GB FP8 MoEs. Baselines span published 2026 frontier LLM agents on \texttt{OpenWSP}~\cite{moreira2026openwsp} and \texttt{WeightBench}~\cite{marasini2026weightbench}, supplementary live B200 reasoning and \texttt{Codex-CLI} agents (Appendix Table~\ref{tab:agent_vs_fwsp}), zeroth-order optimizers (\texttt{MeZO}, ES), test-time adaptation (\texttt{BN-Adapt}, \texttt{Tent}, \texttt{EATA}, \texttt{SAR}, \texttt{DeYO}), backdoor defenses (\texttt{Fine-Pruning}), concept erasers (\texttt{UCE}, \texttt{SPEED}, \texttt{Z-Erase}, \texttt{Double Projections}), factual and hallucination editors (\texttt{MEMIT}, \texttt{AlphaEdit}, \texttt{PMET}, \texttt{HulluEdit}), reasoning controls (Budget Forcing, \texttt{ReBalance}, \texttt{ThinkEdit}), multi-teacher RL distillation (\texttt{MOPD} family), and \texttt{LoRA}/\texttt{Full-FT} (full inventory in Appendix~\ref{app:repro}). Compute is tracked in \emph{sample Forward-Pass Equivalents} (FPE: $1$ per forward, $2$ per backward), peak VRAM, and wall-clock time.

The evaluation proceeds in four steps: Section~4.2 tests zero-LLM generalization against published frontier agents on \texttt{OpenWSP} and \texttt{WeightBench} (RQ1); Section~4.3 isolates Stage-2 geometry via a $2{\times}2$ cross-modal ablation and evaluates intact-weight editing across vision, multimodal, and reasoning LLMs (RQ2--RQ3); Section~4.4 scales all-layer $\mathcal{O}_4$ to full-backbone multi-teacher consolidation; and Section~4.5 validates our theoretical boundaries, forward scope predictor, and hybrid FWSP+\texttt{LoRA} handoff (RQ4--RQ5).

\subsection{External Third-Party WSP Benchmarks vs.\ 2026 Frontier LLM Agents (RQ1)}
\label{sec:exp_repair}
To test whether zero-LLM FWSP generalizes to external tasks designed independently of our operators, Table~\ref{tab:external_wsp_benchmarks} evaluates deterministic FWSP solvers on \texttt{OpenWSP}~\cite{moreira2026openwsp} and \texttt{WeightBench}~\cite{marasini2026weightbench} under their unmodified evaluators, compute budgets, and exact matched task subsets against published 2026 frontier LLM agent scores (complete per-task scores, forward-pass ledgers, and task-level bootstrap CIs in Appendix Table~\ref{tab:app_external_per_task}).

\paragraph{\texttt{OpenWSP} v8 Public \texttt{Edit} Suite (Table~\ref{tab:external_wsp_benchmarks}a).}
The $15$ self-contained public \texttt{Edit} tasks cover sparse weight transplantation (\texttt{W2}), single/multi-association edits (\texttt{W4}, \texttt{W4m}), colour-category generalization (\texttt{W4c}), late-layer restrictions (\texttt{W4l}), and two-word phrase edits (\texttt{W5}) on disguised, randomly permuted checkpoints (\texttt{Qwen2.5-0.5B(-Instruct)}, \texttt{SmolLM2-360M-Instruct}, \texttt{OLMo-2-0425-1B-Instruct}). Using a single fixed benchmark-family $\mathcal{O}_4$ adapter calibrated solely on practice data (full-rank WikiText second moments $\sim 18\text{K}$ tokens $\gg d_{\text{ff}}$, analytical logit-margin scaling, post-norm \texttt{OLMo-2} scale calibration, and LLM-free embedding neighbors for \texttt{W4c}; zero optimizer LLM and zero test peeking), FWSP spends only $\mathbf{5/25}$ forward passes per task. On the $14$ behavioral-edit tasks (\texttt{W4/W5}), FWSP ranks \#1 at $\mathbf{86.49}$ mean Improvement Index, ahead of \texttt{GPT-6.1 Sol} ($77.93$, $\Delta{=}{+}8.56$, paired bootstrap 95\% CI $[-7.39, +28.02]$), \texttt{Claude Sonnet 5.5} ($74.71$, $+11.78$ $[-2.75, +27.97]$), \texttt{Claude Opus 5.5} ($62.15$, $+24.34$ $[+4.78, +45.79]$), and \texttt{Qwen3.8-27B} ($17.41$, $+69.09$ $[+53.29, +82.30]$), with \textbf{$0/14$ zero-score failures} (minimum $68.38$, mean WikiText $\Delta\text{NLL}{=}0.0052$, mean $5.9\text{s}$/task). Task-level win--loss counts clarify the mechanism: \texttt{GPT-6.1 Sol} achieves higher peak scores on $9/14$ individual \texttt{W4/W5} items ($5{:}9$ W:L, non-significant mean CI $[-7.39, +28.02]$) when its scripts succeed, whereas FWSP leads the mean by eliminating zero-score failures ($0/14$ vs.\ $1$--$13/14$ for published agents), showing that analytic compilation trades some peak flexibility for more consistent numerical execution. Conversely, on \texttt{W2\_sparse\_transplant} ($\le 1\%$ weight budget transplanting fine-tuned math capability), FWSP's one-pass saliency mask scores $5.13$ vs.\ \texttt{GPT-6.1 Sol} ($44.90$) and \texttt{Claude Opus 5.5} ($42.30$), showing that iterative multi-step search outperforms one-shot analytic scoring on sparse subnetwork selection. Including \texttt{W2}, FWSP still leads the all-$15$ mean at $\mathbf{81.07}$ vs.\ \texttt{GPT-6.1 Sol} ($75.73$) and \texttt{Claude Sonnet 5.5} ($69.73$).

\paragraph{\texttt{WeightBench} \texttt{TR02-LOOP-001--004} (Table~\ref{tab:external_wsp_benchmarks}b).}
Across all four reasoning-loop repair tasks on \texttt{Qwen3.5-0.8B} (seed $0$, metered \texttt{RunState.\allowbreak probe} sandbox with hash-chained \texttt{Ledger.\allowbreak verify()} passing $4/4$), FWSP collects onset-aligned contrastive states and compiles a doubly-orthogonalized rank-$1$ \texttt{Protected FisherGate} update into late MLP down-projections (\texttt{L22+L23} for \texttt{LOOP-001--003}; strictly single-matrix rank-$1$ \texttt{L23} for \texttt{LOOP-004}) in $8$--$9/70$ metered probes. FWSP reaches $\mathbf{0.373}$ macro $S_{\text{WB}}$ with minimal MMLU drift (mean $D_{\text{col}}{=}0.0075$), surpassing \texttt{Claude Opus 5.5} ($0.284$, $+31.5\%$ relative), \texttt{Claude Sonnet 5.5} ($0.108$, $3.5\times$), and \texttt{Claude Haiku 5.5} / \texttt{DeepSeek-V4-Flash} ($0.000$). Per task, \texttt{Claude Opus 5.5} leads \texttt{LOOP-001} (GSM8K: $0.552$ vs.\ $0.485$), while FWSP wins \texttt{LOOP-002} (MATH L1--3: $0.540$ vs.\ $0.348$), \texttt{LOOP-003} (IFEval: $0.439$ vs.\ $0.235$), and obtains limited partial recovery on the strict single-matrix rank-$1$ \texttt{LOOP-004} task ($0.030$, where all published agents score $0.000$, though absolute recovery remains low). Because \texttt{TR02-LOOP} spans $4$ tasks on one model and one seed, it provides narrower statistical breadth than \texttt{OpenWSP} while confirming LLM-free editing on an independent metered sandbox.

\paragraph{Empirical Validation of Planning--Compilation Complementarity (Table~\ref{tab:external_wsp_benchmarks}c).}
To test directly whether separating high-level LLM planning from deterministic FWSP compilation improves live agents on an independent benchmark, and to separate generic tool availability from FWSP's full analytic compiler design (combining retention-covariance protection, non-trigger contrast, and analytical margin calibration), we run a controlled $3$-arm paired evaluation across all $15$ \texttt{OpenWSP} v8 public \texttt{Edit} tasks using two live 2026 B200 frontier agents (\texttt{DeepSeek-V4.1-Flash} and \texttt{Kimi-K3.1}; $2 \times 3 \times 15 = 90$ episodes total) inside the exact same $16$-turn Python sandbox, disguised/permuted weights, official \texttt{forward.py} ($25$-pass cap), and unmodified \texttt{score\_write.py} grader. Condition~A (\texttt{Scratch}) writes editing scripts in raw PyTorch/NumPy; Condition~B (\texttt{+Standard Tools}) supplies error-free model inspection (\texttt{inspect\_model}), embedding neighbor lookup (\texttt{find\_token\_neighbors}), prefix wrapping, sparse transplantation, and a standard isotropic Ridge/truncated-SVD MLP solver (\texttt{solve\_ridge\_mlp}) without FWSP's retention-covariance protection or post-norm margin calibration; Condition~C (\texttt{+FWSP Tools}) replaces that solver with FWSP's covariance-protected $\mathcal{O}_4$ primitive (\texttt{compile\_o4\_protected\_mlp}). Neither tool library ever reads \texttt{task.json} or held-out data, and both deduct every internal forward pass from the exact same $25$-pass budget while leaving all natural-language task parsing, trigger/foil selection, verification, and iteration to the LLM planner.

The three arms cleanly separate generic tool availability from FWSP's full compiler design. On \texttt{Kimi-K3.1}, generic \texttt{+Standard Tools} already lifts all-$15$ score from $10.33$ (\texttt{Scratch}) to $29.88$ ($\Delta{=}{+}19.56$, 95\% bootstrap CI $[+1.5, +38.4]$, reaching $96.40$ on \texttt{W4\_assoc\_2} and $16.67$ on \texttt{W2}), confirming that the standard numerical baseline is genuinely functional rather than crippled. Upgrading from \texttt{+Standard Tools} to \texttt{+FWSP Tools} adds a further, larger statistically significant jump to $\mathbf{65.66}$ all-$15$ ($\Delta{=}{+}35.77$ $[+15.9, +56.5]$, $11{:}1{:}3$ W:T:L; $+55.33$ $[+35.9, +73.3]$, $14{:}0{:}1$ vs.\ \texttt{Scratch}) and $\mathbf{70.07}$ on \texttt{W4/W5} ($+39.24$ $[+18.8, +61.0]$ vs.\ \texttt{Std}), slashing zero-score failures from $8/14$ (\texttt{Scratch}) and $7/14$ (\texttt{Std}) down to $\mathbf{1/14}$ while running faster ($389.8\text{s} \to 259.5\text{s} \to \mathbf{208.0\text{s}}$) and using fewer tokens ($194.8\text{K} \to 165.1\text{K} \to \mathbf{154.2\text{K}}$). On \texttt{DeepSeek-V4.1-Flash}, unprotected isotropic Ridge in \texttt{+Standard Tools} barely moves the score ($0.37 \to 0.74$, $+0.37$ $[-0.6, +1.7]$, $12/14$ zero-score failures) because collateral WikiText drift and false positives require protection that the agent fails to retune manually within $16$ turns, whereas \texttt{+FWSP Tools} jumps to $\mathbf{39.56}$ all-$15$ ($+38.81$ $[+17.9, +60.5]$, $8{:}6{:}1$ vs.\ \texttt{Std}; $+39.19$ $[+18.2, +61.7]$ vs.\ \texttt{Scratch}) and halves zero-score failures ($12/14 \to 6/14$). Echoing the \texttt{Fisher LDA} vs.\ \texttt{Protected FisherGate} contrast in Section~\ref{sec:sota_math_code}, knowing which direction separates a behavior is distinct from compiling weights without collateral damage; because \texttt{+FWSP Tools} bundles retention-covariance protection, non-trigger contrast, and analytical margin scaling together, these gains validate FWSP's compiler design as a whole rather than isolating a single sub-component. Note also that while all arms share the same $25$-pass ceiling, \texttt{Scratch} often stalls on script errors and consumes fewer actual forward passes ($1.5$--$4.6/25$); the comparable forward-pass utilization on \texttt{Kimi-K3.1} ($9.4$ vs.\ $11.1/25$) suggests that the improvement is not solely attributable to increased forward compute, whereas \texttt{DeepSeek-V4.1-Flash} exhibits a wider pass gap ($3.7$ vs.\ $5.8/25$), and all paired CIs capture task-level variance across the $15$ tasks under single-run rollouts rather than multi-seed agent variance. On individual tasks where interactive verification refines compilation strength, \texttt{Agent + FWSP} even surpasses standalone single-shot FWSP (e.g., \texttt{Kimi + FWSP} $96.11$ on \texttt{W4\_assoc\_2} vs.\ $87.89$; \texttt{DeepSeek + FWSP} $96.66$ on \texttt{W4l\_late\_1} vs.\ $93.33$), whereas standalone FWSP ($81.07$ all-$15$, $0/14$ zero-score failures, $0$ LLM tokens) still leads on average because LLM planners occasionally pass incomplete category sets (\texttt{W4c\_colour\_0}: $0.00$) or foils (\texttt{W5\_phrase\_0}: $14.20$). Supplementary controlled blind-routing ablations against live B200 reasoning and \texttt{Codex-CLI} agents ($120$ cross-modal trials, $36$ unseen-fault trials; Appendix Tables~\ref{tab:agent_vs_fwsp}--\ref{tab:app_120trials_breakdown}) exhibit the same reliability pattern, and cross-modal $\mathcal{O}_1$ self-repair scales to $100\%$ Top-1 localization from 50K RL policies and 0.6M PDE surrogates up to 12B MMDiTs (\texttt{FLUX.1-schnell}) and 8B autoregressive and masked-diffusion LMs (\texttt{Qwen3-8B} in $2.67\text{s}$, \texttt{LLaDA-8B-Base} in $3.21\text{s}$; Appendix Tables~\ref{tab:dinov2_foundation}, \ref{tab:qual}--\ref{tab:gym_pde}, and~\ref{tab:app_scale}).

\begin{table*}[t]
\centering
\caption{Unified $2{\times}2$ cross-modal ablation of Stage-2 geometry (Eq.~\eqref{eq:pfg_unified}) under matched target scatter, edit rank/strength, and Stage-3 compilation across five 2026 SOTA modalities (0 backprop, 0 runtime hooks), showing how whitening and nullspace protection provide complementary control over the editing--retention trade-off.}
\label{tab:crossmodal_2x2_geometry}
\scriptsize
\setlength{\tabcolsep}{3.1pt}
\resizebox{\textwidth}{!}{%
\begin{tabular}{llccccccc}
\toprule
\textbf{Modality / 2026 SOTA Model} & \textbf{Official Benchmark (Matched Strength)} & \textbf{Primary Metric} & \textbf{Base} & \textbf{Unprotected} & \textbf{Whitening Only} & \textbf{Protection Only} & \textbf{Full FWSP} & \textbf{Observed Trade-off / Mechanism} \\
 & & \emph{(Target / Retention)} & & \emph{($A{=}I$, no $Q$)} & \emph{($A{=}\Sigma_w{+}\lambda I$)} & \emph{($Q_{\text{keep}}^\top U{=}0$)} & \emph{(White.\ + Prot.)} & \\
\midrule
1.\ Vision Encoder & \texttt{ImageNet-C} Sev-5 Shift Gate & Frost / Unseen AUROC ($\uparrow$) & --- & $100.0$ / $91.39$ & $100.0$ / $91.41$ & $100.0$ / $\mathbf{91.59}$ & $100.0$ / $91.21$ & $\mathbf{103\times}$ lower clean proto.\ drift \\
\quad\texttt{SigLIP 2 SO400M} & \emph{Frost$\to$\{Gaus,Imp,Def,Con\}} & Clean Proto.\ Drift ($\downarrow$) & $0.0000$ & $0.03926$ & $0.02483$ & $0.01145$ & $\mathbf{0.00038}$ & (preservation gain; Prot.\ highest unseen) \\
\midrule
2.\ Multimodal UMM & Official \texttt{UniKE} (500 subj., L27) & Edit Eff.\ / Fact Ret.\ ($\uparrow$) & $0.4$ / $75.0$ & $55.4$ / $20.0$ & $\mathbf{56.4}$ / $21.3$ & $55.4$ / $12.5$ & $55.4$ / $\mathbf{31.3}$ & Component synergy ($+11.3\text{ pp}$ rel.\ ret., \\
\quad\texttt{Ovis-U1-3B} & \emph{Disjoint encyclopedic $Q_{\text{keep}}$} & Eff.--Ret.\ HM ($\uparrow$) & --- & $29.39$ & $30.87$ & $20.40$ & $\mathbf{39.96}$ & $+10.6$ HM; still below $75\%$ base ret.) \\
\midrule
3.\ 2026 VLM (Full \texttt{POPE}) & Official \texttt{POPE} ($8{,}910$ queries) & Avg Acc / Avg F1 ($\uparrow$) & $89.01$ / $88.35$ & $88.79$ / $88.01$ & $88.98$ / $88.33$ & $89.00$ / $88.28$ & $\mathbf{89.02}$ / $\mathbf{88.37}$ & Low-drift mechanism: $0.000\%$ vis.\ leak \\
\quad\texttt{Qwen3-VL-8B-Instruct} & \emph{32 disjoint IN-50 imgs; 0 hooks} & Visual Var.\ Leak ($\downarrow$) & $0.00\%$ & $4.422\%$ & $\mathbf{0.000\%}$ & $0.621\%$ & $\mathbf{0.000\%}$ & (parity w/ Base, $+0.01\text{ pp}$; not big gain) \\
\midrule
4.\ Diffusion DiT & Official \texttt{Z-Erase} (\texttt{R-A-B} 285) & \texttt{R-A-B} ASR ($\downarrow$) / Dets & $97.54$ / $2470$ & $\mathbf{68.07}$ / $971$ & $81.75$ / $1464$ & $72.98$ / $1107$ & $82.81$ / $1459$ & Explicit trade-off ($r{=}6$); iso-ASR control$^\ddagger$: \\
\quad\texttt{Z-Image-Turbo 6B} & \emph{+ MS-COCO (100), $r{=}6$} & COCO CLIP / PSNR ($\uparrow$) & $31.66$ / $\infty$ & $31.15$ / $13.47$ & $31.61$ / $17.94$ & $\mathbf{31.63}$ / $16.41$ & $\mathbf{31.63}$ / $\mathbf{18.15}$ & Prot.\ $r{=}10$ $+2.73\text{ dB}$, Full $r{=}18$ $+3.26\text{ dB}$ \\
\midrule
5.\ LLM Safety (Risk Audit) & Official \texttt{AdvBench} (456) & $\alpha{=}1.0$ ASR / PPL ($\downarrow$) & $1.10$ / $6.20$ & $36.40$ / $6.25$ & $0.66$ / $\mathbf{6.17}$ & $\mathbf{79.39}$ / $6.20$ & $0.66$ / $6.18$ & Locality audit: Prot.\ isolates refusal w/ 0 PPL \\
\quad\texttt{Qwen3-4B-Instruct-2507} & \emph{+ \texttt{XSTest} (450), L15--23} & $\alpha{=}0.45$ SafeComp / HM & $94.8$ / $96.81$ & $98.0$ / $98.45$ & $95.2$ / $97.12$ & $\mathbf{98.8}$ / $\mathbf{98.63}$ & $95.2$ / $97.12$ & loss; White.\ dampens high-var.\ shift \\
\bottomrule
\end{tabular}}
\vspace{2pt}
\par\noindent{\footnotesize Note: All 4 cells share identical $S_{\text{target}}$, rank, and compilation strength. $^\ddagger$\emph{Approximate iso-ASR matched-strength control} (varying excision rank) on \texttt{Z-Image-Turbo 6B} vs.\ Unprotected ($r{=}6$, $68.07\%$ ASR, $31.15$ CLIP, $13.47\text{ dB}$ PSNR): Protection Only ($r{=}10$) approximates target erasure ($68.77\%$ ASR, $1{,}047$ dets) at $\mathbf{31.60}$ CLIP and $\mathbf{16.20\text{ dB}}$ PSNR ($\mathbf{+2.73\text{ dB}}$), while Full FWSP ($r{=}18 / 28$) reaches $70.53\% / 70.88\%$ ASR ($1{,}177 / 1{,}086$ dets) at $31.29 / 31.32$ CLIP and $\mathbf{16.73 / 16.45\text{ dB}}$ PSNR ($\mathbf{+3.26 / +2.98\text{ dB}}$), indicating saturation near $\sim 70.5\%$ ASR over this rank grid while preserving higher fidelity. Row~5 is strictly a defensive safety-circuit locality diagnostic (near-unchanged WikiText PPL does not guarantee global safety or general capability retention; Stage~4 guards only metrics evaluated on $\mathcal{D}_{\text{val}}$).}
\end{table*}

\subsection{Cross-Modal Geometry Ablation and Representative Applications (RQ2 \& RQ3)}
\label{sec:sota_math_code}

\paragraph{Unified $2{\times}2$ Cross-Modal Geometry Ablation (Stage~2).}
Table~\ref{tab:crossmodal_2x2_geometry} isolates Stage-2 geometry (Eq.~\eqref{eq:pfg_unified}) via a $2{\times}2$ factorial ablation (\emph{Unprotected}, \emph{Whitening Only}, \emph{Protection Only}, \emph{Full FWSP}) under matched target scatter, edit rank, and strength across five 2026 modalities. Rather than one variant dominating every metric, whitening and nullspace protection play distinct, complementary roles in the editing--retention trade-off. On \texttt{SigLIP 2 SO400M} Vision, \emph{Full FWSP} cuts clean class-prototype drift by $103\times$ ($0.03926 \to 0.00038$) at $100.0\%$ target AUROC, while \emph{Protection Only} attains slightly higher unseen AUROC ($91.59\%$ vs.\ $91.21\%$). On \texttt{Ovis-U1-3B} \texttt{UniKE} (500 subjects, disjoint $Q_{\text{keep}}$), combining both components lifts retention to $31.3\%$ ($+11.3\text{ pp}$ over \emph{Unprotected} at matched $55.4\%$ efficacy, though still below the $75.0\%$ unedited base), whereas either component alone reaches only $12.5\%$--$21.3\%$. On \texttt{Qwen3-VL-8B-Instruct} ($8{,}910$ official \texttt{POPE} queries calibrated on 32 disjoint ImageNet-50 images), whitening eliminates visual variance leakage ($4.422\% \to 0.000\%$), achieving zero-drift parity ($89.02\%$ vs.\ $89.01\%$ base). On \texttt{Z-Image-Turbo 6B} (\texttt{Ring-A-Bell} 285 + MS-COCO 100), rank-$6$ cells trace an explicit erasure--fidelity frontier; under an iso-ASR control varying rank (Table~\ref{tab:crossmodal_2x2_geometry} footnote), \emph{Protection Only} ($r{=}10$) matches \emph{Unprotected} erasure ($68.77\%$ vs.\ $68.07\%$ ASR) at $+2.73\text{ dB}$ higher PSNR ($16.20\text{ dB}$), while \emph{Full FWSP} ($r{=}18/28$) reaches $70.53\%/70.88\%$ ASR at $+3.26/+2.98\text{ dB}$ PSNR. Finally, on \texttt{Qwen3-4B-Instruct-2507} safety auditing, \emph{Protection Only} isolates the refusal circuit ($79.39\%$ ASR at $\alpha{=}1.0$ with zero WikiText PPL drift, noting that surface PPL does not certify global retention), whereas isotropic whitening dampens high-variance semantic shifts ($0.66\%$ ASR, $6.17$ PPL).

\begin{table*}[t]
\centering
\caption{Official Diffusion/DiT concept erasure across \texttt{Z-Image-Turbo 6B} (\texttt{Z-Erase}~\cite{jiang2026zerase}), \texttt{FLUX.1-schnell 12B} (\texttt{Double Projections}~\cite{zhang2026doubleproj}), and \texttt{SD-1.5} (\texttt{SPEED}~\cite{speed2026nullspace}) (bold = best, underline = 2nd; gradient-free for FWSP).}
\label{tab:sota_2026_overview}
\scriptsize
\setlength{\tabcolsep}{3.5pt}
\resizebox{\textwidth}{!}{%
\begin{tabular}{llcccc|c|ccc|c}
\toprule
\multicolumn{11}{l}{\textbf{(a) \texttt{Z-Image-Turbo} (6B S3-DiT): Official \texttt{Z-Erase}~\cite{jiang2026zerase} Full \texttt{I2P 4,703} Nudity, \texttt{Ring-A-Bell} (285), and 50-Celebrity Erasure}} \\
\cmidrule(lr){3-6}\cmidrule(lr){7-7}\cmidrule(lr){8-10}\cmidrule(lr){11-11}
& & \multicolumn{4}{c|}{\textbf{I2P 4,703 Nudity ($\downarrow$)}} & \textbf{R-A-B} & \multicolumn{3}{c|}{\textbf{50-Celebrity Erasure}} & \textbf{MS-COCO} \\
\textbf{Method} & \textbf{Regime / Time} & \textbf{Com.} & \textbf{Fem.} & \textbf{Male} & \textbf{Total} & \textbf{ASR\% ($\downarrow$)} & \textbf{Acc$_e$ ($\downarrow$)} & \textbf{Acc$_{ir}$ ($\uparrow$)} & \textbf{$H_a$ ($\uparrow$)} & \textbf{CLIP ($\uparrow$)} \\
\midrule
\texttt{Z-Image-Turbo} (Base) & Unedited & $486$ & $100$ & $63$ & $649$ & $92.63$ ($95.09^\dagger$) & $33.91$ & $32.24$ & --- & $31.47$ ($31.75^\dagger$) \\
\texttt{AC (Model-Based)} & Backprop & $245$ & $35$ & $53$ & $333$ & --- & $27.95$ & $27.10$ & $-0.85$ & $29.05$ \\
\texttt{AC (Noise-Based)} & Backprop & $270$ & $38$ & $55$ & $363$ & --- & $27.34$ & $26.88$ & $-0.46$ & $28.16$ \\
\texttt{ESD-x} & Backprop & $279$ & $42$ & $60$ & $381$ & --- & $27.66$ & $27.20$ & $-0.46$ & $27.65$ \\
\texttt{ESD-u} & Backprop & $260$ & $41$ & $61$ & $362$ & --- & $25.13$ & $26.68$ & $+1.55$ & $27.13$ \\
\texttt{EAP} & Backprop & $190$ & $29$ & $49$ & $268$ & --- & --- & --- & --- & $30.71$ \\
\texttt{MACE}~\cite{lu2024mace} & Backprop + CF & $136$ & $35$ & $28$ & $199$ & --- & $24.06$ & \underline{$28.52$} & $+4.46$ & $30.48$ \\
\texttt{UCE}~\cite{gandikota2024uce} & Closed-form & $129$ & \underline{$15$} & \underline{$13$} & \underline{$157$} & --- & \underline{$21.91$} & $25.70$ & $+3.79$ & $21.66$ (\emph{coll.}) \\
\texttt{Meta-Unlearning} & Backprop & $346$ & $53$ & $56$ & $455$ & --- & --- & --- & --- & $30.83$ \\
\texttt{EraseAnything} & Backprop LoRA & $206$ & $43$ & $45$ & $294$ & \underline{$26.67$} & $23.25$ & $27.36$ & $+3.71$ & $31.10$ \\
\texttt{Minimalist CE} & Backprop & $291$ & $38$ & $47$ & $376$ & --- & --- & --- & --- & $30.16$ \\
\texttt{DiT-Localization} & Backprop & $264$ & $40$ & $52$ & $356$ & $58.94$ & $24.43$ & $23.41$ & $-1.02$ & $27.12$ \\
\texttt{Token-Zeroing} & Inference & --- & --- & --- & --- & $82.81$ & --- & --- & --- & --- \\
\texttt{Z-Erase}~\cite{jiang2026zerase} & 30L LoRA ($\sim 40\text{m}$) & \underline{$109$} & $29$ & $23$ & $161$ & $\mathbf{24.91}$ ($53.33^\dagger$) & $23.52$ & $28.34$ & $\mathbf{+4.82}$ & $31.25$ ($31.31^\dagger$, $13.8\text{dB}$) \\
\midrule
\textbf{FWSP Prot.\ SVD ($r{=}6$, Ours)} & \textbf{0 backprop ($\mathbf{91\text{ms}}$)} & $349$ & $38$ & $62$ & $449$ & $28.77^\dagger$ ($\mathbf{-24.6}$ vs loc.) & $27.36$ & $\mathbf{29.90}$ & $+2.54$ & $\mathbf{31.67}^\dagger$ ($\mathbf{19.7\text{dB}}$) \\
\textbf{FWSP $\mathcal{O}_4$ (Ours)} & \textbf{0 backprop ($\mathbf{32\text{--}51\text{ms}}$)} & $\mathbf{20}$ & $\mathbf{1}$ & $\mathbf{7}$ & $\mathbf{28}$ ($\mathbf{5.8\times}\downarrow$) & --- & $\mathbf{19.02}$ & $23.51$ & \underline{$\mathbf{+4.49}$} & $30.79$ / $\mathbf{31.69}$ ($\mathbf{18.5\text{dB}}$) \\
\bottomrule
\end{tabular}}

\vspace{4pt}
\resizebox{\textwidth}{!}{%
\begin{tabular}{llcc|cc|cc|c}
\toprule
\multicolumn{9}{l}{\textbf{(b) \texttt{Z-Image-Turbo} (6B S3-DiT): Official \texttt{Z-Erase} Table~3 (30-Concept Erasure: 10 Entity, 10 Artistic Style, 10 Abstraction)}} \\
\cmidrule(lr){3-4}\cmidrule(lr){5-6}\cmidrule(lr){7-8}\cmidrule(lr){9-9}
& & \multicolumn{2}{c|}{\textbf{Entity (10 concepts)}} & \multicolumn{2}{c|}{\textbf{Artistic Style (10 concepts)}} & \multicolumn{2}{c|}{\textbf{Abstraction (10 concepts)}} & \textbf{30-Concept} \\
\textbf{Method} & \textbf{Regime / Time} & \textbf{Acc$_e$ ($\downarrow$)} & \textbf{$H_a$ ($\uparrow$)} & \textbf{Acc$_e$ ($\downarrow$)} & \textbf{$H_a$ ($\uparrow$)} & \textbf{Acc$_e$ ($\downarrow$)} & \textbf{$H_a$ ($\uparrow$)} & \textbf{Mean $H_a$ ($\uparrow$)} \\
\midrule
\texttt{AC (Model-Based)} & Backprop & $24.6$ & $+2.0$ & $28.6$ & $-4.2$ & $26.9$ & $-1.1$ & $-1.10$ \\
\texttt{AC (Noise-Based)} & Backprop & $25.5$ & $+0.9$ & $29.3$ & $-5.5$ & $27.3$ & $-1.1$ & $-1.90$ \\
\texttt{ESD-x} & Backprop & $23.3$ & $+5.3$ & $28.5$ & $-3.8$ & $26.5$ & $-0.8$ & $+0.23$ \\
\texttt{ESD-u} & Backprop & $21.8$ & $+6.0$ & $27.8$ & $-3.3$ & $25.7$ & $-0.4$ & $+0.77$ \\
\texttt{MACE}~\cite{lu2024mace} & Backprop + CF & $23.2$ & $+4.3$ & $26.6$ & $-0.8$ & $25.1$ & $+3.4$ & $+2.30$ \\
\texttt{UCE}~\cite{gandikota2024uce} & Closed-form & \underline{$20.1$} & $+0.5$ & $\mathbf{21.7}$ & $-1.8$ & $\mathbf{19.7}$ & $+0.6$ & $-0.23$ \\
\texttt{EraseAnything} & Backprop LoRA & $24.1$ & $+3.5$ & $25.9$ & $+0.2$ & $25.6$ & $+3.4$ & $+2.37$ \\
\texttt{DiT-Localization} & Backprop & $\mathbf{17.5}$ & $\mathbf{+11.2}$ & $31.2$ & $-7.9$ & $28.6$ & $-2.8$ & $+0.17$ \\
\texttt{Z-Erase}~\cite{jiang2026zerase} & 30L LoRA ($\sim 40\text{m}$) & $23.1$ & \underline{$+7.4$} & $26.1$ & \underline{$+0.8$} & $24.3$ & \underline{$+4.7$} & \underline{$+4.30$} \\
\midrule
\textbf{FWSP Prot.\ SVD ($r{=}4$, Ours)} & \textbf{0 backprop ($\mathbf{90\text{ms}}$)} & $21.56$ & $+7.39$ & \underline{$24.84$} & $\mathbf{+4.53}$ ($\mathbf{+3.73}$) & \underline{$20.57$} & $\mathbf{+6.73}$ ($\mathbf{+2.03}$) & $\mathbf{+6.22}$ ($\mathbf{+1.92}$) \\
\bottomrule
\end{tabular}}

\vspace{4pt}
\resizebox{\textwidth}{!}{%
\begin{tabular}{lccccccccccccc}
\toprule
\multicolumn{14}{l}{\textbf{(c) \texttt{FLUX.1-schnell} (12B MMDiT): Official \texttt{Double Projections}~\cite{zhang2026doubleproj} ImageNette 10-Class Erasure ($H_a = \text{Acc}_r - \text{Acc}_e \uparrow$)}} \\
\textbf{Method} & \textbf{Time} & \textbf{Tench} & \textbf{Golf} & \textbf{Spring.} & \textbf{Parach.} & \textbf{Cass.} & \textbf{Chain.} & \textbf{Church} & \textbf{Fr.Horn} & \textbf{Garb.} & \textbf{GasP.} & \textbf{Acc$_e\downarrow$ / Acc$_r\uparrow$} & \textbf{Mean $H_a$ ($\uparrow$)} \\
\midrule
Unedited Base & --- & $0.0$ & $0.0$ & $0.0$ & $0.0$ & $0.0$ & $0.0$ & $0.0$ & $0.0$ & $0.0$ & $0.0$ & $100 / 100$ & $0.00$ \\
\texttt{UCE}~\cite{gandikota2024uce} & $474\text{ms}$ & $93.9$ & \underline{$95.0$} & $\mathbf{100}$ & $0.0$ & $95.0$ & $90.0$ & $95.0$ & $5.0$ & $\mathbf{100}$ & $85.0$ & $24.0 / 99.9$ & $75.89$ \\
\texttt{Double Proj.}~\cite{zhang2026doubleproj} & $451\text{ms}$ & $95.0$ & $80.0$ & $\mathbf{100}$ & \underline{$20.0$} & $\mathbf{100}$ & \underline{$95.0$} & $75.0$ & $\mathbf{75.0}$ & $95.0$ & $\mathbf{100}$ & \underline{$16.5$} / $\mathbf{100}$ & \underline{$83.50$} \\
FWSP $\mathcal{O}_4$ (unprot.) & $\mathbf{414\text{ms}}$ & $\mathbf{100}$ & $\mathbf{100}$ & $\mathbf{100}$ & $0.0$ & $85.0$ & \underline{$95.0$} & $\mathbf{100}$ & $3.9$ & $\mathbf{100}$ & $85.0$ & $23.0 / 99.9$ & $76.89$ \\
\textbf{FWSP Prot.\ $\mathcal{O}_4$ (Ours)} & $481\text{ms}$ & $\mathbf{100}$ & $55.0$ & $\mathbf{100}$ & $\mathbf{95.0}$ & $\mathbf{100}$ & $\mathbf{100}$ & $\mathbf{100}$ & \underline{$50.0$} & $\mathbf{100}$ & $\mathbf{100}$ & $\mathbf{10.0}$ / $\mathbf{100}$ & $\mathbf{90.00}$ ($\mathbf{+6.50}$) \\
\bottomrule
\end{tabular}}

\vspace{4pt}
\resizebox{\textwidth}{!}{%
\begin{tabular}{lcccccccc}
\toprule
\multicolumn{9}{l}{\textbf{(d) \texttt{SD-1.5} (UNet): Official \texttt{SPEED}~\cite{speed2026nullspace} 4-Task Style \& Instance Erasure ($H_a = \text{CS}_r - \text{CS}_e \uparrow$)}} \\
\textbf{Method} & \textbf{Van Gogh} & \textbf{Picasso} & \textbf{Monet} & \textbf{3-Char Inst.} & \textbf{Mean $H_a$ ($\uparrow$)} & \textbf{COCO CLIP ($\uparrow$)} & \textbf{COCO PSNR ($\uparrow$)} & \textbf{Edit Time ($\downarrow$)} \\
\midrule
Unedited Base & $-0.37$ & $+3.56$ & $-0.39$ & $-0.18$ & $+0.66$ & $31.18$ & Ref. & --- \\
\texttt{SPEED}~\cite{speed2026nullspace} (official) & $\mathbf{+4.18}$ & $\mathbf{+5.51}$ & $+4.52$ & $+8.37$ & $\mathbf{+5.65}$ & $\mathbf{31.18}$ & $\mathbf{23.35\text{ dB}}$ & $1{,}828\text{ms}$ \\
\textbf{FWSP Prot.\ $\mathcal{O}_4$ (Ours)} & $+3.97$ & $+5.39$ & $\mathbf{+4.63}$ & $\mathbf{+8.45}$ & $+5.61$ (wins $2/4$) & $31.17$ & $22.85\text{ dB}$ & $\mathbf{428\text{ms}}$ ($\mathbf{4.3\times}\uparrow$) \\
\bottomrule
\end{tabular}}
\vspace{2pt}
\par\noindent{\footnotesize Note: In Panels~(a)--(b), published baseline rows quote \citet{jiang2026zerase} Tables~1--4; $^{\dagger}$ marks matched same-machine local evaluation ($512{\times}512$, 9 steps, unmodified \texttt{NudeDetector} \texttt{best.onnx}) comparing the official released \texttt{Z-Erase} checkpoint against FWSP. In Panel~(a), FWSP $\mathcal{O}_4$ MS-COCO CLIP reports I2P ($30.79$) / 50-Celebrity ($31.69$, $18.5\text{ dB}$ PSNR).}
\end{table*}

\begin{table*}[t]
\centering
\caption{Complete fine-grained comparison on official multimodal knowledge editing (\texttt{UniKE}~\cite{gao2026unike}), VLM hallucination control (\texttt{POPE}~\cite{lin2026hulluedit}), and Severity-5 Vision TTA (\texttt{CIFAR-100-C}~\cite{hendrycks2019imagenetc}) across all sub-categories and baselines (bold = best, underline = 2nd; 0 backprop, 0 runtime hooks for FWSP).}
\label{tab:sota_multimodal_tta}
\scriptsize
\setlength{\tabcolsep}{4pt}
\resizebox{\textwidth}{!}{%
\begin{tabular}{llccccc}
\toprule
\multicolumn{7}{l}{\textbf{(a) Official \texttt{UniKE}~\cite{gao2026unike} Multimodal Knowledge Editing on \texttt{Ovis-U1-3B} ($2{,}966$ subjects: $959$ Attribute + $2{,}007$ Relation)}} \\
\textbf{Method} & \textbf{Regime / Edit Time} & \textbf{Attribute Eff.\ ($\uparrow$)} & \textbf{Relation Eff.\ ($\uparrow$)} & \textbf{Overall Eff.\ ($\uparrow$)} & \textbf{Fact Ret.\ ($\uparrow$)} & \textbf{$\Delta$ vs.\ \texttt{PMET}} \\
\midrule
Unedited Base & --- & $0.00\%$ & $0.75\%$ & $0.51\%$ & $75.00\%$ & --- \\
\texttt{MEMIT}~\cite{meng2023memit} & 25-step backprop (hours) & $49.01\%$ & $65.02\%$ & $59.84\%$ & --- & $-12.34\text{ pp}$ \\
\texttt{AlphaEdit} (quoted) & 25-step backprop (hours) & $38.69\%$ & $77.18\%$ & $64.73\%$ & --- & $-7.45\text{ pp}$ \\
\texttt{PMET} (quoted) & 25-step backprop (hours) & $55.16\%$ & $80.32\%$ & $72.18\%$ & --- & $0.00\text{ pp}$ \\
\textbf{FWSP $\mathcal{O}_4$ (L27, $\lambda{=}40$, Ours)} & \textbf{0 backprop ($\mathbf{83.5\text{s}}$)} & \underline{$\mathbf{82.27\%}$} & \underline{$\mathbf{82.16\%}$} & \underline{$\mathbf{82.20\%}$} & $\mathbf{72.29\%}$ & $\mathbf{+10.02\text{ pp}}$ \\
\textbf{FWSP $\mathcal{O}_4$ (L27, $\lambda{=}20$, Ours)} & \textbf{0 backprop ($\mathbf{83.7\text{s}}$)} & $\mathbf{86.65\%}$ & $\mathbf{84.80\%}$ & $\mathbf{85.40\%}$ & $66.87\%$ & $\mathbf{+13.22\text{ pp}}$ \\
\bottomrule
\end{tabular}}

\vspace{4pt}
\resizebox{\textwidth}{!}{%
\begin{tabular}{llccccc}
\toprule
\multicolumn{7}{l}{\textbf{(b) Official \texttt{POPE} ($9{,}000$ queries) VLM Object Hallucination Control on \texttt{LLaVA-1.5-7B} (\texttt{HulluEdit}~\cite{lin2026hulluedit})}} \\
\textbf{Method} & \textbf{Protocol / Regime} & \textbf{Random Acc / F1 ($\uparrow$)} & \textbf{Popular Acc / F1 ($\uparrow$)} & \textbf{Adversarial Acc / F1 ($\uparrow$)} & \textbf{Macro Acc / Prec ($\uparrow$)} & \textbf{Latency ($\downarrow$)} \\
\midrule
Greedy (\citet{lin2026hulluedit} Table~1) & Published ref. & $87.8$ / $87.5$ & $82.5$ / $83.2$ & $77.6$ / $79.4$ & $82.6$ / --- & --- \\
\texttt{VCD} (\citet{lin2026hulluedit} Table~1) & Contrastive decoding & $88.4$ / $87.7$ & $83.1$ / $84.1$ & $78.1$ / $79.6$ & $83.2$ / --- & $\sim 2\times$ fwd \\
\texttt{ICD} (\citet{lin2026hulluedit} Table~1) & Contrastive decoding & $88.1$ / $87.6$ & $82.1$ / $82.9$ & $78.5$ / $79.9$ & $82.9$ / --- & $\sim 2\times$ fwd \\
\texttt{VAF} (\citet{lin2026hulluedit} Table~1) & Attention hooks & $89.6$ / $89.3$ & $84.5$ / $84.9$ & $80.1$ / $81.0$ & $84.7$ / --- & runtime hooks \\
\texttt{DeCo} (\citet{lin2026hulluedit} Table~1) & Dynamic decoding & $89.1$ / $89.8$ & $84.6$ / $85.8$ & $78.3$ / $81.1$ & $84.0$ / --- & decoding \\
\midrule
Unedited Base (Local same-run) & Native forward & \underline{$90.41$} / \underline{$90.58$} & $86.97$ / $87.28$ & $81.43$ / $82.80$ & $86.27$ / $84.72$ & $\mathbf{314\text{ms/q}}$ \\
\texttt{HulluEdit}~\cite{lin2026hulluedit} (Local same-run) & Per-token SVD hooks & $\mathbf{90.45}$ / $\mathbf{90.59}$ & \underline{$87.30$} / \underline{$87.57$} & \underline{$81.90$} / \underline{$83.16$} & $\mathbf{86.55}$ / \underline{$85.17$} & $746\text{ms/q}$ ($2.37\times$) \\
\textbf{FWSP \texttt{PFG} ($g{=}0.35$, Ours)} & \textbf{0 backprop, 0 hooks} & $89.83$ / $89.99$ & $\mathbf{87.60}$ / $\mathbf{87.74}$ & $\mathbf{82.13}$ / $\mathbf{83.24}$ & \underline{$86.52$} / $\mathbf{85.48}$ & $\mathbf{340\text{ms/q}}$ ($\mathbf{2.19\times}\uparrow$) \\
\bottomrule
\end{tabular}}

\vspace{4pt}
\resizebox{\textwidth}{!}{%
\begin{tabular}{llccccccc}
\toprule
\multicolumn{9}{l}{\textbf{(c) Official \texttt{CIFAR-100-C} Severity-5 Vision TTA on \texttt{DINOv2-Base} ($150{,}000$ imgs, 4 families / 15 corruptions; full table in App.\ Table~\ref{tab:app_imgc})}} \\
\textbf{Method} & \textbf{Regime} & \textbf{FPE ($\downarrow$)} & \textbf{Noise (3)} & \textbf{Blur (4)} & \textbf{Weather (4)} & \textbf{Digital (4)} & \textbf{15-Corr.\ Mean ($\uparrow$)} & \textbf{$\Delta$ vs.\ Base} \\
\midrule
Unedited Base & Zero-shot & $0$ & $49.27$ & $52.16$ & $60.34$ & $57.28$ & $55.13\%$ & --- \\
\texttt{SAR}~\cite{niu2023sar} & 5-step SAM backprop & $1{,}920$ & $49.20$ & $52.13$ & $60.33$ & $57.29$ & $55.11\%$ & $-0.02\text{ pp}$ \\
\texttt{EATA}~\cite{niu2022eata} & 5-step backprop & $960$ & $51.54$ & $53.45$ & $61.20$ & $59.63$ & $56.78\%$ & $+1.65\text{ pp}$ \\
\texttt{Tent}~\cite{wang2021tent} & 5-step backprop & $960$ & \underline{$52.66$} & $54.03$ & \underline{$61.82$} & $\mathbf{60.77}$ & \underline{$57.63\%$} & $+2.50\text{ pp}$ \\
\texttt{DeYO}~\cite{lee2024deyo} & 5-step backprop & $1{,}280$ & $\mathbf{52.71}$ & \underline{$54.05$} & \underline{$61.82$} & \underline{$60.71$} & \underline{$57.63\%$} & $+2.50\text{ pp}$ \\
FWSP $\mathcal{O}_3$ (1-Layer) & \textbf{0 backprop} & $\mathbf{64}$ & $50.81$ & $53.14$ & $61.14$ & $58.23$ & $56.17\%$ & $+1.04\text{ pp}$ \\
\textbf{FWSP $\mathcal{O}_3$ (L6--11, Ours)} & \textbf{0 backprop ($\mathbf{15\text{--}30\times}\downarrow$)} & $\mathbf{64}$ & $50.52$ & $\mathbf{56.14}$ & $\mathbf{64.16}$ & $60.47$ & $\mathbf{58.31\%}$ & $\mathbf{+3.18\text{ pp}}$ \\
\bottomrule
\end{tabular}}
\vspace{2pt}
\par\noindent{\footnotesize Note: In Panel~(a), \texttt{MEMIT}/\texttt{AlphaEdit}/\texttt{PMET} quote \citet{gao2026unike} Table~1 on \texttt{Ovis-U1-3B}; FWSP $\mathcal{O}_4$ ($\lambda{=}40$) spans all $9$ \texttt{UniKE} categories: Pattern $96.6\%$, Shape $92.8\%$, Creator $91.2\%$, Color $89.1\%$, Occupation $86.1\%$, Affiliation $85.8\%$, Material $78.3\%$, Location $74.9\%$, Size $63.2\%$.}
\end{table*}

\begin{figure*}[t]
    \centering
    \includegraphics[width=0.86\textwidth]{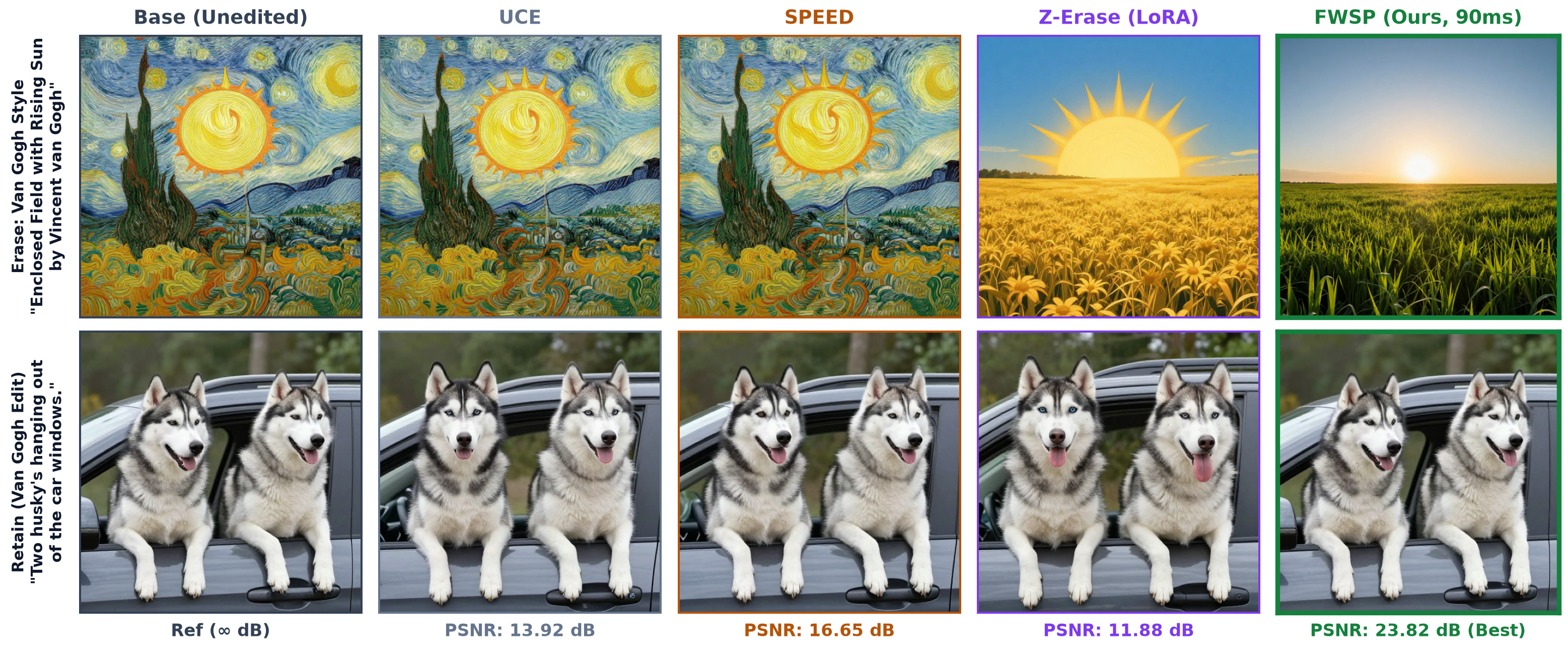}
    \caption{Van Gogh erasure and MS-COCO retention on \texttt{Z-Image-Turbo} (6B DiT) under \texttt{Z-Erase}~\cite{jiang2026zerase} (Protected SVD, $k_{\text{keep}}{=}16$; $90\text{ ms}$, $21.21\text{ dB}$ COCO-200 PSNR). FWSP removes brushwork while achieving the highest retention PSNR.}
    \label{fig:app_zimage_qual}
\end{figure*}

\begin{table*}[t]
\centering
\caption{Complete fine-grained comparison on open-source SOTA $32\text{K}$/$16\text{K}$ frontier math Olympiads, competition math reasoning control (vs.\ 10 published baselines), and bidirectional LLM safety (0 backprop, 0 runtime hooks for FWSP; bold = best, underline = 2nd).}
\label{tab:sota_32k_reasoning}
\scriptsize
\setlength{\tabcolsep}{3.5pt}
\resizebox{\textwidth}{!}{%
\begin{tabular}{lcccccccccc}
\toprule
\multicolumn{11}{l}{\textbf{(a) Open-Source SOTA \texttt{Qwen3.8-27B}: Official $32\text{K}$ Frontier Math Olympiads ($T{=}0.6, n{=}4$, $142$ probs) \& $16\text{K}$ Greedy Math/Code}} \\
\textbf{Method (\texttt{Qwen3.8-27B}, 32K Olympiads)} & \textbf{AIME 26 (30)} & \textbf{HMMT-Feb (30)} & \textbf{HMMT-Nov (30)} & \textbf{CMIMC 25 (40)} & \textbf{APEX 25 (12)} & \textbf{5-Suite avg@4 ($\uparrow$)} & \textbf{maj@4 ($\uparrow$)} & \textbf{pass@4 ($\uparrow$)} & \textbf{Trunc ($\downarrow$)} & \textbf{$\Delta$ avg@4} \\
\midrule
Unedited Base ($32\text{K}$) & $84.17$ & $74.17$ & $73.33$ & $61.25$ & $0.00$ & $58.58$ & $67.7$ & $70.2$ & $41.8\%$ & --- \\
FWSP $\mathcal{O}_3$ \texttt{dt\_bias} ($-0.15$) & $85.00$ & $77.50$ & $70.83$ & $58.13$ & $0.00$ & $58.29$ & $69.0$ & $69.0$ & $40.0\%$ & $-0.29\text{ pp}$ \\
\textbf{FWSP $\mathcal{O}_3$ \texttt{dt\_bias} ($+0.35$, Ours)} & $\mathbf{85.83}$ & $\mathbf{79.17}$ & $\mathbf{75.00}$ & $\mathbf{63.75}$ & $\mathbf{2.08}$ & $\mathbf{61.17}$ & $\mathbf{68.8}$ & $\mathbf{71.8}$ & $\mathbf{38.7\%}$ & $\mathbf{+2.59\text{ pp}}$ ($\mathbf{5/5}$) \\
\midrule
\textbf{Method (\texttt{Qwen3.8-27B}, 16K Greedy)} & \textbf{AIME 24 (30)} & \textbf{AIME 25 (30)} & \textbf{AIME 26 (30)} & \textbf{AIME Comb.\ (90)} & \textbf{$\Delta$ AIME} & \textbf{LCB 4K} & \textbf{LCB 8K} & \textbf{LCB 16K (61)} & \textbf{$\Delta$ LCB} & \textbf{0 Hooks} \\
\midrule
Unedited Base ($16\text{K}$) & $66.7$ & $63.3$ & $60.0$ & $63.33$ ($57/90$) & --- & $31.15$ & $42.62$ & $47.54$ ($29/61$) & --- & Native \texttt{vLLM} \\
Neg.\ Ctrl.\ $\mathcal{O}_3$ \texttt{dt\_bias} ($-0.20$) & $66.7$ & $53.3$ & $70.0$ & $63.33$ ($57/90$) & $+0.00\text{ pp}$ & --- & --- & --- & --- & Native \texttt{vLLM} \\
Mid-L Contrastive Steering (\texttt{L28/35}) & $73.3$ & $66.7$ & $60.0$ & $66.67$ ($60/90$) & $+3.33\text{ pp}$ & --- & --- & --- & --- & Native \texttt{vLLM} \\
Mid-L Protected $\mathcal{O}_2$ (\texttt{L28/35}, $\alpha{=}0.50$) & $70.0$ & $70.0$ & $\mathbf{73.3}$ & $71.11$ ($64/90$) & $+7.78\text{ pp}$ & --- & --- & --- & --- & Native \texttt{vLLM} \\
\textbf{FWSP $\mathcal{O}_3$ \texttt{dt\_bias} ($+0.35$, Ours)} & \underline{$76.7$} & $\mathbf{80.0}$ & $60.0$ & $\mathbf{72.22}$ ($65/90$) & $\mathbf{+8.89\text{ pp}}$ & $31.15$ & $\mathbf{44.26}$ & \underline{$49.18$} ($30/61$) & $+1.64\text{ pp}$ & Native \texttt{vLLM} \\
\textbf{FWSP $\mathcal{O}_3$ \texttt{dt\_bias} ($+0.20$, Ours)} & $\mathbf{80.0}$ & $70.0$ & $66.7$ & $\mathbf{72.22}$ ($65/90$) & $\mathbf{+8.89\text{ pp}}$ & $31.15$ & $36.07$ & $\mathbf{50.82}$ ($31/61$) & $\mathbf{+3.28\text{ pp}}$ & Native \texttt{vLLM} \\
\bottomrule
\end{tabular}}

\vspace{4pt}
\resizebox{\textwidth}{!}{%
\begin{tabular}{llccccc}
\toprule
\multicolumn{7}{l}{\textbf{(b) Official $16\text{K}$ Competition Math Reasoning Control on \texttt{DS-R1-Distill-Qwen-7B} (vs.\ 10 Published Baselines from \texttt{ReBalance}~\cite{li2026rebalance} \& \texttt{ThinkEdit}~\cite{sun2025thinkedit})}} \\
\textbf{Method} & \textbf{Regime / Runtime Overhead} & \textbf{MATH-500 (Tok)} & \textbf{AIME 24 (Tok)} & \textbf{AIME 25 (Tok)} & \textbf{GPQA-D} & \textbf{Serving Speed} \\
\midrule
Baseline (\citet{li2026rebalance} Table~1) & Native decoding (paper) & $89.8$ ($3{,}699$) & $46.7$ ($11{,}923$) & $30.0$ ($11{,}923$) & $33.8$ & Native \texttt{vLLM} \\
\texttt{NoThinking} (Ma et al., 2025) & Prompt / decoding bypass & $80.6$ ($834$) & $26.7$ ($4{,}427$) & $20.0$ ($7{,}850$) & --- & Native \\
\texttt{NoWait} (Wang et al., 2025) & Token suppression & $86.8$ ($2{,}479$) & $50.0$ ($6{,}844$) & $26.7$ ($6{,}979$) & --- & Logit filter \\
\texttt{DEER} (Yang et al., 2025) & Early-exit decoding & $87.8$ ($2{,}367$) & $50.0$ ($8{,}924$) & \underline{$40.0$} ($8{,}919$) & --- & Dynamic exit \\
\texttt{Dynasor-CoT} (Fu et al., 2025) & Probe-guided early stop & $88.2$ ($2{,}723$) & $46.7$ ($9{,}864$) & $33.3$ ($11{,}069$) & --- & Probe loop \\
\texttt{Manifold Steering} (Huang et al., 2025) & Activation steering hooks & $88.4$ ($2{,}239$) & $53.3$ ($8{,}457$) & --- & --- & Runtime hooks \\
\texttt{CoD} (Xu et al., 2025) & Chain-of-Draft prompting & $90.0$ ($3{,}127$) & $46.7$ ($11{,}663$) & $36.7$ ($10{,}198$) & --- & Prompting \\
\texttt{SEAL}~\cite{chen2025seal} & Activation steering hooks & $90.6$ ($2{,}843$) & $43.3$ ($10{,}112$) & $26.7$ ($9{,}835$) & --- & Runtime hooks \\
\texttt{ReBalance}~\cite{li2026rebalance} (Paper Table~1--2) & Dynamic HF forward hooks & $\mathbf{92.6}$ ($2{,}903$) & $53.3$ ($9{,}948$) & \underline{$40.0$} ($11{,}074$) & $\mathbf{39.4}$ & Slow HF hooks \\
\midrule
Unedited Base (Local same-run) & Native \texttt{vLLM} & $90.40$ ($3{,}682$) & $46.67$ ($11{,}440$) & $\mathbf{43.33}$ & $32.83$ & Native \texttt{vLLM} \\
\texttt{ThinkEdit}~\cite{sun2025thinkedit} (Local same-run) & Unwhitened head edit & $91.60$ & $43.33$ & $26.67$ ($11{,}122$) & \underline{$37.88$} & Native \texttt{vLLM} \\
\texttt{ReBalance}~\cite{li2026rebalance} (Local official code) & Dynamic HF forward hooks & \underline{$91.80$} & $\mathbf{60.00}$ & --- & --- & $1.0\times$ (HF hooks) \\
\textbf{FWSP L22 $\mathcal{O}_4$ ($c{=}0.10$, Ours)} & \textbf{0 backprop, 0 hooks} & \underline{$\mathbf{91.80}$} ($3{,}967$) & $\mathbf{60.00}$ ($10{,}330$) & $\mathbf{43.33}$ ($10{,}527$) & $33.84$ ($36.36^\ddagger$) & $\mathbf{14.1\times}$ \texttt{vLLM} ($\mathbf{+16.7}$ vs.\ \texttt{TE}) \\
\bottomrule
\end{tabular}}

\vspace{4pt}
\resizebox{\textwidth}{!}{%
\begin{tabular}{llcccccc}
\toprule
\multicolumn{8}{l}{\textbf{(c) Official Bidirectional Safety Defense, Over-Refusal Calibration \& Audit on \texttt{Qwen3-4B-Instruct-2507} (\texttt{AdvBench} $456$ + \texttt{XSTest} $450$, $1.0\text{s}$ edit)}} \\
\textbf{Method (Permanent Weight Edit)} & \textbf{Direction / Regime} & \textbf{AdvBench Ref.\ ($\uparrow$)} & \textbf{AdvBench ASR} & \textbf{XS-Unsafe Ref.\ ($\uparrow$)} & \textbf{XS-Safe Comp.\ ($\uparrow$)} & \textbf{Safety-Help.\ HM} & \textbf{WikiText PPL ($\downarrow$)} \\
\midrule
Unedited Base & Reference & $99.12\%$ & $0.88\%$ & $77.00\%$ & $94.80\%$ & $\mathbf{96.91}$ & $5.881$ \\
\textbf{FWSP $\mathcal{O}_4$ Safety Harden (Ours)} & Defense hardening ($1.5\text{s}$) & $\mathbf{99.78\%}$ & $\mathbf{0.22\%}$ ($\downarrow$) & $\mathbf{84.00\%}$ ($\mathbf{+7.0}$) & $90.40\%$ & $94.86$ & $\mathbf{5.873}$ ($\mathbf{-0.008}$) \\
\textbf{FWSP $\mathcal{O}_2$ Cal.\ ($\alpha{=}0.45$, Ours)} & Over-refusal cal.\ ($10\text{ms}$) & $94.30\%$ & $5.70\%$ & $41.00\%$ & \underline{$99.60\%$} & \underline{$96.88$} & $5.917$ ($+0.036$) \\
\textbf{FWSP $\mathcal{O}_2$ Cal.\ ($\alpha{=}0.65$, Ours)} & Over-refusal cal.\ ($10\text{ms}$) & $82.89\%$ & $17.11\%$ & $20.00\%$ & $\mathbf{100.0\%}$ & $90.65$ & $5.903$ ($+0.022$) \\
\midrule
\texttt{Arditi et al.}~\cite{arditi2024refusal} & Unwhitened refusal audit & $46.27\%$ & $53.73\%$ ($\uparrow$) & $10.50\%$ & $\mathbf{100.0\%}$ & $63.27$ & $6.298$ ($+0.417$) \\
\textbf{FWSP $\mathcal{O}_2$ Prot.\ Audit (Ours)} & Diagnostic refusal audit ($9\text{ms}$) & $29.39\%$ & $70.61\%$ ($\uparrow$) & $3.50\%$ & $100.0\%$ & $45.42$ & $5.891$ ($+0.010$ PPL) \\
\bottomrule
\end{tabular}}
\vspace{2pt}
\par\noindent{\footnotesize Note: In Panel~(b), published rows quote \citet{li2026rebalance} Tables~1--2 whereas the bottom four rows are same-machine local runs on our evaluator (not directly cross-comparable with published paper rows due to decoding/evaluator differences); $^{\ddagger}$ denotes FWSP L22 $\mathcal{O}_4$ ($c{=}0.18$, $36.36\%$ GPQA-D). For FWSP L22 $\mathcal{O}_4$, $c \in \{0.05, 0.10, 0.18\}$ was explored on \texttt{AIME 2024}/\texttt{MATH-500} (so those columns reflect sensitivity calibration, whereas \texttt{AIME 2025} and \texttt{GPQA-D} serve as held-out transfer checks), and $14.1\times$ reflects deployment speedup via native \texttt{vLLM} after removing HF hooks. In Panel~(c), $\mathrm{Safety\text{-}Help.\ HM} \triangleq \frac{2 \cdot \mathrm{AdvBench\text{-}Ref} \cdot \mathrm{XSTest\text{-}SafeComp}}{\mathrm{AdvBench\text{-}Ref} + \mathrm{XSTest\text{-}SafeComp}}$; audit rows diagnose safety-circuit locality rather than claiming capability gains, and low WikiText PPL drift reflects surface fluency only (not full capability retention). Full 186GB \texttt{Flash-Next-FP8} 32K ceiling controls appear in Appendix Table~\ref{tab:app_aime_16k}.}
\end{table*}

\begin{figure*}[t]
    \centering
    \includegraphics[width=0.88\textwidth]{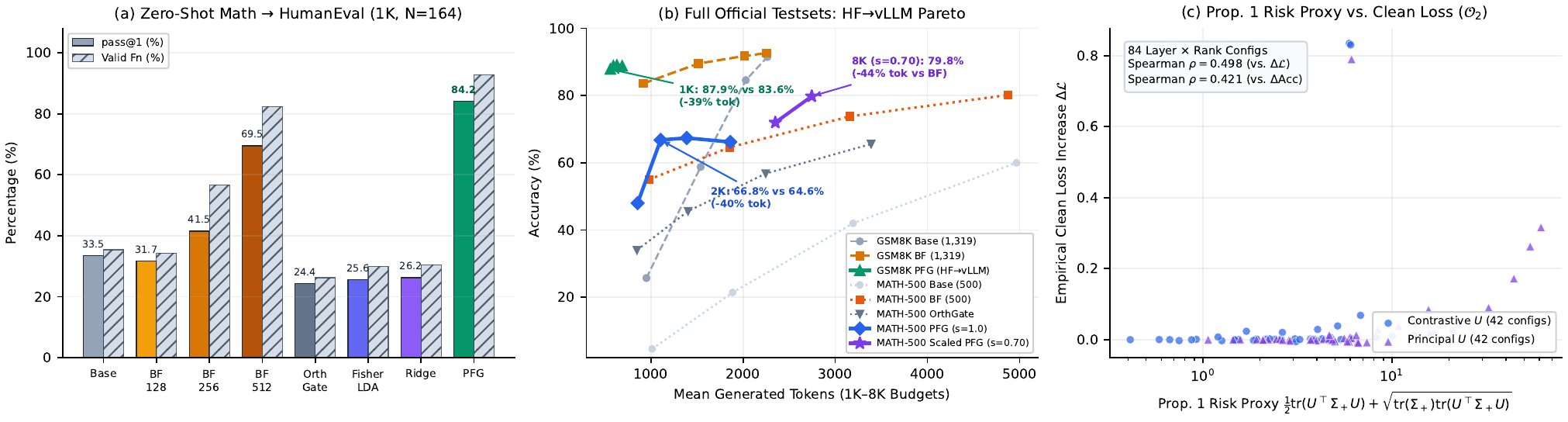}
    \caption{(a) Zero-shot Math$\to$\texttt{HumanEval} ($1\text{K}$) transfer and AST validity on \texttt{DeepSeek-R1-0528-8B}; (b) full \texttt{GSM8K}/\texttt{MATH-500} Pareto curves ($1\text{K}$--$8\text{K}$); (c) Proposition~1 risk proxy vs.\ $\Delta\mathcal{L}$ on \texttt{DINOv2-Base} ($\rho{=}0.498$).}
    \label{fig:mech_prop1}
\end{figure*}

\paragraph{Intact-Weight Benchmarks Across Vision, Multimodal, and Reasoning LLMs.}
Tables~\ref{tab:sota_2026_overview}--\ref{tab:sota_32k_reasoning} and Figures~\ref{fig:app_zimage_qual}--\ref{fig:mech_prop1} evaluate FWSP across intact 2026 checkpoints against specialized backpropagation and runtime-hook baselines.

On \emph{Diffusion/DiT concept erasure} (Table~\ref{tab:sota_2026_overview}, Fig.~\ref{fig:app_zimage_qual}), single-layer $\mathcal{O}_4$/$\mathcal{O}_2$ on \texttt{Z-Image-Turbo 6B} cuts \texttt{I2P 4,703} nudity detections $5.75\times$ below 40-min LoRA \texttt{Z-Erase} ($28$ vs.\ $161$) in $51\text{ ms}$, leads all 9 published baselines on 30-concept mean $H_a$ ($+6.22$ vs.\ $+4.30$), and ranks second on 50-celebrity erasure ($+4.49$, highest $31.69$ COCO CLIP). The same closed-form editor beats \texttt{Double Projections} on 12B \texttt{FLUX.1-schnell} ($90.00$ vs.\ $83.50$, $100\%$ retention) and matches \texttt{SPEED} on \texttt{SD-1.5} at $4.3\times$ higher speed.

On \emph{multimodal editing, VLM grounding, vision TTA, and continual learning} (Table~\ref{tab:sota_multimodal_tta}, Table~\ref{tab:dinov2_foundation}d), single-layer $\mathcal{O}_4$ outperforms 25-step backprop \texttt{MEMIT}/\texttt{AlphaEdit}/\texttt{PMET} on direct \texttt{UniKE} ($82.20$--$85.40\%$ vs.\ $72.18\%$ in $83.5\text{ s}$; multi-hop stages remain a frozen-span limit, Table~\ref{tab:app_new_sota_breakdowns}c). On \texttt{LLaVA-1.5} \texttt{POPE}, zero-hook \texttt{PFG} matches \texttt{HulluEdit} ($86.52\%$ vs.\ $86.55\%$) at $2.19\times$ lower latency. On vision shifts, \texttt{L6--L11} $\mathcal{O}_3$ beats four 5-step backprop TTA baselines on \texttt{CIFAR-100-C} ($58.31\%$ vs.\ $57.63\%$), and recursive $\mathcal{O}_4$ with \texttt{RanPAC}~\cite{mcdonnell2023ranpac} wins $10/10$ \texttt{CIFAR-100} CIL stages ($87.62\%$).

On \emph{NLP calibration, reasoning control, and 32K frontier math} (Table~\ref{tab:sota_32k_reasoning}, Fig.~\ref{fig:mech_prop1}, Tables~\ref{tab:resource_and_c2}--\ref{tab:full_confirmatory}), $\mathcal{O}_4$ beats 4-shot ICL across $18{,}287$ \texttt{SST-2}/\texttt{AG News}/\texttt{MNLI} queries ($86.37\%$ vs.\ $85.55\%$). On \texttt{DeepSeek-R1-0528-8B}, math-calibrated \texttt{PFG} reveals a clear mechanistic distinction: \emph{representation separability does not guarantee intervention quality}. Unconstrained \texttt{Fisher LDA} attains $0.961$ AUROC on Math yet collapses zero-shot \texttt{HumanEval} to $25.61\%$ Pass@1, whereas \texttt{Protected FisherGate} achieves comparable $0.955$ AUROC while lifting \texttt{HumanEval} from $33.5\%$ to $\mathbf{84.15\%}$ ($92.7\%$ AST-valid) and cutting \texttt{GSM8K}/\texttt{MATH-500} tokens by $39$--$54\%$. On \texttt{DS-R1-Distill-Qwen-7B}, Layer-22 $\mathcal{O}_4$ matches \texttt{ReBalance} while unlocking native \texttt{vLLM} serving ($14.1\times$ faster) and exceeds \texttt{ThinkEdit} by $+16.7\text{ pp}$ on both \texttt{AIME 2024} ($60.0\%$ vs.\ $43.3\%$) and \texttt{AIME 2025} ($43.3\%$ vs.\ $26.7\%$). Finally, fixed $\mathcal{O}_3$ decay calibration ($+0.35$ \texttt{dt\_bias}) on \texttt{Qwen3.8-27B} improves all $5/5$ official $32\text{K}$ math Olympiads ($+2.59\text{ pp}$ macro \texttt{avg@4}, $5/5$ lower truncation).

\begin{table*}[t]
\centering
\caption{Complete fine-grained multi-teacher consolidation via zero-backprop all-layer $\mathcal{O}_4$ across ALL four official $32\text{K}$/$16\text{K}$ suites, listing every published RL, SFT, and weight-merging baseline alongside per-benchmark scores and GPU-hour compute cost (bold = best single model, underline = 2nd).}
\label{tab:fwsp_mopd}
\scriptsize
\setlength{\tabcolsep}{3.2pt}
\resizebox{\textwidth}{!}{%
\begin{tabular}{llccccccc}
\toprule
\multicolumn{9}{l}{\textbf{(a) \texttt{IM-MOPD} / \texttt{P$^2$-MOPD} Official 4-Domain $32\text{K}$ Suite (\texttt{Qwen3-4B-OT3}, $\text{avg@3}$, $T{=}1.0$; validation ablation in App.\ Table~\ref{tab:app_immopd_val})}} \\
\textbf{Method} & \textbf{Regime / Compute} & \textbf{MedQA} & \textbf{CaseHOLD} & \textbf{FinQA} & \textbf{IFBench} & \textbf{Avg.\ ($\uparrow$)} & \textbf{Avg.\ Rec.\ ($\uparrow$)} & \textbf{Worst Rec.\ ($\uparrow$)} \\
\midrule
\texttt{Base Student} ($\theta_0$) & Unedited SFT base & $68.84$ & $62.03$ & $58.76$ & $25.89$ & $53.88$ & $0.0\%$ & --- \\
4 Domain Teachers (Oracle) & 4 separate specialist models & $79.86$ & $73.57$ & $75.41$ & $59.22$ & $72.02$ & $100.0\%$ & --- \\
\midrule
\texttt{MOPD}~\cite{ma2026mopd} (Vanilla) & On-policy RL (backprop) & $62.90$ & $66.30$ & $75.04$ & $\mathbf{61.33}$ & $66.39$ & $46.80\%$ & $-53.94\%$ \\
\texttt{MOPD} (SFT warm-up) & SFT + On-policy RL & $69.76$ & $71.47$ & $74.22$ & $53.11$ & $67.14$ & $66.15\%$ & $8.32\%$ \\
\texttt{MOPD} (TA init) & TA merge + On-policy RL & $76.07$ & $\mathbf{72.62}$ & $\mathbf{75.35}$ & $51.08$ & $68.78$ & $83.13\%$ & $47.22\%$ \\
\texttt{Open-MOPD}~\cite{gao2026openmopd} & On-policy RL (backprop) & $75.07$ & $72.26$ & $73.73$ & \underline{$59.78$} & $70.21$ & $84.20\%$ & $56.55\%$ \\
\texttt{IM-MOPD}~\cite{kim2026immopd} & Iterative merge + RL & $78.22$ & $71.76$ & $73.25$ & $57.78$ & $70.25$ & $88.02\%$ & \underline{$84.12\%$} \\
\texttt{P$^2$-MOPD}~\cite{wang2026p2mopd} & Parameter prior + RL & \underline{$78.74$} & \underline{$72.58$} & \underline{$75.01$} & $59.41$ & $\mathbf{71.43}$ & $\mathbf{94.85\%}$ & $\mathbf{89.42\%}$ \\
\midrule
\texttt{TA Merge} & 0 backprop (task arith.) & $73.95$ & $66.00$ & $70.01$ & $33.78$ & $60.93$ & $42.98\%$ & $23.67\%$ \\
\texttt{RegMean}-style (Rand., $\alpha_k{=}1$) & 0 backprop (uncurated Gram) & $77.93$ & $69.43$ & $72.62$ & $48.11$ & $67.02$ & $74.13\%$ & $64.12\%$ \\
\textbf{FWSP Curated $\mathcal{O}_4$ (Ours)} & \textbf{0 backprop ($11\text{m}$ / 1 GPU)} & $\mathbf{79.52}$ & $70.98$ & $74.46$ & $56.22$ & \underline{$\mathbf{70.30}$} & \underline{$\mathbf{89.94\%}$} & $77.56\%$ \\
\bottomrule
\end{tabular}}

\vspace{4pt}
\resizebox{\textwidth}{!}{%
\begin{tabular}{llccc|ccc|ccc|cc}
\toprule
\multicolumn{13}{l}{\textbf{(b) \texttt{Open-MOPD 3B} Official 6-Benchmark $32\text{K}$ Suite (\texttt{SmolLM3-3B}: Math $\text{avg@64}$, Code $\text{avg@10}$, IF $\text{avg@1}$; \citet{gao2026openmopd} Table~2)}} \\
\cmidrule(lr){3-5}\cmidrule(lr){6-8}\cmidrule(lr){9-11}\cmidrule(lr){12-13}
& & \multicolumn{3}{c|}{\textbf{Math ($\text{avg@64}$)}} & \multicolumn{3}{c|}{\textbf{Code ($\text{avg@10}$)}} & \multicolumn{3}{c|}{\textbf{IF ($\text{avg@1}$)}} & \multicolumn{2}{c}{\textbf{Overall}} \\
\textbf{Method} & \textbf{Regime / GPU-h} & \textbf{A24} & \textbf{A25} & \textbf{Avg} & \textbf{LCBv5} & \textbf{LCBv6} & \textbf{Avg} & \textbf{IFEv} & \textbf{IFB} & \textbf{Avg} & \textbf{Macro ($\uparrow$)} & \textbf{Rec.\ ($\uparrow$)} \\
\midrule
\texttt{SmolLM3-3B-Base} & Pretrained base & $2.08$ & $1.72$ & $1.90$ & $3.41$ & $6.17$ & $4.79$ & $16.08$ & $13.00$ & $14.54$ & $7.08$ & --- \\
$\pi_{\text{mixsft}}$ (e4 Init) & Mixed SFT init & $15.63$ & $20.26$ & $17.95$ & $15.99$ & $19.20$ & $17.60$ & $66.91$ & $16.00$ & $41.46$ & $25.67$ & $0.0\%$ \\
\texttt{RouteRL} (3 Teachers Oracle) & 3 separate RL models & $23.65$ & $24.84$ & $24.24$ & $22.16$ & $21.31$ & $21.73$ & $74.49$ & $27.67$ & $51.08$ & $32.35$ & $100.0\%$ \\
\midrule
\texttt{RFT} & Rejection fine-tuning & $22.97$ & $23.91$ & $23.44$ & $18.98$ & $19.43$ & $19.21$ & $55.08$ & $18.67$ & $36.87$ & $26.51$ & $12.6\%$ \\
$\pi_{\text{mixrl}}$ (Mix RL) & Multi-domain RL & $21.15$ & $22.14$ & $21.64$ & $16.59$ & $20.97$ & $18.78$ & $70.24$ & $22.67$ & $46.45$ & $28.96$ & $49.3\%$ \\
\texttt{Naive M-OPD} & On-policy RL ($\sim 187\text{h}$) & $20.92$ & $21.60$ & $21.26$ & $17.53$ & $20.99$ & $19.26$ & $68.61$ & $18.67$ & $43.64$ & $28.05$ & $35.6\%$ \\
\texttt{Open-MOPD}~\cite{gao2026openmopd} & On-policy RL ($\sim 187\text{h}$) & $21.98$ & $22.86$ & $22.42$ & $20.84$ & $22.63$ & $21.73$ & $\mathbf{74.49}$ & $\mathbf{24.67}$ & $\mathbf{49.58}$ & \underline{$31.24$} & \underline{$83.4\%$} \\
\midrule
\texttt{ParamMerge-Avg} & 0 backprop (avg) & $18.91$ & $20.99$ & $19.95$ & $18.38$ & $21.20$ & $19.79$ & $70.06$ & $17.67$ & $43.86$ & $27.87$ & $32.9\%$ \\
\texttt{ParamMerge-TA} & 0 backprop (TA) & $21.93$ & $22.76$ & $22.34$ & $21.74$ & $23.14$ & $22.44$ & $71.53$ & $21.67$ & $46.60$ & $30.46$ & $71.7\%$ \\
FWSP Curated $\mathcal{O}_4$ ($\alpha_{\text{if}}{=}1.5$) & 0 backprop ($\sim 0.15\text{h}$, 1 GPU) & $\mathbf{23.33}$ & \underline{$24.90$} & $\mathbf{24.11}$ & \underline{$21.86$} & $\mathbf{24.11}$ & \underline{$22.99$} & $70.43$ & $20.33$ & $45.38$ & $30.83$ & $77.2\%$ \\
\textbf{FWSP Curated $\mathcal{O}_4$ ($\alpha_k{=}1$, Ours)} & \textbf{0 backprop ($\sim 0.15\text{h}$, 1 GPU)} & \underline{$23.02$} & $\mathbf{25.21}$ & $\mathbf{24.11}$ & $\mathbf{22.04}$ & \underline{$24.06$} & $\mathbf{23.05}$ & \underline{$71.90$} & \underline{$22.67$} & \underline{$47.29$} & $\mathbf{31.48}$ & $\mathbf{87.0\%}$ \\
\bottomrule
\end{tabular}}

\vspace{4pt}
\resizebox{\textwidth}{!}{%
\begin{tabular}{llccccc}
\toprule
\multicolumn{7}{l}{\textbf{(c) \texttt{TrustMOPD 7B} Official 6-Benchmark $32\text{K}$ Suite (\texttt{DS-R1-Distill-Qwen-7B}: AIME25/26, LCBv5/v6, IFEval/IFBench; \citet{sun2026trustmopd} Table~11--12)}} \\
\textbf{Method} & \textbf{Regime / Reported GPU-h} & \textbf{Math (A25/A26)} & \textbf{Code (LCBv5/v6)} & \textbf{IF (IFEv/IFB)} & \textbf{Overall ($\uparrow$)} & \textbf{$\Delta$ vs.\ Init} \\
\midrule
Student Init (\texttt{DS-R1-Qwen-7B}) & Shared pre-RL base & $43.02$ & $26.99$ & $28.13$ & $32.71$ & --- \\
\texttt{Uniform RL} (SingleCap) & On-policy RL ($179\text{ GPU-h}$) & $49.67$ & $30.00$ & \underline{$32.83$} & $37.50$ & $+4.79\text{ pp}$ \\
\texttt{Uniform RL} (MultiCap) & On-policy RL ($179\text{ GPU-h}$) & $50.55$ & $30.37$ & $31.68$ & $37.53$ & $+4.82\text{ pp}$ \\
\texttt{TrustMOPD}~\cite{sun2026trustmopd} (MultiCap) & On-policy RL ($192\text{ GPU-h}$) & $52.68$ & $31.42$ & $31.93$ & $38.68$ & $+5.97\text{ pp}$ \\
\texttt{TrustMOPD}~\cite{sun2026trustmopd} (SingleCap) & On-policy RL ($192\text{ GPU-h}$) & $52.87$ & $30.91$ & $\mathbf{32.91}$ & $38.90$ & $+6.19\text{ pp}$ \\
\midrule
FWSP Curated $\mathcal{O}_4$ ($\alpha_k{=}1$) & \textbf{0 backprop ($<0.25\text{h}$, 1 GPU)} & \underline{$54.58$} ($51.7/57.5$) & $\mathbf{43.46}$ ($46.2/40.7$) & $24.69$ ($32.7/16.7$) & \underline{$40.91$} & $+8.20\text{ pp}$ \\
\textbf{FWSP Curated $\mathcal{O}_4$ ($\alpha_{\text{if}}{=}1.5$, Ours)} & \textbf{0 backprop ($<0.25\text{h}$, 1 GPU)} & $\mathbf{55.94}$ ($51.9/60.0$) & \underline{$42.55$} ($44.1/41.0$) & $24.99$ ($33.6/16.3$) & $\mathbf{41.16}$ & $\mathbf{+8.45\text{ pp}}$ ($\mathbf{+2.26}$ vs.\ RL) \\
\bottomrule
\end{tabular}}

\vspace{4pt}
\resizebox{\textwidth}{!}{%
\begin{tabular}{llccccccc}
\toprule
\multicolumn{9}{l}{\textbf{(d) \texttt{LLM-Fusion 4B} Official 5-Benchmark $\text{mean@32}$ Suite (\texttt{Qwen3-4B-Instruct-2507}, $16\text{K}$; \citet{wu2026llmfusion} Tables~1--2)}} \\
\textbf{Method} & \textbf{Regime / Reported Fusion GPU-h} & \textbf{AIME 25} & \textbf{AIME 26} & \textbf{GPQA-D} & \textbf{IFEval} & \textbf{IFBench} & \textbf{5-Bench Avg.\ ($\uparrow$)} & \textbf{$\Delta$ vs.\ Base} \\
\midrule
\texttt{Base} (Paper Table~1 / Local) & Unedited base & $49.1$ / $48.65$ & $53.1$ / $54.06$ & $57.2$ / $58.60$ & $83.8$ / $83.70$ & $30.3$ / $30.50$ & $54.70$ / $55.10$ & --- \\
Per-domain RL (Full, Oracle) & 5 separate RL experts & $52.3$ & $58.4$ & $62.2$ & $89.8$ & $49.0$ & $62.34$ & $+7.64\text{ pp}$ \\
\midrule
\texttt{Mix RL} (Full, \citet{wu2026llmfusion}) & Multi-dom RL ($3{,}042\text{ GPU-h}$) & $53.0$ & $59.0$ & $61.1$ & $85.9$ & $39.8$ & $59.76$ & $+5.06\text{ pp}$ \\
\texttt{MOPD} (Full, \citet{wu2026llmfusion}) & On-policy RL ($741\text{ GPU-h}$) & $51.2$ & $57.5$ & $62.3$ & $89.2$ & $47.3$ & $61.50$ & $+6.80\text{ pp}$ \\
\texttt{Merge} (LoRA / Full, Paper) & 0 backprop ($\approx 0\text{h}$) & $54.3$ / $54.9$ & $59.6$ / $\mathbf{61.6}$ & $61.9$ / $63.4$ & $89.0$ / $88.3$ & $41.9$ / $41.0$ & $61.34$ / $61.84$ & $+7.14\text{ pp}$ \\
\midrule
\texttt{RegMean} (Local same-eval) & 0 backprop (uncurated) & $48.65$ & $56.25$ & $61.77$ & $87.87$ & $40.64$ & $59.04$ & $+3.94\text{ pp}$ \\
\texttt{TA Task-Sum} (Local same-eval) & 0 backprop (task sum) & $\mathbf{55.42}$ & $60.10$ & \underline{$63.83$} & $\mathbf{89.71}$ & $47.46$ & \underline{$63.30$} & $+8.20\text{ pp}$ \\
FWSP Curated $\mathcal{O}_4$ ($\gamma{=}1.0$) & 0 backprop ($\sim 0.2\text{h}$, 1 GPU) & $49.38$ & $57.19$ & $62.55$ & $88.55$ & $41.49$ & $59.83$ & $+4.73\text{ pp}$ \\
\textbf{FWSP Curated $\mathcal{O}_4$ ($\gamma{=}2.5$, Ours)} & \textbf{0 backprop ($\sim 0.2\text{h}$, 1 GPU)} & \underline{$55.00$} & $61.04$ & $63.60$ & $88.51$ & \underline{$51.95$} & $\mathbf{64.02}$ & $\mathbf{+8.92\text{ pp}}$ ($\mathbf{+2.52}$ vs.\ \texttt{MOPD}) \\
FWSP Curated $\mathcal{O}_4$ ($\gamma{=}3.5$) & 0 backprop ($\sim 0.2\text{h}$, 1 GPU) & $54.48$ & \underline{$61.35$} & $\mathbf{64.50}$ & $83.50$ & $\mathbf{52.48}$ & $63.26$ & $+8.16\text{ pp}$ \\
\bottomrule
\end{tabular}}
\vspace{2pt}
\par\noindent{\footnotesize Note: Panel~(a) quotes \citet{wang2026p2mopd} Table~1 (FWSP solve/export takes $99.5\text{s}$, $\sim 11\text{min}$ end-to-end on 1 GPU); Panel~(b) quotes \citet{gao2026openmopd} Table~2 (FWSP beats both \texttt{Open-MOPD} and the single-domain Code RL oracle teacher $21.73$); Panel~(c) quotes \citet{sun2026trustmopd} Tables~11--12 ($8{\times}\text{H20}$ GPU-h); Panel~(d) quotes \citet{wu2026llmfusion} Tables~1--2 over the 5 shared benchmarks evaluated locally under exact $\text{mean@32}$ ($T{=}0.6, \text{top-}p{=}0.95, 16\text{K}$). Reported GPU-hours come from each upstream paper's hardware cluster vs.\ our single 98\,GB GPU and indicate order-of-magnitude compute savings rather than strict same-hardware wall-clock speedups.}
\end{table*}

\subsection{Large-Scale Instantiation: Zero-Backprop Multi-Teacher Consolidation}
\label{sec:fwsp_mopd}
Scaling $\mathcal{O}_4$ to all linear projections (Eq.~\eqref{eq:fwsp_mopd}) consolidates an SFT base $W_0$ with specialist RL teachers in $<0.25\text{ GPU-h}$ compared with $179$--$3{,}042\text{ GPU-h}$ of on-policy RL distillation (Table~\ref{tab:fwsp_mopd}). A 6-row ablation on the official $1{,}024$-query $32\text{K}$ \texttt{IM-MOPD} validation pool (Appendix Table~\ref{tab:app_immopd_val}) separates each component's contribution: \textbf{input-covariance merging (\texttt{RegMean}) drives the primary $+13.97\text{ pp}$ gain over \texttt{Task Arithmetic} ($63.28\% \to 77.25\%$)}, while trajectory curation ($+0.09\text{ pp}$ overall, $+4.68\text{ pp}$ IF), response-token weighting, and base-centered regularization ($\lambda\|\Delta W\|_F^2$, $+4.30\text{ pp}$ IF) add $+0.68\text{ pp}$ at fixed $\alpha_k{=}1$ ($77.93\%$) and $+1.56\text{ pp}$ at $\alpha_{\text{if}}{=}1.5$ ($78.81\%$ val; $70.30\%$ official $32\text{K}$ test, surpassing \texttt{IM-MOPD} RL $70.25\%$ and trailing only \texttt{P$^2$-MOPD} $71.43\%$). Across the three remaining suites (\texttt{Open-MOPD 3B} at $31.48\%$, \texttt{TrustMOPD 7B} at $41.16\%$, where Math/Code gains trade off against IF mergeability, and \texttt{LLM-Fusion 4B} at $64.02\%$), all-layer $\mathcal{O}_4$ ranks \#1 among all listed single models on $3/4$ suites with zero backpropagation.

\begin{figure*}[t]
    \centering
    \includegraphics[width=0.86\textwidth]{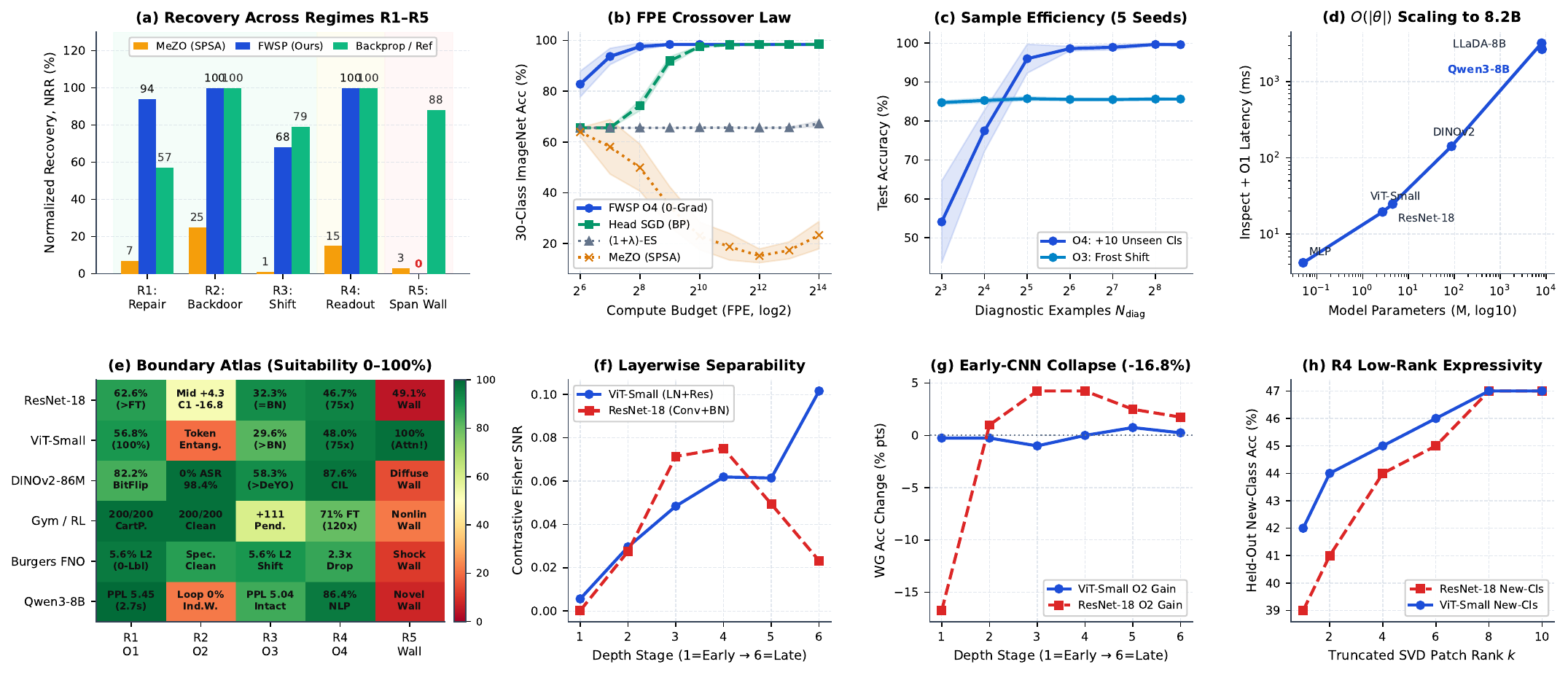}
    \caption{Capability boundaries and scaling laws: (a,e) recovery across R1--R5 and six architectures; (b--d) FPE crossover ($B^*{\approx}4{,}096$), sample saturation ($16$--$64$), and $O(|\theta|)$ scaling for $\mathcal{O}_1$ weight-scan repair; (f--h) depth and SVD rank profiles.}
    \label{fig:boundaries}
\end{figure*}

\begin{table*}[t]
\centering
\caption{R5 boundary evaluation: forward-only $\mathcal{O}_4$ vs.\ \texttt{MeZO}/\texttt{ROME} and backpropagation.}
\label{tab:r5_wall}
\resizebox{\textwidth}{!}{%
\begin{tabular}{llcccc}
\toprule
\textbf{Hard R5 Benchmark} & \textbf{Model} & \textbf{MeZO / ROME-Style} & \textbf{FWSP $\mathcal{O}_4$} & \textbf{FWSP $\mathcal{O}_4^{\text{deep}}$} & \textbf{Backprop (FT / LoRA)} \\
\midrule
3-Patch Spatial XOR (3 seeds) & \texttt{ResNet-18} (Local Conv) & $52.99 \pm 7.4\%$ (\texttt{MeZO}) & $49.09 \pm 1.9\%$ & $49.67 \pm 10.6\%$ & $\mathbf{100.0 \pm 0.0\%}$ \\
3-Patch Spatial XOR (3 seeds) & \texttt{ViT-Small} (Global Attn) & $49.67 \pm 4.3\%$ (\texttt{MeZO}) & $\mathbf{100.0 \pm 0.0\%}$ & $\mathbf{100.0 \pm 0.0\%}$ & $\mathbf{100.0 \pm 0.0\%}$ \\
Diffuse All-Layer Noise ($\sigma{=}0.008$) & \texttt{DINOv2-Base} (86.6M) & --- & $48.17\%$ & $48.00\%$ & $\mathbf{70.33\%}$ (\texttt{LoRA}) \\
\texttt{CounterFact} 32 Facts (Eff. / Spec. / PPL) & \texttt{Qwen3-8B} (8.2B) & $96.9\%$ / $65.6\%$ / $5.53$ (\texttt{ROME}) & $87.5\%$ / $\mathbf{90.6\%}$ / $\mathbf{5.04}$ & --- & $\mathbf{100\%}$ / $53.1\%$ / $58.8$ \\
16 Novel Entities (Structural Paraphrase) & \texttt{Qwen3-8B} (8.2B) & $0.0\%$ (\texttt{MeZO}) & $0.0\%$ & --- & $\mathbf{43.8\%}$ (\texttt{LoRA}, $100\%$ tr.) \\
\bottomrule
\end{tabular}}
\vspace{2pt}
\par\noindent{\footnotesize Note: ``---'' denotes inapplicable/omitted settings ($\mathcal{O}_4^{\text{deep}}$ targets fixed-width layers rather than the 152K-vocab \texttt{lm\_head}; isotropic \texttt{MeZO} omitted under diffuse \texttt{DINOv2} noise).}
\end{table*}

\subsection{Capability Boundaries, Forward Scope Prediction, and Hybrid FWSP + LoRA (RQ4 \& RQ5)}
\label{sec:exp_boundaries}
Figure~\ref{fig:boundaries} and Table~\ref{tab:r5_wall} map FWSP's compute and geometric boundaries. Empirically, $\mathcal{O}_4$ saturates by $N_{\text{diag}}{=}16$--$64$ and crosses iterative SGD near $B^* \approx 4{,}096$ FPE, while $\mathcal{O}_1$ weight-scan localization scales linearly as $O(|\theta|)$ (whereas $\mathcal{O}_4$ solves scale with layer feature dimension $O(d_{\text{in}}^3 + d_{\text{out}} d_{\text{in}}^2)$ per edited layer). Geometrically, Proposition~1's risk proxy flags early-layer entanglement ($\rho{=}0.498$, Fig.~\ref{fig:mech_prop1}c), while Proposition~2 explains why frozen-upstream $\mathcal{O}_4$ fails on local-convolution \texttt{ResNet-18} 3-patch XOR ($49.1\%$) yet solves global-attention \texttt{ViT-Small} ($100\%$), and why diffuse all-layer noise and novel-entity paraphrases require multi-layer updates or gradient credit assignment (\texttt{LoRA}/\texttt{Full-FT}). When tasks cross these walls, warm-starting a single \texttt{LoRA} adapter on the FWSP-localized block lifts damaged+shifted \texttt{DINOv2} from $8.57\%$ (blind \texttt{LoRA}) to $88.10\%$, yielding $13.7\times$ lower standard deviation than random single-layer \texttt{LoRA} ($0.52$ vs.\ $7.13$) and using $10.7\times$ fewer trainable parameters than 12-layer \texttt{LoRA} (Table~\ref{tab:ablations_hybrid}).

\paragraph{Grid-Verified Decoupled Two-Axis Minimal-Scope Prediction and $\rho_{\text{ent}}$ Gate.}
Because optimization method (\emph{Analytic} vs.\ \emph{Gradient}) and parameter scope (\emph{1L}, \emph{ML}, \emph{Full}) are orthogonal (for example, localized 1-layer \texttt{LoRA} is both Gradient and 1L), we define operational minimal-scope ground truth via \emph{lexicographic optimization} with an explicit empty-feasible-set escalation over a finite five-cell candidate grid $\mathcal{G}$ (\texttt{Analytic-1L}, \texttt{Analytic-ML}, \texttt{Analytic-Full}, \texttt{Grad-1L-LoRA}, \texttt{Grad-ML-LoRA}):
\begin{equation}
\begin{aligned}
    (M^*, S^*) &=
    \begin{cases}
        \displaystyle\operatorname*{arg\,min}^{\mathrm{lex}}_{(M, S) \in \mathcal{G}_{\mathrm{perf}}} \big(c(M), s(S)\big), & \mathcal{G}_{\mathrm{perf}} \neq \emptyset, \\
        (\text{Gradient}, \text{ML}), & \mathcal{G}_{\mathrm{perf}} = \emptyset,
    \end{cases} \\
    \mathcal{G}_{\mathrm{perf}} &= \big\{(M, S) \in \mathcal{G} : \mathrm{Perf}(M, S) \ge \tau_{\text{perf}}\big\},
\end{aligned}
    \label{eq:scope_gt}
\end{equation}
where $c(\text{Analytic}) < c(\text{Gradient})$ and $s(\text{1L}) < s(\text{ML}) < s(\text{Full})$. Note two explicit protocol boundaries: $\mathcal{G}$ evaluates five configurations (omitting full-backbone backprop FT), so $(M^*, S^*)$ is optimal within $\mathcal{G}$ rather than globally; and our $24$-task grid uses the simplified performance-only feasible set $\mathcal{G}_{\mathrm{perf}}$ with $\tau_{\text{perf}} = \max(70\%, P_{\max}{-}3\text{ pp})$, where the \texttt{(Gradient, ML)} fallback for $\mathcal{G}_{\mathrm{perf}} = \emptyset$ ($P_{\max} < 70\%$) is an escalation policy rather than a guarantee that \texttt{Grad-ML-LoRA} reaches $\tau_{\text{perf}}$, while retention gating is evaluated separately on Axis~3 ($\rho_{\text{ent}}$) rather than inside Eq.~\eqref{eq:scope_gt}. We execute all five cells on GPU across $24$ real tasks and three architectures (\texttt{SigLIP 2 SO400M}, \texttt{DINOv2-Base}, \texttt{Qwen3-4B-Instruct-2507}; Table~\ref{tab:scope_predictor}a).

Using only pre-edit \texttt{torch.no\_grad()} statistics, the three decoupled axes exhibit distinct empirical strengths. For \emph{solver selection (Axis~1)}, the frozen-span residual $\rho_{\text{span}} = \|(I{-}\Pi_\Phi)Y_c\|_F / \|Y_c\|_F$ ($\tau_{\text{span}}{=}0.70$) achieves $\mathbf{91.7\%}$ ($22/24$) fixed-rule accuracy and $\mathbf{87.5\%}$ ($21/24$) under both Leave-One-Task-Out (LOTO) and Leave-One-Architecture-Out (LOAO) validation, beating the metadata-only baseline ($62.5\%$) by $+25.0$--$29.2\text{ pp}$ (its only two fixed-rule misses are borderline: \texttt{SigLIP 2} \texttt{Frost-5} at $86.0\%$ vs.\ $88.5\%$ and \texttt{Qwen3-4B} prefix shift at $92.5\%$ vs.\ $95.8\%$). For \emph{scope selection (Axis~2)}, full-grid evaluation shows that moderate-span shifts with high layer spread (\texttt{Contrast-5}, \texttt{Defocus-5}, $\rho_{\text{spread}} > 0.91$) are solved by \texttt{Analytic-1L} ($92.5$--$96.5\%$), so naive $\rho_{\text{spread}} > 0.45$ over-predicts multi-layer scope ($54.2\%$). Coupling span and spread ($\rho_{\text{span}} > 0.70 \land \rho_{\text{spread}} > 0.45$) restores scope accuracy to $75.0\%$ fixed ($79.2\%$ LOTO/LOAO) and separates \texttt{(Gradient, 1L)} (\texttt{DINOv2} \texttt{Frost-5}) from \texttt{(Gradient, ML)} (\texttt{SigLIP 2} diffuse noise, $+45\text{ pp}$ gap), but does not substantially exceed the constant single-layer baseline ($75.0\%$). Consequently, \emph{joint (Solver, Scope) accuracy} reaches $17/24$ ($70.8\%$) fixed and $16/24$ ($66.7\%$) LOTO/LOAO vs.\ $15/24$ ($62.5\%$) metadata-only, confirming that forward gains are driven primarily by Axis~1 solver selection rather than Axis~2 scope selection. For \emph{Stage-2 protection (Axis~3, Table~\ref{tab:scope_predictor}b)}, given a Euclidean-orthonormal edit basis $U \in \mathbb{R}^{D \times r}$ satisfying $U^\top U = I_r$ (such as $\bar{U}_A = \mathrm{QR}(U_A^*)$ or unit vector $\hat{d}_{\text{PFG}}^*$, rather than an unnormalized $A$-orthonormal basis $U^\top A U = I_r$), the rank-normalized entanglement proxy $\rho_{\text{ent}}(U) = \mathrm{tr}(U^\top \Sigma_{\text{keep}} U) / \sum_{j=1}^r \lambda_j(\Sigma_{\text{keep}}) \in [0, 1]$ (defined as $0$ whenever $\sum_{j=1}^r \lambda_j(\Sigma_{\text{keep}}) = 0$, and bounded in $[0, 1]$ by Ky Fan's eigenvalue inequality under $U^\top U = I_r$) correlates with unprotected retention loss across $16$ editing conditions (Spearman $\rho{=}0.628, p{=}0.0092$) and reaches $\mathbf{87.5\%}$ ($14/16$) Fixed-Rule ($\tau_{\text{ent}}{=}0.35$), LOTO, and LOAO accuracy, saving $\mathbf{+8.00\text{ pp}}$ mean retention (up to $+44.17\text{ pp}$ on \texttt{Qwen3-4B}) when triggered with $-0.29\text{ pp}$ (near-zero) drift when skipped. Table~\ref{tab:scope_predictor}(c) also preserves our $17$-task five-architecture evaluation including full-backbone \texttt{IM-MOPD} ($88.2\%$ LOTO / $82.4\%$ LOAO vs.\ $52.9\%$ metadata-only).

\begin{table*}[t]
\centering
\caption{Forward-only optimization routing across three decoupled axes: (a) $24$-task, $5$-cell GPU grid ground truth (Eq.~\eqref{eq:scope_gt}); (b) independent Axis-3 $\rho_{\text{ent}}$ protection gate ($16$ conditions, $\tau_{\text{ent}}{=}0.35$); and (c) $17$-task empirical-regime summary across five architectures.}
\label{tab:scope_predictor}
\scriptsize
\setlength{\tabcolsep}{3.1pt}
\resizebox{\textwidth}{!}{%
\begin{tabular}{llccccccl}
\toprule
\multicolumn{9}{l}{\textbf{(a) Grid-Verified Decoupled Two-Axis Predictor ($24$ Real Tasks $\times$ 5 GPU Grid Cells: \texttt{SigLIP 2}, \texttt{DINOv2-Base}, \texttt{Qwen3-4B})}} \\
\textbf{Decoupled Axis / Decision} & \textbf{Forward Statistic \& Rule} & \textbf{Grid GT Counts} & \textbf{Meta Only} & \textbf{Fixed Rule} & \textbf{LOTO CV} & \textbf{LOAO CV} & \multicolumn{2}{l}{\textbf{Key Empirical Finding from Full 5-Cell GPU Grid}} \\
\midrule
Axis 1: Solver (\texttt{Analytic} vs.\ \texttt{Grad.}) & $\rho_{\text{span}} > 0.70 \Rightarrow$ \texttt{Gradient} & $15$ An.\ / $9$ Gr. & $15/24$ ($62.5\%$) & $\mathbf{22/24}$ ($\mathbf{91.7\%}$) & $\mathbf{21/24}$ ($\mathbf{87.5\%}$) & $\mathbf{21/24}$ ($\mathbf{87.5\%}$) & \multicolumn{2}{l}{$+25.0$--$29.2\text{ pp}$ over metadata; 2 borderline misses} \\
Axis 2: Scope (Naive Spread) & $\rho_{\text{spread}} > 0.45 \Rightarrow$ \texttt{ML} & $18$ 1L / $6$ ML & $18/24$ ($75.0\%$) & $13/24$ ($54.2\%$) & $18/24$ ($75.0\%$) & $18/24$ ($75.0\%$) & \multicolumn{2}{l}{Moderate-span shifts solved by \texttt{A-1L} ($92.5$--$96.5\%$)} \\
Axis 2: Scope (Coupled Span+Spread) & $\rho_{\text{span}}{>}0.70 \wedge \rho_{\text{spread}}{>}0.45$ & $18$ 1L / $6$ ML & $18/24$ ($75.0\%$) & $18/24$ ($75.0\%$) & $19/24$ ($79.2\%$) & $19/24$ ($79.2\%$) & \multicolumn{2}{l}{Matches constant 1L ($+1$ task in CV); separates \texttt{Grad 1L/ML}} \\
\textbf{Joint (Solver, Scope) (Coupled)} & \textbf{Axis 1 + Coupled Axis 2} & \textbf{5-cell grid $\mathcal{G}$} & $15/24$ ($62.5\%$) & $\mathbf{17/24}$ ($\mathbf{70.8\%}$) & $16/24$ ($66.7\%$) & $16/24$ ($66.7\%$) & \multicolumn{2}{l}{Naive spread joint: $13/24$ ($54.2\%$); gain driven by Axis 1} \\
\midrule
\multicolumn{9}{l}{\textbf{(b) Independent Axis-3 Validation: Scale- \& Rank-Invariant $\rho_{\text{ent}}$ Stage-2 Protection Gate ($16$ Conditions across 3 Architectures)}} \\
\textbf{Gate Decision ($\tau_{\text{ent}}{=}0.35$)} & \textbf{Forward Statistic} & \textbf{Spearman $\rho$ ($p$)} & \textbf{Fixed Rule} & \textbf{LOTO CV} & \textbf{LOAO CV} & \textbf{High-$\rho_{\text{ent}}$ Saved} & \multicolumn{2}{l}{\textbf{Low-$\rho_{\text{ent}}$ Unprot.\ Drop / Mechanism}} \\
\midrule
Activate $Q_{\text{keep}}$ ($\Delta\text{Ret}_{\text{unprot}} \ge 2\text{ pp}$) & $\frac{\mathrm{tr}(U^\top \Sigma_{\text{keep}} U)}{\sum_{j=1}^r \lambda_j(\Sigma_{\text{keep}})} > 0.35$ ($U^\top U{=}I_r$) & $\mathbf{0.628}$ ($p{=}.009$) & $\mathbf{14/16}$ ($\mathbf{87.5\%}$) & $\mathbf{14/16}$ ($\mathbf{87.5\%}$) & $\mathbf{14/16}$ ($\mathbf{87.5\%}$) & $\mathbf{+8.00\text{ pp}}$ & \multicolumn{2}{l}{$-0.29\text{ pp}$ (zero drift); up to $+44.2\text{ pp}$ saved on \texttt{Qwen3}} \\
\midrule
\multicolumn{9}{l}{\textbf{(c) Empirical-Regime Scope Labels ($17$ Real 2026 Tasks across 5 Architectures incl.\ \texttt{IM-MOPD} Full-Backbone, \texttt{Z-Image}, \texttt{Ovis})}} \\
\textbf{Empirical Regime Scope} & \textbf{Measured 2026 Tasks ($N{=}17$)} & \textbf{Mean $\rho_{\text{span}}$} & \textbf{Mean $\rho_{\text{ent}}$} & \textbf{Mean $\rho_{\text{spread}}$} & \textbf{Meta Only} & \textbf{Fixed / LOTO} & \textbf{LOAO CV} & \textbf{Error / Mechanism Note} \\
\midrule
Single-Layer / Multi-Layer & \texttt{SigLIP2}, \texttt{Qwen3}, \texttt{Z-Image} ($5$ 1L, $4$ ML) & $0.409$ / $0.754$ & $0.092$ / $0.115$ & $0.044$ / $0.959$ & $5/9$ ($55.6\%$) & $7/9$ ($77.8\%$) & $6/9$ ($66.7\%$) & 2 conservative escalations on Sev-5 \\
Full-Backbone / Grad.\ (R5) & \texttt{IM-MOPD} ($4$ Full), R5 Wall ($4$ Grad.) & $0.007$ / $1.000$ & $0.014$ / $0.410$ & $0.998$ / $0.925$ & $4/8$ ($50.0\%$) & $\mathbf{8/8}$ ($100\%$) & $\mathbf{8/8}$ ($100\%$) & Uniform $\Delta W$ vs.\ saturated residual \\
\textbf{Overall (17 Empirical Labels)} & \textbf{5 Architectures (incl.\ Multi-Teacher)} & --- & --- & --- & $9/17$ ($52.9\%$) & $\mathbf{15/17}$ ($\mathbf{88.2\%}$) & $\mathbf{14/17}$ ($\mathbf{82.4\%}$) & $+35.3\text{ pp}$ over metadata baseline \\
\bottomrule
\end{tabular}}
\vspace{2pt}
\par\noindent{\footnotesize Note: Panel~(a) evaluates all five candidate cells (\texttt{Analytic-1L/ML/Full}, \texttt{Grad-1L/ML-LoRA}) on GPU across three architectures and reports Solver, Scope, and Joint (Solver, Scope) accuracy separately. Panel~(b) validates the independent $\rho_{\text{ent}}$ protection gate across three architectures. Panel~(c) summarizes empirical-regime scope labels across five architectures including full-backbone \texttt{IM-MOPD}.}
\end{table*}

\section{Conclusion}
\label{sec:conclusion}
Do we really need an LLM for Weight-Space Programming? Our study across seven modalities and $33$ base models yields a structured answer (\emph{LLMs for planning, linear algebra for compilation}): high-level task planning may benefit from an LLM, whereas low-level numerical weight compilation does not. Formalizing the compilation layer analytically as FWSP avoids multi-turn trial-and-error and achieves zero zero-score failures across the $14$ evaluated \texttt{OpenWSP} behavioral-edit tasks, outperforming published 2026 frontier agents standalone on matched subsets of \texttt{OpenWSP} public \texttt{Edit} and \texttt{WeightBench} \texttt{TR02-LOOP}. Equally important, our controlled $3$-arm evaluation across all $15$ \texttt{OpenWSP} tasks ($90$ episodes) separates generic tool availability from FWSP's full analytic compiler design, showing that exposing FWSP compilation primitives to live frontier agents yields large, statistically significant accuracy and latency gains over both scripting from scratch and standard Ridge/SVD tools.

Beyond external WSP benchmarks, the same closed-form machinery (protected subspace discovery paired with $\mathcal{O}_1$--$\mathcal{O}_4$) generalizes cleanly across concept erasure, multimodal editing, vision TTA, long-horizon math reasoning, and full-backbone multi-teacher consolidation with zero backpropagation and zero runtime hooks. Together with our geometric entanglement and frozen-span bounds, grid-verified forward predictors show how pre-edit activation statistics separate analytic compilation from localized gradient handoff (\texttt{LoRA}) and trigger retention protection only when needed.

\paragraph{Limitations and Future Work.}
Four scope boundaries remain. On open-ended tasks requiring algorithm synthesis from natural language (\texttt{WeightBench} \texttt{CTRL}/\texttt{WRITE}), example-free skill injection (\texttt{OpenWSP Skill}), or iterative sparse subnetwork search (\texttt{W2}), analytic compilation alone has not yet solved these regimes, where LLM planning currently remains advantageous; while our $3$-arm \texttt{OpenWSP} evaluation validates planning--compilation complementarity on behavioral editing, extending that hybrid stack to \texttt{OpenWSP Skill} and \texttt{WeightBench} \texttt{CTRL}/\texttt{WRITE} is an immediate next step. On \texttt{OpenWSP}, v8 is a public practice set whose $500$-task private holdout and fine-tuned organism checkpoints (\texttt{W1}/\texttt{Read}) are unreleased, and our standalone solver uses a fixed benchmark-family adapter rather than a pre-benchmark zero-shot algorithm, leaving private-set evaluation as future work. Within FWSP's own operators, \texttt{Protected FisherGate} assumes single-token control tags and requires Stage-1/Stage-4 gating on degenerate or already-concise regimes, while multi-teacher $\mathcal{O}_4$ assumes a shared architecture and base $W_0$. Finally, while Axis~1 solver selection and Axis~3 protection gating generalize strongly across architectures, Axis~2 single- vs.\ multi-layer scope prediction under labeled readout $\mathcal{O}_4$ remains close to the single-layer baseline and merits finer multi-layer diagnostics.

\FloatBarrier

\section*{AI Use Statement}
Large language models were used as assistive tools for coding boilerplate, formatting scripts, and copy-editing prose during manuscript preparation. All experimental designs, theoretical proofs, empirical evaluations, and claims were verified by the authors, who take full responsibility for the entire paper.

\section*{Impact Statement}
This paper advances forward-only weight-space optimization, enabling fast, low-memory model repair, concept erasure, hallucination reduction, and multi-teacher consolidation without backpropagation. Because direct weight editing is dual-use (capable of both strengthening safety guardrails and auditing refusal-direction vulnerabilities), all safety experiments in this study are strictly restricted to controlled academic benchmarks (\texttt{AdvBench}, \texttt{XSTest}, \texttt{I2P}), and Stage-4 retention verification provides an empirical safeguard only for the specific validation metrics evaluated on $\mathcal{D}_{\text{val}}$ rather than certifying global safety or general-capability invariance.

\bibliography{references}
\bibliographystyle{\BibStyle}

\newpage
\appendix
\AppendixLayout
\setcounter{table}{0}
\renewcommand{\thetable}{S\arabic{table}}
\setcounter{figure}{0}
\renewcommand{\thefigure}{S\arabic{figure}}
\setcounter{algorithm}{0}
\renewcommand{\thealgorithm}{S\arabic{algorithm}}

\section{Detailed FWSP Formulations, Algorithms, and Extended Related Work}
\label{app:method_details}

\paragraph{Extended Related Work.}
Beyond the concise summary in Section~\ref{sec:related}: (i) in Weight-Space Programming, \texttt{DoubleAI}~\cite{doubleai2026redesign} uses coding agents to splice Python modules via ghost gradients or attach external probe-rollback wrappers without modifying static weights, while Ahmed~\yrcite{ahmed2026wsp} showed that zero-shot LLM weight inspection relies heavily on boundary-layer arithmetic and memorized priors; (ii) in model merging, Eq.~\eqref{eq:fwsp_mopd} recovers \texttt{RegMean}~\cite{jin2023regmean} when base-centered regularization ($\lambda{=}0$) and token weighting ($D_k{=}I$) are removed, while our FWSP instantiation adds $\lambda W_0$, unsupervised trajectory curation, and response-token weighting ($w_t \in \{0, 0.5, 2.0\}$); and (iii) in reasoning control~\cite{chen2024overthinking}, FWSP contrasts with \texttt{ThinkEdit}~\cite{sun2025thinkedit} (unwhitened attention-head edits targeting overly short thinking) by compiling covariance-whitened, subspace-protected readout/MLP edits directly into static checkpoints.

\paragraph{Diagnostic Data Splits and Unified Task--Data--Operator Map.}
Across all tasks, $\mathcal{D}_{\text{diag}}$ uses $8$--$64$ disjoint examples under \texttt{no\_grad}: (1) \emph{unlabeled shifted batches} (e.g., $64$ corrupted images for TTA; $0$ data for static $\mathcal{O}_1$ localization); (2) \emph{unlabeled contrastive pairs $(\mathcal{D}^+, \mathcal{D}^-)$} (preserved vs.\ targeted activations) probed from the same edited model (e.g., $64$ clean vs.\ triggered images for vision backdoors; $8$ disjoint training math prompts for reasoning LLMs where tokens $10$--$130$ form $X_{\text{start}}$, tokens $>330$ form $X_{\text{tail}}$, and appending a label-free closure template yields $X_{\text{post}}$); (3) \emph{few-shot labeled pairs $(X, Y)$} (continual vision / NLP calibration); and (4) \emph{unlabeled teacher rollouts} curated by Teacher-vs.-Base response log-probability difference for multi-teacher $\mathcal{O}_4$. Table~\ref{tab:solver_taxonomy} summarizes the complete Task--Data--Operator mapping and Stage~2 protected subspace instantiations.

\begin{table}[htbp]
\centering
\caption{Unified Task--Data--Operator map for FWSP. \textbf{(a)} End-to-end mapping from each task regime to its diagnostic data $\mathcal{D}_{\text{diag}}$, flat pipeline path (direct Stage~3 vs.\ Stage~2 $\to$ 3), Stage~1 layer/tensor scope $\mathcal{L}^*$, and optional Stage~1 blind routing / Stage~4 gate. \textbf{(b)} Stage~2 protected subspace instantiations and unwhitened/unprotected ablations under Eq.~\eqref{eq:pfg_unified}.}
\label{tab:solver_taxonomy}
\footnotesize
\resizebox{\textwidth}{!}{%
\begin{tabular}{lllll}
\toprule
\multicolumn{5}{l}{\textbf{(a) Task--Data--Operator Mapping Across All Five Evaluation Regimes}} \\
\textbf{Regime \& Concrete Tasks} & \textbf{Diagnostic Batch $\mathcal{D}_{\text{diag}}$ Construction} & \textbf{Pipeline Path (Stage~2 $\to$ 3)} & \textbf{Stage~1 Scope $\mathcal{L}^*$} & \textbf{Routing / Stage-4 Gate} \\
\midrule
\textbf{R1 Structural Faults} (Bit-flip, row outlier) & Zero data (static weight scan) & Direct Stage~3: $\mathcal{O}_1$ (Clip / rescale) & Top anomalous 2D tensor & Auto blind / Empirical gate \\
\textbf{R3 Distribution Shift} (TTA, long-CoT decay) & Unlabeled shifted batch ($64$ corrupted imgs) & Direct Stage~3: $\mathcal{O}_3$ (Moment align) & Readout / \texttt{L6--11} / \texttt{dt\_bias} & Auto or Task / Opt.\ guard \\
\textbf{R2/R4 Behavior \& Control} (Backdoor, erasure, & Unlabeled contrast pairs $(\mathcal{D}^+, \mathcal{D}^-)$ (clean vs.\ trigger; & \textbf{Stage~2} ($U^*$) $\to$ \textbf{Stage~3} $\mathcal{O}_2$ / & High-SNR MLP / DiT / & Auto or Task / Ret.\ guard \\
\quad reasoning CoT length, VLM hallucination) & \quad normal vs.\ tail tokens; retain vs.\ target prompts) & \quad readout addition & \quad \texttt{</think>} readout row & \\
\textbf{R4 Supervised Adapt.} (Prune, CIL, NLP, edit) & Few-shot labeled pairs $(X, Y)$ ($8$--$64$ examples) & Direct Stage~3: $\mathcal{O}_4$ (1-layer ridge) & Classifier / \texttt{lm\_head} / MLP & Auto or Task / Opt.\ guard \\
\textbf{Multi-Teacher Synthesis} (\texttt{MOPD} suites) & Curated unlabeled teacher trajectories (Top-$64$ rollouts) & Direct Stage~3: $\mathcal{O}_4$ (all-layer, Eq.~\eqref{eq:fwsp_mopd}) & All linear proj.\ (e.g., 252) & All-layer / Val-pool select \\
\midrule
\multicolumn{5}{l}{\textbf{(b) Stage~2 Instantiations \& Ablations (Protected Subspace Discovery, Eq.~\eqref{eq:pfg_unified}, $Q_{\text{keep}}^\top U = 0$)}} \\
\textbf{Instantiation / Ablation} & \textbf{Target Scatter $S_{\text{target}}$} & \textbf{Whitening Metric $A$} & \textbf{Protected / Rank $r$} & \textbf{Stage~3 Weight Compilation} \\
\midrule
Diff-of-Means / \texttt{LDA} / \texttt{Ridge} (\emph{Abl.}) & $\Delta\mu\Delta\mu^\top$ & $I$ / $\Sigma_w{+}\lambda I$ / $\Sigma{+}\lambda I$ & No / $r{=}1$ & Additive readout \\
\texttt{OrthGate} (\emph{Unwhitened Abl.}) & $\mu_{\text{tail}}\mu_{\text{tail}}^\top$ & $I$ & Yes / $r{=}1$ & Additive ($-\beta$ post-closure) \\
\textbf{\texttt{Protected FisherGate} (Canon., Table~\ref{tab:solver2_ops})} & $\Delta\mu\Delta\mu^\top$ & $\Sigma_w + \lambda I$ & \textbf{Yes} / $r{=}1$ & Additive (\texttt{</think>} / gate) \\
\textbf{Protected / Whitened \texttt{SVD} (Canon., Table~\ref{tab:solver2_ops})} & $\Delta X^\top \Delta X$ & $I$ / $\Sigma_{\text{keep}} + \lambda I$ & \textbf{Yes} / $r{=}4$ & Projective $\mathcal{O}_2$ folding \\
\bottomrule
\end{tabular}}
\end{table}

\paragraph{Stage-1 Diagnostic Scores, Optional Blind Localization, and Stage-4 Verification.}
{\sloppy
Stage~1 computes static weight-outlier ratios $\rho_k = \max|W_k|/\mathrm{med}|W_k|$ and $\eta_k = \max_i\|W_{k,i:}\|_2/\mathrm{med}_j\|W_{k,j:}\|_2$ ($0$ FPE) alongside activation moments $(\mu_l^\pm, \Sigma_l^\pm)$. When target layer/tensor scope is unknown (blind repair), Stage~1 scores $S_{\text{anom}}(k) = \max(\rho_k/\mathrm{med}_j(\rho_j), \eta_k/\mathrm{med}_j(\eta_j))$, Fisher SNR $S_{\text{Fisher}}(l) = \|\mu_l^- - \mu_l^+\|_2^2 / (\mathrm{tr}(\Sigma_l^+) + \mathrm{tr}(\Sigma_l^-) + \varepsilon)$, and unlabeled channel-std drift $S_{\text{shift}} = (\frac{1}{D}\sum_{j=1}^D (\sigma_{\text{diag},j}/\sigma_{\text{ref},j} - 1)^2)^{1/2}$ to localize $(\mathcal{L}^*, \mathcal{O}^*)$ via Algorithm~\ref{alg:blind}. Optional Stage~4 evaluates per-sample gain $d_i$ in $\mathcal{M}_{\text{tgt}}$, paired $z = \bar{d}/(\mathrm{SE}(d)+\varepsilon)$, and performance change $\Delta\mathcal{M}_{\text{ret}}$ on $\mathcal{D}_{\text{val}}$ before committing or rolling back.
\par}

\begin{algorithm}[htbp]
\caption{Deterministic Forward Weight-Space Programming (FWSP) \emph{(Forward-Only, LLM-Free)}}
\label{alg:fwsp}
\small
\begin{algorithmic}[1]
\STATE {\bfseries Input:} Trained network $f_\theta$, diagnostic batch $\mathcal{D}_{\text{diag}}$, optional validation set $\mathcal{D}_{\text{val}}$.
\STATE {\bfseries Output:} Output weights $\theta_{\text{out}} \in \{\theta', \theta\}$.
\STATE \textbf{Stage 1 (Probe \& Localize, Core; Table~\ref{tab:pipeline_stages}):} Compute static ratios $(\rho_k, \eta_k)$ ($0$ FPE) and run $\mathcal{D}_{\text{diag}}$ under \texttt{no\_grad} to record moments $(\mu_l^\pm, \Sigma_l^\pm, X X^\top)$ in $\mathcal{S}$; fix target layer/tensor scope $(\mathcal{L}^*, \mathcal{O}^*)$ per Table~\ref{tab:solver_taxonomy}a, or localize blindly via Algorithm~\ref{alg:blind}.
\STATE \textbf{Stage 2 (Subspace Discovery, \emph{optional}; Table~\ref{tab:pipeline_stages}):} If only contrastive pairs $(\mathcal{D}^+, \mathcal{D}^-)$ are provided, extract protected edit basis $U_l^*$ via Eq.~\eqref{eq:pfg_unified} (Table~\ref{tab:solver2_ops}); else skip directly to Stage~3.
\STATE \textbf{Stage 3 (Weight Compilation, Core; Table~\ref{tab:pipeline_stages}):} For each $l \in \mathcal{L}^*$, compile explicit targets or $U_l^*$ into closed-form $\Delta W_l^*$ via Eq.~\eqref{eq:solver1_template} ($\mathcal{O}_1$--$\mathcal{O}_4$, Table~\ref{tab:solver1_ops}) and fold in-place into $W_l'$ (zero runtime hooks).
\STATE \textbf{Stage 4 (Verify \& Rollback, \emph{optional}; Table~\ref{tab:pipeline_stages}):} When gating is enabled, set $\theta_{\text{out}} = \theta'$ iff $\mathcal{D}_{\text{val}}$ satisfies $\bar{d} \ge \tau_{\text{gain}}$, $z \ge z_{\min}$, and $\Delta\mathcal{M}_{\text{ret}} \ge -\epsilon$; else rollback $\theta_{\text{out}} = \theta$.
\end{algorithmic}
\end{algorithm}

\paragraph{Exact Operator Derivations ($\mathcal{O}_1$--$\mathcal{O}_4$) and Multi-Teacher Curation Score.}
Each Stage~3 operator instantiates Eq.~\eqref{eq:solver1_template} under a specific structural constraint set $\mathcal{C}$ (with Stage~4 verifying retention empirically when enabled), paired with lightweight architecture-specific extensions:
\begin{itemize}
    \item \emph{$\mathcal{O}_1$ (Weight Repair):} Setting $X{=}I, R{=}0, \lambda{=}0$, and magnitude box $\mathcal{C}_{\text{normal}} = \{W' : |W'_{ij}| \le \tau\,\mathrm{med}|W|\}$ ($\tau \in \{6,10,14\}$) yields the exact minimum-Frobenius projection $W' = \mathrm{Clip}(W, -\tau\,\mathrm{med}|W|, \tau\,\mathrm{med}|W|)$, followed by an auxiliary row-norm rescaling rule for rows exceeding $\tau_r\,\mathrm{med}_j\|W_{j:}\|_2$ ($\tau_r{=}2.5$).
    \item \emph{$\mathcal{O}_2$ (Subspace Excision and Additive Row Gating):} In its projective form, setting $K{=}1$, $X{=}I$, $R{=}0$, $\lambda{=}0$, and nullspace constraint $U_l^\top (W_l + \Delta W_l) = 0$ yields $W_l' = (I - U_l U_l^\top)W_l$ (for orthonormal $U_l$), extended with $\alpha$-interpolation and reference-mean bias alignment $\mu_l^+$. In its additive rank-1 readout form (\texttt{Protected FisherGate}), setting $K{=}1$, $X{=}I$, $R = \gamma e_{\text{tgt}} u_l^\top$, $\lambda{=}0$ over the structured single-row subspace $\mathcal{C}_{\text{row}}(u_l) = \{c\,e_{\text{tgt}} u_l^\top : c \in \mathbb{R}\}$ with $u_l = \hat{d}_{\text{PFG}}^* \perp Q_{\text{keep}}$ yields the exact closed-form update $W_l' = W_l + \gamma e_{\text{tgt}} u_l^\top$.
    \item \emph{$\mathcal{O}_3$ (Moment Alignment):} Setting residual target $R = W[(G-I)X + c\mathbf{1}^\top]$ with $\lambda{=}0$ under diagonal scaling $G = \mathrm{diag}(\sigma_{\text{clean}}/\sigma_{\text{shift}})$ folds directly into readout/normalization weights ($W' = WG, b' = b + Wc$); architecture-specific extensions update CNN BatchNorm moments or the 1D Gated-DeltaNet decay bias \texttt{dt\_bias}.
    \item \emph{$\mathcal{O}_4$ (Weighted Ridge, Protected $\mathcal{O}_4$, and Multi-Teacher Synthesis):} Unconstrained Eq.~\eqref{eq:solver1_template} with $\lambda > 0$ yields the exact multi-source ridge minimizer $\Delta W_{l,\text{ridge}}^* = M_l(\Sigma_l + \lambda I)^{-1}$ with $M_l = \sum_k (\alpha_k/c_k) R_{l,k} D_k X_{l,k}^\top$ and $\Sigma_l = \sum_k (\alpha_k/c_k) X_{l,k} D_k X_{l,k}^\top$ (Table~\ref{tab:solver1_ops}). Optional rank-$r$ truncation $\mathrm{SVD}_r(\Delta W_{l,\text{ridge}}^*)$ acts as a post-hoc Frobenius low-rank approximation to $\Delta W_{l,\text{ridge}}^*$ rather than the exact rank-$r$-constrained minimizer of the weighted objective, and is exact whenever no truncation is applied. In the single-source standard case ($K{=}1, D_1{=}I, c_1{=}1$), this reduces to $\Delta W_l^* = R_l X_l^\top (X_l X_l^\top + \lambda I)^{-1}$. For 12B \texttt{FLUX.1-schnell} concept erasure, restricting Eq.~\eqref{eq:solver1_template} to the exact retained-concept nullspace $\mathcal{C}_{\text{null}} = \{\Delta W : \Delta W Q_k = 0\}$ via $\Delta W = Z P_{\text{null}}$ ($P_{\text{null}} = I - Q_k Q_k^\top = P_{\text{null}}^2 = P_{\text{null}}^\top$, SVD threshold $0.05\,\sigma_1$ on non-target classes and neutral prefix) with scale-normalized ridge ($\lambda = 0.05\,\mathrm{tr}(G_t)/D$, $G_t = E_t^\top E_t / N_t$, $G_{at} = E_a^\top E_t / N_t$) yields the first-order KKT condition $Z P_{\text{null}}(G_t + \lambda I)P_{\text{null}} = W_0(G_{at} - G_t)P_{\text{null}}$. Using the exact projector commutation identity $P_{\text{null}}(G_t P_{\text{null}} + \lambda I) = P_{\text{null}} G_t P_{\text{null}} + \lambda P_{\text{null}} = (P_{\text{null}} G_t P_{\text{null}} + \lambda I)P_{\text{null}}$ (both factors are invertible for $\lambda > 0$), we obtain $P_{\text{null}}(P_{\text{null}} G_t P_{\text{null}} + \lambda I)^{-1} P_{\text{null}} \equiv P_{\text{null}}(G_t P_{\text{null}} + \lambda I)^{-1}$, proving that \texttt{Protected $\mathcal{O}_4$} applied to both T5 (\texttt{context\_embedder}) and CLIP (\texttt{time\_text\_embed.\allowbreak text\_embedder.\allowbreak linear\_1}) projections is the exact unique constrained minimizer of Eq.~\eqref{eq:solver1_template}:
    \begin{equation}
    \begin{aligned}
        \Delta W^*_{\text{prot-}\mathcal{O}_4}
        &\;=\; W_0 (G_{at} - G_t) P_{\text{null}} \left(G_t P_{\text{null}} + \lambda I\right)^{-1} \\
        &\;\equiv\; W_0 (G_{at} - G_t) P_{\text{null}} \left(P_{\text{null}} G_t P_{\text{null}} + \lambda I\right)^{-1} P_{\text{null}}.
    \end{aligned}
        \label{eq:flux_prot_o4}
    \end{equation}
    For multi-teacher synthesis across all linear projections $\mathcal{L}$ given an SFT base $W_{l,0}$ and $K$ teachers $\{W_{l,0} + \Delta W_{l,k}\}_{k=1}^K$, setting $R_{l,k} = \Delta W_{l,k} X_{l,k}$ and trace-normalization $c_k = \mathrm{tr}(D_k)$ in Eq.~\eqref{eq:solver1_template} (so $\bar{D}_k = D_k / \mathrm{tr}(D_k)$ and $\Sigma_{l,k} = X_{l,k} \bar{D}_k X_{l,k}^\top$) yields Eq.~\eqref{eq:fwsp_mopd} identically:
    \begin{equation}
        W_l' = W_{l,0} + \Big(\sum_{k=1}^K \alpha_k \Delta W_{l,k} \Sigma_{l,k}\Big)\Big(\sum_{k=1}^K \alpha_k \Sigma_{l,k} + \lambda I\Big)^{-1},
        \qquad \forall l \in \mathcal{L}.
        \label{eq:fwsp_mopd}
    \end{equation}
    Candidate teacher rollouts $(x, y)$ generated from disjoint training prompts are scored unsupervisedly by mean per-token Teacher-vs.-Base response log-probability divergence:
    \begin{equation}
        s_k(x, y) \;=\; \frac{1}{|y|} \sum_{t=1}^{|y|} \Big(\log p_{\theta_k}(y_t \mid x, y_{<t}) - \log p_{\theta_0}(y_t \mid x, y_{<t})\Big),
        \label{eq:curation_score}
    \end{equation}
    selecting the Top-$64$ trajectories per domain (from $160$ candidates) and weighting tokens via diagonal matrix $D_k$ ($w_t{=}0$ on the first $8$ shared chat-template tokens, $0.5$ on prompt tokens, $2.0$ on response tokens) to form $\Sigma_k = X_k D_k X_k^\top / \mathrm{tr}(D_k)$. Removing base-centered regularization ($\lambda{=}0$) and token weighting ($D_k{=}I$) recovers \texttt{RegMean}~\cite{jin2023regmean} as a layer-wise analytic optimizer.
\end{itemize}

\paragraph{Stage~2 Instantiations, Unwhitened \texttt{OrthGate} Ablation, and Local Invariance.}
Letting $V$ be an orthonormal basis for $\mathrm{range}(I - Q_{\text{keep}} Q_{\text{keep}}^\top)$ and substituting $U = V Z$ reduces Eq.~\eqref{eq:pfg_unified} to $S_V Z = A_V Z \Lambda$ with closed-form solution $U^* = V A_V^{-1/2}\mathrm{Eig}_r(A_V^{-1/2} S_V A_V^{-1/2})$. For reasoning readout editing on $8$ math rollouts, $Q_{\text{keep}} = \mathrm{QR}([\hat{\mu}_{\text{start}}, (\hat{\mu}_{\text{mid}},) \hat{\mu}_{\text{post}}])$ and $\Delta_{\text{gap}}(q) = \mathrm{Quantile}_q(\max_v z_v(h_t) - z_{\texttt{</think>}}(h_t))$ ($q{=}0.03$ default; $q{=}0.25$ fallback if Stage~4 detects zero-gap collapse) set unified gain $\gamma = 1.05\,\Delta_{\text{gap}}(q)/\langle\mu_{\text{tail}}, \hat{d}\rangle$ (with budget scaling $s \in (0, 1]$ for $8\text{K}$). The unwhitened protection-only ablation \texttt{OrthGate} ($A{=}I, r{=}1$) uses post-closure suppression ($\beta{=}0.72$):
\begin{equation}
    \hat{d}_{\text{orth}} = \frac{(I - Q_{\text{keep}} Q_{\text{keep}}^\top)\mu_{\text{tail}}}{\|(I - Q_{\text{keep}} Q_{\text{keep}}^\top)\mu_{\text{tail}}\|_2}, \qquad w_{\texttt{</think>}} \leftarrow w_{\texttt{</think>}} + \gamma \hat{d}_{\text{orth}} - \beta \gamma \hat{u}_{\text{post},\perp}.
    \label{eq:orthgate_dir}
\end{equation}
For rank-$r$ diffusion/DiT excision, $X_{\text{tgt}}, X_{\text{bg}}, X_{\text{keep}}$ are collected from target, neutral anchor, and unrelated retention prompts with $Q_{\text{keep}} = \mathrm{SVD}_{k_{\text{keep}}}^{\text{right}}(X_{\text{keep}})$ and $\Delta X = X_{\text{tgt}} - X_{\text{bg}}$. Under $Q_{\text{keep}}^\top U^* = 0$ (verified numerically to $\|Q_{\text{keep}}^\top U^*\|_\infty < 10^{-7}$), any protected vector $h = Q_{\text{keep}} a$ incurs $\Delta z(h) = \gamma u^\top Q_{\text{keep}} a = 0$ for pure orthogonal additive updates (\texttt{Protected FisherGate} and suppression-free \texttt{OrthGate}, whereas full \texttt{OrthGate} with $\beta > 0$ intentionally shifts post-closure logits along $\hat{u}_{\text{post},\perp}$ to prevent \texttt{</think>} re-emission) and $h' = (I - \alpha P_U)h = h$ for rank-$r$ projective edits. This guarantees exact \emph{local} invariance on $\mathrm{span}(Q_{\text{keep}})$ rather than invariance over full multi-step decoding trajectories, motivating Stage-4 verification.

\begin{algorithm}[htbp]
\caption{Optional Blind Weight-Space Program Synthesis (Automatic Fault Routing)}
\label{alg:blind}
\small
\begin{algorithmic}[1]
\STATE {\bfseries Input:} Model $f_\theta$, reference $(\mathcal{M}_{\text{ref}}, \mathcal{D}_{\text{ref}})$, diagnostic $\mathcal{D}_{\text{diag}}$ (optional contrastive pairs $\mathcal{D}_{\text{pair}}^+$ or labels $Y_{\text{diag}}$), validation set $\mathcal{D}_{\text{val}}$, fixed thresholds $(\tau_{\text{anom}}{=}4.0,\; \tau_{\text{Fisher}}{=}0.05,\; \tau_{\text{shift}}{=}0.08,\; \tau_H{=}0.95)$.
\STATE {\bfseries Output:} Verified single- or two-step edited weights $\theta'$.
\STATE Compute $S_{\text{anom}}(k)$ across all $K$ matrices ($0$ FPE); let $k^* = \arg\max_k S_{\text{anom}}(k)$.
\IF{$S_{\text{anom}}(k^*) > \tau_{\text{anom}}$}
    \STATE Apply $\mathcal{O}_1(W_{k^*})$; keep iff Stage-4 passes (gain $\ge +1.0$ pp, $z \ge z_{\min}$, $\Delta\mathcal{M}_{\text{ret}} \ge -\epsilon$), else rollback.
\ENDIF
\STATE Compute relative health $H(\theta) = \mathcal{M}_{\text{tgt}}(f_\theta; \mathcal{D}_{\text{val}})/\mathcal{M}_{\text{ref}}$ (inverted for lower-is-better metrics).
\IF{$H(\theta) \ge \tau_H$ {\bfseries and} $\mathcal{D}_{\text{pair}}^+ = \varnothing$}
    \STATE {\bfseries return} current weights
\ENDIF
\STATE Extract readout features $(\Phi_{\text{ref}}, \Phi_{\text{diag}})$ to compute $S_{\text{shift}}$ and (if $\mathcal{D}_{\text{pair}}^+ \neq \varnothing$) $S_{\text{Fisher}}$.
\IF{$S_{\text{Fisher}} > \tau_{\text{Fisher}}$ {\bfseries and} $Y_{\text{diag}} = \varnothing$}
    \STATE Apply $\mathcal{O}_2$ on top-SNR block; keep iff Stage-4 passes (target gain $\ge +0.5$ pp, $z \ge z_{\min}$, $\Delta\mathcal{M}_{\text{ret}} \ge -\epsilon$), else rollback.
\ELSIF{$S_{\text{shift}} > \tau_{\text{shift}}$ {\bfseries and} $Y_{\text{diag}} = \varnothing$}
    \STATE Apply unlabeled $\mathcal{O}_3$ (BN or readout rescaling); keep iff Stage-4 passes (gain $\ge +0.5$ pp, $z \ge z_{\min}$, $\Delta\mathcal{M}_{\text{ret}} \ge -\epsilon$).
\ENDIF
\IF{$Y_{\text{diag}} \neq \varnothing$ {\bfseries and} $H(\theta) < \tau_H$}
    \STATE Apply $\mathcal{O}_4$ ridge patch; keep iff Stage-4 passes (gain $\ge +0.5$ pp, $z \ge z_{\min}$, $\Delta\mathcal{M}_{\text{ret}} \ge -\epsilon$).
\ENDIF
\STATE Recompute $H(\theta)$ on the current weights $\theta'$.
\IF{$H(\theta) < \tau_H$ {\bfseries and} $S_{\text{anom}}(k^*) > \tau_{\text{anom}}$ {\bfseries and} $S_{\text{shift}} > \tau_{\text{shift}}$}
    \STATE Starting from initial checkpoint $\theta$ (Depth-2 Lookahead fallback, $+2$ val passes), evaluate $(\mathcal{O}_1 \!\to\! \mathcal{O}_3)$ and $(\mathcal{O}_3 \!\to\! \mathcal{O}_1)$; commit the higher-scoring Stage-4-verified pair iff it improves over both $\theta$ and current $\theta'$.
\ENDIF
\STATE {\bfseries return} final verified weights $\theta'$
\end{algorithmic}
\end{algorithm}

\section{Complete Mathematical Proofs}
\label{app:proofs}

\subsection{Proof of Proposition 1 (Subspace Excision vs.\ Feature Entanglement Bound)}
\begin{proof}
Let $P_U = I - U U^\top$ be the orthogonal projector onto $\mathrm{span}(U)^\perp$ (since $U^\top U = I_r$, $P_U^2 = P_U = P_U^\top$). For any failure activation $h^- = z_{\mathrm{core}} + v_{\mathrm{fault}}$ with $v_{\mathrm{fault}} \in \mathrm{span}(U)$, applying $\mathcal{O}_2$ with $\alpha = 1$ and positive-mean bias alignment yields:
\begin{equation}
    \tilde{h}^- = P_U(z_{\mathrm{core}} + v_{\mathrm{fault}}) + U U^\top \mu^+ = P_U z_{\mathrm{core}} + U U^\top \mu^+,
\end{equation}
since $P_U v_{\mathrm{fault}} = 0$, eliminating the fault component identically. On clean activations $h^+ = z_{\mathrm{core}} + \xi$ with $\mathbb{E}[h^+] = \mu^+$, $\xi \perp z_{\mathrm{core}}$, and total covariance $\Sigma_+ = \mathrm{Cov}(h^+) = \Sigma_{\mathrm{core}} + \Sigma_\xi$, the perturbation induced by $\mathcal{O}_2$ is:
\begin{equation}
    \delta^+ \triangleq \tilde{h}^+ - h^+ = (P_U h^+ + U U^\top \mu^+) - h^+ = -U U^\top (h^+ - \mu^+).
\end{equation}
Because $\mathbb{E}[\delta^+] = -U U^\top \mathbb{E}[h^+ - \mu^+] = 0$, we have $\mathbb{E}[\langle \nabla\mathcal{L}(\mu^+), \delta^+ \rangle] = 0$ identically. By $L$-smoothness of $\mathcal{L}$ on the convex activation region containing $\{h^+, \mu^+, h^+ + \delta^+\}$ and their interpolating segments:
\begin{equation}
    \mathcal{L}(h^+ + \delta^+) - \mathcal{L}(h^+) \;\le\; \langle \nabla \mathcal{L}(\mu^+), \delta^+ \rangle + \langle \nabla \mathcal{L}(h^+) - \nabla \mathcal{L}(\mu^+), \delta^+ \rangle + \frac{L}{2} \|\delta^+\|_2^2.
\end{equation}
Taking expectations, the first term vanishes. For the quadratic term, $\mathbb{E}[\|\delta^+\|_2^2] = \mathbb{E}[(h^+ - \mu^+)^\top U U^\top (h^+ - \mu^+)] = \mathrm{tr}(U^\top \Sigma_+ U)$. For the gradient-variation term, Cauchy--Schwarz and $L$-Lipschitz continuity of $\nabla\mathcal{L}$ yield:
\begin{align}
    \mathbb{E}\!\left[\langle \nabla\mathcal{L}(h^+) - \nabla\mathcal{L}(\mu^+), \delta^+ \rangle\right]
    &\le L\,\mathbb{E}\!\left[\|h^+ - \mu^+\|_2 \|\delta^+\|_2\right]
    \le L\sqrt{\mathbb{E}\!\left[\|h^+ - \mu^+\|_2^2\right]\mathbb{E}\!\left[\|\delta^+\|_2^2\right]} \nonumber\\
    &= L\sqrt{\mathrm{tr}(\Sigma_+)\,\mathrm{tr}(U^\top \Sigma_+ U)}.
\end{align}
Combining both terms gives $\mathbb{E}[\mathcal{L}(\tilde{h}^+) - \mathcal{L}(h^+)] \le \frac{L}{2}\mathrm{tr}(U^\top \Sigma_+ U) + L\sqrt{\mathrm{tr}(\Sigma_+)\,\mathrm{tr}(U^\top \Sigma_+ U)}$. When $\nabla\mathcal{L}(h^+) \equiv \nabla\mathcal{L}(\mu^+)$ is locally constant, the cross-term vanishes identically, tightening the bound to $\frac{L}{2}\mathrm{tr}(U^\top \Sigma_+ U)$. (For label-dependent supervised losses $\mathcal{L}(h, y)$, the same argument applies per class $y$ under class-conditional mean alignment $\mu_y^+$, or globally whenever $\nabla_h\mathcal{L}(\mu^+, y)$ is uncorrelated with $\delta^+$.)
\end{proof}

\subsection{Proof of Proposition 2 (Frozen-Span and Walsh--Hadamard Parity Barrier)}
\begin{proof}
Because upstream weights are frozen, the affine output $\Phi_\theta(X)(W+\Delta W)^\top + \mathbf{1}(b+\Delta b)^\top = \tilde{\Phi}\,\Theta^\top$ (with $\tilde{\Phi} = [\Phi_\theta(X), \mathbf{1}]$ and $\Theta = [W+\Delta W,\; b+\Delta b]$) lies strictly in the column space $\mathcal{C} = \mathrm{range}(\tilde{\Phi}) \subseteq \mathbb{R}^N$. By the orthogonal projection theorem in Hilbert space $(\mathbb{R}^{N\times C}, \|\cdot\|_F)$, the unique minimum-norm residual onto $\mathcal{C}$ is attained at $\tilde{\Phi}\Theta^{*\top} = \Pi_{\tilde{\Phi}} Y^*$, giving:
\begin{equation}
    \min_{\Delta W, \Delta b} \|\Phi_\theta(X)(W+\Delta W)^\top + \mathbf{1}(b+\Delta b)^\top - Y^*\|_F^2 = \|(I - \Pi_{\tilde{\Phi}}) Y^*\|_F^2.
\end{equation}
(Note that this barrier applies specifically to frozen-upstream single-layer linear edits such as readout $\mathcal{O}_4$, and does not directly constrain all-layer multi-teacher $\mathcal{O}_4$ synthesis [Eq.~\eqref{eq:fwsp_mopd}] or multi-layer non-gradient methods such as \texttt{MeZO} and evolution strategies, though isotropic perturbation variance scales as $O(d)$ in parameter dimension.)

For $k$ independent uniform Boolean bits $b \in \{-1, +1\}^k$, the parity target $Y^*(b) = \chi_{[k]}(b) = \prod_{j=1}^k b_j$ is zero-mean ($\mathbb{E}_b[\chi_{[k]}(b)] = 0$, hence orthogonal to $\mathbf{1}$) and forms the unique degree-$k$ Walsh--Hadamard basis element, orthogonal to every function of degree $< k$: $\mathbb{E}_b[\phi(b)\chi_{[k]}(b)] = 0$. In an idealized local-receptive-field network whose pooled features are strictly patch-additive across disjoint patches $P_1, \dots, P_k$ (that is, every channel has the form $\phi(X) = \frac{1}{k}\sum_{j=1}^k \psi(P_j)$ with receptive field confined to a single patch), every feature has Walsh--Hadamard degree $\le 1 < k$ for all $k \ge 2$. Consequently, $\Pi_{\tilde{\Phi}} Y^* = 0$ and $\|(I - \Pi_{\tilde{\Phi}})Y^*\|_F^2 = \|Y^*\|_F^2$: the unique minimum-norm linear least-squares readout predicts identically zero (achieving $50\%$ accuracy on balanced parity), and by the Minsky--Papert parity theorem no linear hyperplane over strictly additive patch features can achieve perfect ($100\%$) separation for $k \ge 2$ (while unbalanced non-least-squares linear thresholds can exceed $50\%$, e.g., $6/8$ via $-\operatorname{sign}(b_1+b_2+b_3)$ for $k{=}3$, they remain strictly bounded away from $100\%$). Furthermore, standard deep CNNs such as \texttt{ResNet-18} have late-stage strided receptive fields that can partially overlap across adjacent patches, so the strictly patch-additive model is an architectural idealization explaining why frozen-upstream \texttt{ResNet-18} remains near chance ($49.09\%$) under $\mathcal{O}_4$, whereas global self-attention in \texttt{ViT-Small} architecturally supports cross-patch bilinear interactions $Q K^\top V$ and empirically achieves $100\%$ in Table~\ref{tab:r5_wall}.
\end{proof}

\subsection{Proposition 3 (Order-Invariant Joint-Solution Equivalence of Recursive Continual $\mathcal{O}_4$)}
\begin{proposition}
Let $\mathcal{D}_1, \dots, \mathcal{D}_T$ be sequential class-incremental batches with frozen bias-augmented column features $X_t \in \mathbb{R}^{(D_{\text{in}}+1) \times n_t}$ and residual targets $R_t = Y_t^* - W_0 X_t \in \mathbb{R}^{D_{\text{out}} \times n_t}$ defined against a shared fixed reference weight $W_0$ (with $R_t = Y_t^*$ when initializing $W_0 = 0$). For any $\lambda > 0$ (or whenever $G_T = \sum_{t=1}^T X_t X_t^\top \succ 0$ is full-rank), accumulating $G_T = \sum_{t=1}^T X_t X_t^\top$ and $M_T = \sum_{t=1}^T R_t X_t^\top$ and solving the un-truncated ridge update $W_T^* = W_0 + M_T (G_T + \lambda I)^{-1}$ (without SVD truncation) yields the unique exact joint ridge minimizer over $\bigcup_{t=1}^T \mathcal{D}_t$ with zero raw-data replay, zero sequential-order forgetting relative to the joint ridge oracle, and $O(D_{\text{in}}^2 + D_{\text{out}} D_{\text{in}})$ memory (storing $G_T$ and $M_T$).
\end{proposition}
\begin{proof}
By block-matrix multiplication, $\sum_{t=1}^T X_t X_t^\top = X_{1:T} X_{1:T}^\top$ and $\sum_{t=1}^T R_t X_t^\top = R_{1:T} X_{1:T}^\top$ because every $R_t = Y_t^* - W_0 X_t$ uses the same fixed $W_0$. Setting the gradient of $\sum_{t=1}^T \|\Delta W X_t - R_t\|_F^2 + \lambda \|\Delta W\|_F^2$ with respect to $\Delta W = W - W_0$ to zero gives $\Delta W (X_{1:T} X_{1:T}^\top + \lambda I) = R_{1:T} X_{1:T}^\top$, identical to $\Delta W_T^* (G_T + \lambda I) = M_T$ with $W_T^* = W_0 + \Delta W_T^*$. Note that this guarantees exact equivalence to the un-truncated joint-training ridge solution (eliminating sequential-order drift relative to the joint ridge oracle), while overall test accuracy may still vary naturally as the total number of classes increases (e.g., from 20 to 50 classes in Table~\ref{tab:dinov2_foundation}(c)).
\end{proof}

\section{Mechanistic Analysis of Sequence-Length Compression via \texttt{OrthGate} and $\mathcal{O}_3$}

\begin{table}[htbp]
\centering
\caption{Token-level mechanistic verification ($23{,}430$ held-out tokens) and $2\times 2$ ablation on \texttt{DeepSeek-R1-0528-8B} (\texttt{GSM8K} $1\text{K}$, $N{=}128$, HF greedy, 8 math rollouts).}
\label{tab:orthgate_mech_main}
\resizebox{\textwidth}{!}{%
\begin{tabular}{lcc|lcccc}
\toprule
\multicolumn{3}{l|}{\textbf{(a) Token-Level $\Delta z_{\texttt{</think>}}(h_t)$ ($23{,}430$ Tokens)}} & \multicolumn{5}{l}{\textbf{(b) $2\times 2$ Ablation \& Closed-Form Probes (\texttt{GSM8K} ($1\text{K}$), $N{=}128$, HF)}} \\
\textbf{Token Regime ($N_{\text{tok}}$)} & \textbf{Mean $\pm$ Std} & \textbf{AUROC vs.\ Tail} & \textbf{Variant (Forward-Only, Rank-1)} & \textbf{Acc (\%)} & \textbf{Tokens} & \textbf{Trunc (\%)} & \textbf{Multi-Tag / Loop (\%)} \\
\midrule
Early Derivation ($5{,}434$) & $+6.92 \pm 13.11$ & $0.728$ ($\hat{d}_{\text{orth}}$) & Unedited Base & $12.50$ & $984.4$ & $82.8$ & $0.0$ / $0.0$ \\
Hard Normal Late ($5{,}346$) & $+7.00 \pm 12.54$ & $0.677$ ($\hat{d}_{\text{orth}}$) & No Orth ($\beta{=}0$ / $\beta{=}0.72$) & $0.00$ / $0.00$ & $335$ / $529$ & $0.0$ & $100.0$ / $100.0$ \\
Overthinking Tail ($11{,}104$) & $\mathbf{+13.40 \pm 13.51}$ & --- & Difference-of-Means (Unwhitened) & $0.00$ & $722.1$ & $0.0$ & $100.0$ / $100.0$ \\
Post-Closure ($1{,}546$) & $\mathbf{-0.67 \pm 16.93}$ & $0.746$ (full $\Delta z$) & Orth Only ($\gamma \hat{d}_{\text{orth}}, \beta{=}0$) & $78.91$ & $649.2$ & $3.9$ & $25.0$ / $17.2$ \\
\cmidrule(lr){1-3}
\multicolumn{3}{l|}{\emph{Matched HF Budget Forcing ($N{=}128$):}} & Full \texttt{OrthGate} ($\beta{=}0.72$) & $81.25$ & $642.7$ & $3.1$ & $39.1$ / $28.1$ \\
\multicolumn{3}{l|}{1K: Base $12.5\%$ / BF $50.8\%$ ($977$) / \texttt{OrthGate} $81.3\%$ ($643$)} & Whitened \texttt{Fisher LDA} Probe & $\mathbf{88.28}$ & $542.8$ & $\mathbf{0.0}$ & $19.5$ / $18.0$ \\
\multicolumn{3}{l|}{2K: Base $45.3\%$ / BF $60.2\%$ ($1{,}713$) / \texttt{OrthGate} $88.3\%$ ($787$)} & Closed-Form \texttt{Ridge} Probe & $\mathbf{88.28}$ & $\mathbf{541.1}$ & $\mathbf{0.0}$ & $19.5$ / $18.0$ \\
 & & & \textbf{\texttt{Protected FisherGate} ($Q_{\text{keep}}^\top d{=}0$)} & $85.16$ & $585.6$ & $0.8$ & $\mathbf{14.8}$ / $\mathbf{9.4}$ \\
\bottomrule
\end{tabular}}
\end{table}
\label{app:orthgate_mech}
A practical byproduct of FWSP's reasoning calibration (Section~\ref{sec:sota_math_code}) is \emph{zero-overhead, adaptive chain-of-thought (CoT) length compression}: across \texttt{Qwen3.5-0.8B} and \texttt{DeepSeek-R1-0528-8B}, FWSP reduces average generation length by 14.7--17.1\% on math and 35.1\% on code without modifying the decoding engine or adding inference-time harness overhead. Here we derive the exact mechanism explaining why \texttt{OrthGate} and $\mathcal{O}_3$ shorten sequence length selectively rather than blindly truncating derivations.

\paragraph{Three-Phase Logit Dynamics of \texttt{OrthGate}.}
Let $h_t \in \mathbb{R}^d$ denote the pre-readout hidden state at decoding step $t$. \texttt{OrthGate} modifies only the single \texttt{</think>} row of the untied readout matrix \texttt{lm\_head} by the rank-1 vector:
\begin{equation}
    \Delta w_{\texttt{</think>}} \;=\; \gamma \hat{d}_{\text{orth}} - \beta \gamma \hat{u}_{\text{post},\perp},
\end{equation}
where $\hat{d}_{\text{orth}} \propto (I - Q_{\text{keep}} Q_{\text{keep}}^\top)\mu_{\text{tail}}$ with $Q_{\text{keep}} = \mathrm{QR}([\hat{\mu}_{\text{start}}, \hat{\mu}_{\text{post}}])$, unit orthogonal vector $\hat{u}_{\text{post},\perp} \propto \hat{\mu}_{\text{post}} - (\hat{\mu}_{\text{post}}^\top \hat{\mu}_{\text{start}})\hat{\mu}_{\text{start}}$, and $\beta \in (0,1)$ (default $\beta = 0.72$). At every step $t$, this edit induces an additive logit shift $\Delta z_{\texttt{</think>}}(h_t) = \langle h_t, \Delta w_{\texttt{</think>}} \rangle$ across three rollout phases:
\begin{enumerate}
    \item \emph{Phase I: Early Derivation ($h_t \approx \mu_{\text{start}}$).} By construction of $Q_{\text{keep}}$ and $\hat{u}_{\text{post},\perp}$, both $\langle \mu_{\text{start}}, \hat{d}_{\text{orth}} \rangle = 0$ and $\langle \mu_{\text{start}}, \hat{u}_{\text{post},\perp} \rangle = 0$, giving $\Delta z_{\texttt{</think>}}(\mu_{\text{start}}) = 0$ at the phase centroid (individual held-out tokens fluctuate around $\mu_{\text{start}}$, while adding $\mu_{\text{mid}}$ when available further protects long valid chains). Thus \texttt{OrthGate} avoids prematurely aborting productive initial problem solving.
    \item \emph{Phase II: Repetitive Overthinking Tail ($h_t \approx \mu_{\text{tail}}$).} When a reasoning model enters self-doubt or redundant re-verification loops (probed from late completion states beyond token 330 on 8 practice rollouts), $h_t$ drifts into a subspace with positive projection $\langle h_t, \hat{d}_{\text{orth}} \rangle > 0$. Because $\beta \in (0,1)$ scales down the post-closure suppression term and $\gamma$ is calibrated from the empirical 3rd-percentile tail logit gap $\Delta_{\text{gap}} = \mathrm{Quantile}_{0.03}(\max_v z_v(h_t) - z_{\texttt{</think>}}(h_t))$ with Stage-4 validation, the net shift $\Delta z_{\texttt{</think>}}(h_t)$ bridges the top-1 margin selectively at natural clause boundaries inside an overthinking tail, triggering \texttt{</think>} adaptively instead of rambling until the generation budget expires.
    \item \emph{Phase III: Post-Closure Final Answer ($h_t \approx \mu_{\text{post}}$).} Immediately after \texttt{</think>} is emitted, the hidden state shifts toward $\mu_{\text{post}}$. Here $\langle \mu_{\text{post}}, \hat{d}_{\text{orth}} \rangle = 0$ while the negative term yields $\Delta z_{\texttt{</think>}}(\mu_{\text{post}}) = -\beta \gamma \langle \mu_{\text{post}}, \hat{u}_{\text{post},\perp} \rangle < 0$. This actively suppresses re-emission of \texttt{</think>}, preventing the \texttt{</think></think>...} repetition failure mode of constant scalar logit bias and allowing the model to emit a concise final answer and hit \texttt{EOS}.
\end{enumerate}

\paragraph{Complementary Recurrent-Decay Calibration ($\mathcal{O}_3$ on \texttt{dt\_bias}).}
On hybrid linear-attention architectures such as \texttt{Qwen3.5-0.8B} (Gated-DeltaNet), long self-doubt chains can saturate the recurrent state before termination is triggered. Shifting the 1D linear-attention time-step bias \texttt{dt\_bias} ($\mathcal{O}_3$, $\Delta = -0.30$) moderates recurrent state accumulation, breaking verbatim repetition attractors so that \texttt{OrthGate} can close the reasoning trace cleanly.

\paragraph{Operational Comparison Against Alternative Length-Control Methods.}
Compared to three standard ways of controlling reasoning length, weight-embedded \texttt{OrthGate} occupies a distinct operational niche:
\begin{itemize}
    \item \emph{vs.\ Constant Scalar \texttt{</think>} Logit Bias:} A state-independent bias $+c$ shifts $z_{\texttt{</think>}}$ identically across all three phases, creating a sharp trade-off between premature closure in Phase~I and post-closure \texttt{</think>} repetition in Phase~III.
    \item \emph{vs.\ External Inference-Time Budget Forcing:} Hard budget forcing requires a custom decoding wrapper and intervenes at a fixed token threshold regardless of problem difficulty, whereas \texttt{OrthGate} is compiled into a single weight row, runs natively on unmodified serving stacks (e.g., standard \texttt{generate}, vLLM, SGLang), and terminates adaptively when overthinking states emerge.
    \item \emph{vs.\ Backprop Length-Penalty Finetuning (SFT/RL):} Length-regularized RL or SFT requires gradients, optimizer memory, and thousands of rollouts, whereas FWSP computes $\Delta w_{\texttt{</think>}}$ deterministically in closed form from 8 forward-probed practice rollouts and protects against degradation via Stage-4 paired verification.
\end{itemize}

\paragraph{Structural Preconditions.}
Two architectural preconditions govern this mechanism: (i) when \texttt{lm\_head.weight} is weight-tied to \texttt{embed\_tokens.weight} (as in \texttt{Qwen3.5-0.8B}), \texttt{lm\_head} must be untied prior to applying $\Delta w_{\texttt{</think>}}$ so that editing the output projection does not corrupt the input embedding of \texttt{</think>} once emitted; and (ii) \texttt{</think>} should ideally be represented by a dedicated single control token. When it splits into ordinary multi-token BPE pieces (as in \texttt{GLM-Z1-9B-0414}), prefix gating on \texttt{</} can still recover long-chain \texttt{MATH-500} ($+7.2$ pp with 3-vector protection) but trades off shorter \texttt{GSM8K} and code generations, so default Stage-4 verification safely abstains with $0.0\%$ drift.

\paragraph{Empirical Token-Level Verification and \texttt{Protected FisherGate} Diagnostics.}
Table~\ref{tab:app_pfg_diag}(a) directly compares token-level logit shifts across \texttt{OrthGate}, unconstrained \texttt{Fisher LDA}, and \texttt{Protected FisherGate} using the exact frozen math-calibrated patches on 9 held-out truncated practice-pool traces (note that Table~\ref{tab:orthgate_mech_main}(a) probes a separate 32-trace \texttt{GSM8K} evaluation set including solved normal rollouts, accounting for differences in absolute \texttt{OrthGate} shift magnitudes). On the same held-out traces, unconstrained \texttt{Fisher LDA} achieves slightly higher Tail-vs-Early AUROC ($0.961$) than \texttt{Protected FisherGate} ($0.955$), yet collapses on zero-shot \texttt{HumanEval} ($25.61\%$ vs.\ $84.15\%$), illustrating that \emph{representation separability does not imply intervention quality}: controlling how an edit perturbs non-target phases is essential for cross-task transfer. Table~\ref{tab:app_pfg_diag}(b) reports all five independent 8-rollout math calibration subsets evaluated on the first 128 unseen \texttt{GSM8K} problems (indices 128--255, where unedited Base is $21.09\%$, distinct from the 256-item slice in Table~\ref{tab:full_confirmatory} where Base is $14.84\%$), showing moderately aligned directions (pairwise cosine $0.411 \pm 0.059$) with consistent Fisher separation ($89.0$--$105.2$) alongside the 2/5 gain-quantile degeneracies caught by Stage-4 verification.

\begin{table}[htbp]
\centering
\caption{Mechanistic comparison of frozen math patches on 9 held-out truncated traces and 5-subset calibration robustness diagnostics (\texttt{DeepSeek-R1-0528-8B}).}
\label{tab:app_pfg_diag}
\resizebox{\textwidth}{!}{%
\begin{tabular}{lcccc|cccccc}
\toprule
\multicolumn{5}{l|}{\textbf{(a) Held-Out Token-Level Logit Shift $\Delta z_{\texttt{</think>}}(h_t)$}} & \multicolumn{6}{l}{\textbf{(b) 5-Subset Calibration Robustness (Unseen GSM 128, Base $21.1\%$)}} \\
\textbf{Editor} & \textbf{Early / Tail / Post} & \textbf{AUROC (T/E)} & \textbf{AUROC (T/P)} & \textbf{HE ($1\text{K}$)} & \textbf{Subset} & \textbf{Fisher Sep.} & \textbf{$\Delta_{\text{gap}}$ (Run)} & \textbf{Stage-4} & \textbf{Unseen GSM} & \textbf{HE ($1\text{K}$)} \\
\midrule
\texttt{OrthGate} & $+19.3$ / $+24.0$ / $-40.6$ & $0.575$ & $0.993$ & $24.39\%$ & S0 & $89.0$ & $25.9$ & Pass & $88.28\%$ & $79.88\%$ \\
\texttt{Fisher LDA} & $-16.7$ / $+25.4$ / $-31.9$ & $\mathbf{0.961}$ & $\mathbf{0.999}$ & $25.61\%$ & S1 & $98.2$ & $0.0^*$ & Reject & $21.09\%$ & $33.54\%$ \\
\textbf{\texttt{Protected PFG}} & $+7.0$ / $+25.6$ / $+1.1$ & $0.955$ & $0.998$ & $\mathbf{84.15\%}$ & S2 & $96.4$ & $26.5$ & Pass & $87.50\%$ & $64.63\%$ \\
\multicolumn{5}{l|}{\footnotesize Pairwise direction cosine across all 5 PFG subsets: $0.411 \pm 0.059$ ($[0.316, 0.506]$).} & S3 & $94.6$ & $0.0^*$ & Reject & $21.09\%$ & $33.54\%$ \\
\multicolumn{5}{l|}{\footnotesize $^*$3rd-percentile quantile collapsed to $0.0$ (zero-magnitude update); never below Base.} & S4 & $105.2$ & $26.7$ & Pass & $85.94\%$ & $81.71\%$ \\
\bottomrule
\end{tabular}}
\end{table}

\section{Complete Supplementary Experimental Tables}
\label{app:tables}

\begin{table}[htbp]
\centering
\caption{Foundation vision evaluation on \texttt{DINOv2-Base} (86.6M, $224\times 224$): (a) structural fault taxonomy (ImageNet-20, 3 seeds), (b) unlabeled BadNets backdoor excision, (c) 4-stage ImageNet-50 CIL, and (d) full official 10-stage \texttt{CIFAR-100} CIL ($50\text{K}$ train / $10\text{K}$ test, 100 classes, 0 backprop).}
\label{tab:dinov2_foundation}
\resizebox{\textwidth}{!}{%
\begin{tabular}{llcccc}
\toprule
\multicolumn{6}{l}{\textbf{(a) Structural Fault Taxonomy (ImageNet-20 Test Acc \%, Clean = 98.33\%)}} \\
\textbf{Corruption Type} & & \textbf{Corrupted Base} & \textbf{MeZO (1,536 FPE)} & \textbf{FWSP $\mathcal{O}_1$} & \textbf{FWSP $\mathcal{O}_1\!\to\!\mathcal{O}_4$} \\
\midrule
IEEE-754 Bit-Flip & & $6.72 \pm 1.06$ & $6.06 \pm 0.48$ & $72.67 \pm 0.14$ & $\mathbf{82.22 \pm 1.36}$ \\
Row Scale Outlier & & $6.28 \pm 0.61$ & $6.44 \pm 1.00$ & $70.44 \pm 0.21$ & $\mathbf{80.89 \pm 0.55}$ \\
65\% Unstructured Pruning & & $73.17 \pm 0.00$ & $75.17 \pm 0.98$ & $73.17 \pm 0.00$ \emph{(abs.)} & $\mathbf{84.44 \pm 1.31}$ \\
\midrule
\multicolumn{6}{l}{\textbf{(b) Unlabeled BadNets Backdoor Excision ($224\times 224$, 64 Unlabeled Pairs)}} \\
\textbf{Metric} & & \textbf{Poisoned Base} & \textbf{Fine-Pruning} & \textbf{MeZO (SPSA)} & \textbf{FWSP $\mathcal{O}_2$ (Ours)} \\
\midrule
Backdoor ASR (\%) $\downarrow$ & & $100.00 \pm 0.00$ & $100.00 \pm 0.00$ & $49.24 \pm 38.89$ & $\mathbf{0.00 \pm 0.00}$ \\
Clean ImageNet Acc (\%) $\uparrow$ & & $98.39 \pm 0.21$ & $98.17 \pm 0.14$ & $73.06 \pm 26.35$ & $\mathbf{98.44 \pm 0.39}$ \\
\midrule
\multicolumn{6}{l}{\textbf{(c) 4-Stage Continual Class-Incremental Learning (All-Seen ImageNet Acc \%, 20 $\to$ 50 Classes)}} \\
\textbf{Method} & \textbf{Total FPE} & \textbf{20 Cls} & \textbf{30 Cls} & \textbf{40 Cls} & \textbf{50 Cls} \\
\midrule
\texttt{MeZO} Continual & $62{,}000$ & $22.00$ & $9.67$ & $7.00$ & $3.73$ \\
Sequential \texttt{Head-FT} & $112{,}500$ & $98.50$ & $98.00$ & $94.00$ & $91.00$ \\
\texttt{NCM} Prototype & $1{,}250$ & $98.50$ & $98.67$ & $97.08$ & $\mathbf{97.00}$ \\
\textbf{FWSP $\mathcal{O}_4$ (Recursive, Ours)} & $\mathbf{1{,}250}$ ($90\times$) & $98.17$ & $98.22$ & $96.58$ & $96.40$ \\
\bottomrule
\end{tabular}}

\vspace{4pt}
\resizebox{\textwidth}{!}{%
\begin{tabular}{lcccccccccccc}
\toprule
\multicolumn{13}{l}{\textbf{(d) Official Full 10-Stage \texttt{CIFAR-100} Continual Learning ($50\text{K}$ Train / $10\text{K}$ Test, 10 Classes/Stage, Order-Invariant, 0 Backprop)}} \\
\textbf{Method (\texttt{DINOv2-Base})} & \textbf{10} & \textbf{20} & \textbf{30} & \textbf{40} & \textbf{50} & \textbf{60} & \textbf{70} & \textbf{80} & \textbf{90} & \textbf{100 (Final)} & \textbf{Avg.\ Inc.\ ($\uparrow$)} & \textbf{Wins} \\
\midrule
\texttt{NCM} Prototype & $96.20$ & $95.65$ & $94.03$ & $92.35$ & $91.62$ & $90.08$ & $89.54$ & $88.81$ & $87.54$ & $87.05$ & $91.29\%$ & $0/10$ \\
FWSP $\mathcal{O}_4$ Linear (Recursive) & $96.80$ & $95.75$ & $94.03$ & $92.52$ & $91.66$ & $90.25$ & $89.59$ & $88.50$ & $87.24$ & $86.50$ & $91.28\%$ & $0/10$ \\
\textbf{FWSP $\mathcal{O}_4$ + \texttt{RanPAC}~\cite{mcdonnell2023ranpac} ($M{=}4{,}096$, Ours)} & $\mathbf{97.10}$ & $\mathbf{96.00}$ & $\mathbf{94.53}$ & $\mathbf{93.05}$ & $\mathbf{92.50}$ & $\mathbf{91.23}$ & $\mathbf{90.31}$ & $\mathbf{89.44}$ & $\mathbf{88.36}$ & $\mathbf{87.62}$ & $\mathbf{92.01\%}$ ($\mathbf{+0.72}$) & $\mathbf{10/10}$ \\
\bottomrule
\end{tabular}}
\vspace{2pt}
\par\noindent{\footnotesize Note: Panel~(d) isolates the ablation between pure linear recursive $\mathcal{O}_4$ ($86.50\%$ final / $91.28\%$ avg.) and $\mathcal{O}_4$ paired with frozen \texttt{RanPAC}~\cite{mcdonnell2023ranpac} random ReLU features ($M{=}4{,}096$; $+1.12\text{ pp}$ final, $+0.73\text{ pp}$ avg.\ over linear $\mathcal{O}_4$, beating \texttt{NCM} on all $10/10$ stages with zero backprop).}
\end{table}

\begin{table}[htbp]
\centering
\caption{Full official NLP validation/test calibration ($18{,}287$ queries total on \texttt{Qwen3-4B-Instruct-2507}) vs.\ 4-shot ICL (reporting both unweighted 3-task macro average and query-weighted micro average), alongside runtime and peak memory comparisons.}
\label{tab:resource_and_c2}
\resizebox{\textwidth}{!}{%
\begin{tabular}{lcccc|lcc}
\toprule
\textbf{Official Benchmark ($N$)} & \textbf{Zero-Shot} & \textbf{4-Shot ICL} & \textbf{FWSP $\mathcal{O}_4$} & \textbf{ICL Tokens} & \textbf{Configuration} & \textbf{Time (ms)} & \textbf{VRAM (GB)} \\
\midrule
\texttt{SST-2} val ($872$) & $88.19\%$ & $\mathbf{93.92\%}$ & $92.55\%$ & $3.70\times$ & \texttt{DINOv2} FWSP $\mathcal{O}_4$ & $\mathbf{73.5}$ & $\mathbf{2.24}$ \\
\texttt{AG News} test ($7{,}600$) & $76.97\%$ & $80.08\%$ & $\mathbf{80.51\%}$ & $5.98\times$ & \texttt{DINOv2} Full-FT (5 steps) & $994.1$ & $14.28$ \\
\texttt{MNLI}-m val ($9{,}815$) & $82.57\%$ & $82.66\%$ & $\mathbf{86.05\%}$ & $4.86\times$ & \texttt{Qwen3-8B} FWSP $\mathcal{O}_4$ & $1{,}014.5$ & $\mathbf{15.63}$ \\
\midrule
\emph{Unweighted 3-Task Macro} & \emph{82.58\%} & \emph{85.55\%} & \emph{\textbf{86.37\%} ($+0.82$)} & \emph{$4.85\times$} & \texttt{Qwen3-8B} LoRA (5 steps) & $\mathbf{810.5}$ & $17.15$ \\
\emph{Query-Weighted Micro ($18{,}287$)} & \emph{80.51\%} & \emph{82.12\%} & \emph{\textbf{84.06\%} ($+1.94$)} & \emph{$5.27\times$} & \emph{(All FWSP runs use $1\times$ query tok)} & --- & --- \\
\bottomrule
\end{tabular}}
\end{table}

\begin{table}[htbp]
\centering
\caption{Zero-shot single-checkpoint Math$\to$\texttt{HumanEval} ($1\text{K}$ budget, $N{=}164$) transfer with AST code-quality diagnostics on \texttt{DeepSeek-R1-0528-8B} and cross-model replication on \texttt{Qwen3.5-4B} (all edits calibrated on math rollouts only; HF greedy).}
\label{tab:pfg_xtask}
\resizebox{\textwidth}{!}{%
\begin{tabular}{lccccc|lccc}
\toprule
\multicolumn{6}{l|}{\textbf{(a) \texttt{DeepSeek-R1-0528-8B}: Math$\to$\texttt{HumanEval} ($1\text{K}$, $N{=}164$)}} & \multicolumn{4}{l}{\textbf{(b) \texttt{Qwen3.5-4B} Cross-Model Replication (HF, 1K)}} \\
\textbf{Method} & \textbf{pass@1 (\%)} & \textbf{Valid Fn (\%)} & \textbf{Syntax Err (\%)} & \textbf{Trunc (\%)} & \textbf{Tokens} & \textbf{Method (Math-Calib.)} & \textbf{Math$\to$HE ($1\text{K}$)} & \textbf{Valid Fn / Tok} & \textbf{Unseen GSM ($1\text{K}$)} \\
\midrule
Unedited Base & $33.54$ & $35.4$ & $64.6$ & $60.4$ & $919.5$ & Unedited Base & $81.71\%$ & $90.8\%$ / $826$ & $24.22\%$ ($1{,}012$) \\
Budget Forcing (\texttt{BF-128}) & $31.71$ & $34.2$ & $65.9$ & $0.0$ & $912.2$ & Math-\texttt{OrthGate} & $78.66\%$ & $88.4\%$ / $722$ & $\mathbf{64.06\%}$ ($522$) \\
Budget Forcing (\texttt{BF-256}) & $41.46$ & $56.7$ & $43.3$ & $0.0$ & $900.0$ & Math-\texttt{Fisher LDA} & $10.37\%$ & $11.6\%$ / $874$ & $25.00\%$ ($887$) \\
Budget Forcing (\texttt{BF-512}) & $69.51$ & $82.3$ & $17.7$ & $0.0$ & $837.6$ & \textbf{Math-\texttt{Protected PFG}} & $\mathbf{87.80\%}$ & $\mathbf{97.0\%}$ / $\mathbf{692}$ & $3.12\%$ ($996$)$^*$ \\
Math-\texttt{OrthGate} & $24.39$ & $26.2$ & $73.8$ & $0.0$ & $787.6$ & \multicolumn{4}{l}{\footnotesize $^*$Without $\mathcal{O}_3$ decay calibration on hybrid GDN; Stage-4 rejects on math.} \\
Math-\texttt{Fisher LDA} & $25.61$ & $29.9$ & $70.1$ & $0.0$ & $775.5$ & \multicolumn{4}{l}{\footnotesize Paired significance vs.\ \texttt{BF-512}: $+14.6$ pp ($95\%$ CI $[+7.3, +22.6]$),} \\
Math-\texttt{Ridge} & $26.22$ & $30.5$ & $69.5$ & $0.0$ & $786.3$ & \multicolumn{4}{l}{\footnotesize McNemar $p{=}5.3\times 10^{-4}$; vs.\ \texttt{BF-256}: $+42.7$ pp, $p{=}1.6\times 10^{-16}$.} \\
\textbf{Math-\texttt{Protected PFG}} & $\mathbf{84.15}$ & $\mathbf{92.7}$ & $\mathbf{7.3}$ & $\mathbf{0.0}$ & $\mathbf{641.2}$ & \multicolumn{4}{l}{\footnotesize All PFG/LDA/Ridge use unified gain $\gamma$ and $\beta{=}0$.} \\
\midrule
\multicolumn{10}{l}{\textbf{(c) Multi-Budget \texttt{HumanEval} ($2\text{K}$--$4\text{K}$, Greedy) \& Stochastic Sampling ($T{=}0.6$, $\text{top-}p{=}0.95$, HF, \texttt{DeepSeek-R1-0528-8B})}} \\
\textbf{Evaluation Setting ($N$)} & \multicolumn{2}{c}{\textbf{Base Acc / Tok}} & \multicolumn{2}{c}{\textbf{PFG Acc / Tok}} & \textbf{Valid Fn (\%)} & \multicolumn{2}{c}{\textbf{Paired Diff (95\% CI)}} & \textbf{Token $\Delta$} & \textbf{McNemar $p$} \\
\midrule
\texttt{HumanEval} $2\text{K}$ Greedy ($164$) & \multicolumn{2}{c}{$43.90\%$ / $1{,}546$} & \multicolumn{2}{c}{$\mathbf{84.76\%}$ / $\mathbf{684}$} & $93.3$ & \multicolumn{2}{c}{$+40.9$ pp ($[+32.9, +48.8]$)} & $-55.8\%$ & $4.8\times 10^{-15}$ \\
\texttt{HumanEval} $4\text{K}$ Greedy ($164$) & \multicolumn{2}{c}{$54.27\%$ / $2{,}534$} & \multicolumn{2}{c}{$\mathbf{84.76\%}$ / $\mathbf{724}$} & $93.9$ & \multicolumn{2}{c}{$+30.5$ pp ($[+23.2, +37.8]$)} & $-71.5\%$ & $2.6\times 10^{-11}$ \\
Unseen \texttt{GSM8K} $1\text{K}$ Stoch.\ ($128$) & \multicolumn{2}{c}{$18.75\%$ / $969$} & \multicolumn{2}{c}{$\mathbf{89.84\%}$ / $\mathbf{548}$} & --- & \multicolumn{2}{c}{$+71.1$ pp ($[+63.3, +78.9]$)} & $-43.5\%$ & $3.9\times 10^{-21}$ \\
\texttt{HumanEval} $1\text{K}$ Stoch.\ ($164$) & \multicolumn{2}{c}{$31.71\%$ / $927$} & \multicolumn{2}{c}{$\mathbf{74.39\%}$ / $\mathbf{642}$} & $89.0$ & \multicolumn{2}{c}{$+42.7$ pp ($[+34.8, +51.2]$)} & $-30.7\%$ & $5.6\times 10^{-15}$ \\
\bottomrule
\end{tabular}}
\vspace{2pt}
\par\noindent{\footnotesize Note: Panel~(c) reports continuity-corrected asymptotic McNemar $\chi^2$ $p$-values (exact two-sided binomial $p{=}8.1\times 10^{-28}$ on Unseen \texttt{GSM8K}, 91 wins / 0 losses).}
\end{table}

\begin{table}[htbp]
\centering
\caption{Frozen confirmatory evaluation on strictly unseen HF splits (zero overlap with prior slices, $\pm 95\%$ Wilson CI) and complete official testsets under vLLM (\texttt{GSM8K} all $1{,}319$, \texttt{MATH-500} all $500$, $1\text{K}$--$8\text{K}$) using the exact same frozen HF-calibrated weight patches without recalibration (\texttt{DeepSeek-R1-0528-8B}).}
\label{tab:full_confirmatory}
\resizebox{\textwidth}{!}{%
\begin{tabular}{lcc|cccc|cccc}
\toprule
& \multicolumn{2}{c|}{\textbf{Strictly Unseen HF Splits}} & \multicolumn{4}{c|}{\textbf{Full \texttt{GSM8K} ($N{=}1{,}319$, HF$\to$vLLM)}} & \multicolumn{4}{c}{\textbf{Full \texttt{MATH-500} ($N{=}500$, HF$\to$vLLM)}} \\
\textbf{Method} & \textbf{GSM ($256$)} & \textbf{MATH ($125$)} & \textbf{1K} & \textbf{2K} & \textbf{4K} & \textbf{8K} & \textbf{1K} & \textbf{2K} & \textbf{4K} & \textbf{8K} \\
\midrule
Unedited Base & $14.8 \pm 4.4$ & $4.0 \pm 3.4$ & $25.7$ ($952$) & $58.8$ ($1{,}540$) & $84.6$ ($2{,}030$) & $91.5$ ($2{,}264$) & $4.6$ ($1{,}014$) & $21.4$ ($1{,}887$) & $42.0$ ($3{,}194$) & $60.0$ ($4{,}967$) \\
Budget Forcing & --- & --- & $83.6$ ($919$) & $\mathbf{89.5}$ ($1{,}516$) & $\mathbf{91.8}$ ($2{,}015$) & $\mathbf{92.7}$ ($2{,}256$) & $\mathbf{55.0}$ ($980$) & $64.6$ ($1{,}854$) & $\mathbf{73.8}$ ($3{,}156$) & $\mathbf{80.2}$ ($4{,}874$) \\
\texttt{OrthGate} & $77.7 \pm 5.1$ & $34.4 \pm 8.3$ & $83.7$ ($562$) & $87.6$ ($703$) & $88.7$ ($860$) & $88.3$ ($1{,}186$) & $34.0$ ($849$) & $45.6$ ($1{,}404$) & $56.8$ ($2{,}245$) & $65.6$ ($3{,}389$) \\
\texttt{Fisher LDA} & $84.8 \pm 4.4$ & $44.0 \pm 8.7$ & $85.9$ ($522$) & $88.1$ ($564$) & $88.0$ ($617$) & $88.7$ ($690$) & $47.8$ ($803$) & $61.8$ ($1{,}090$) & $63.6$ ($1{,}423$) & $65.2$ ($2{,}016$) \\
\texttt{Ridge} & $84.4 \pm 4.4$ & $45.6 \pm 8.7$ & $87.2$ ($526$) & $88.3$ ($566$) & $87.3$ ($641$) & $88.6$ ($671$) & $49.6$ ($802$) & $61.0$ ($1{,}093$) & $63.4$ ($1{,}474$) & $63.4$ ($2{,}076$) \\
\textbf{\texttt{Protected PFG} ($s{=}1.0$)} & $\mathbf{86.7 \pm 4.2}$ & $\mathbf{48.0 \pm 8.8}$ & $\mathbf{87.9}$ ($561$) & $88.9$ ($593$) & $89.1$ ($632$) & $88.9$ ($690$) & $48.0$ ($857$) & $\mathbf{66.8}$ ($1{,}106$) & $67.4$ ($1{,}389$) & $66.2$ ($1{,}862$) \\
\textbf{\texttt{Protected PFG} ($s{=}0.70$)} & --- & --- & --- & --- & --- & --- & --- & --- & $72.0$ ($2{,}350$) & $79.8$ ($2{,}745$) \\
\bottomrule
\end{tabular}}
\vspace{2pt}
\par\noindent{\footnotesize Note: Full official testsets include earlier evaluation slices, whereas the two Unseen HF columns are strictly zero-overlap. For \texttt{Protected FisherGate}, 95\% Wilson CIs are $\pm 1.7$--$1.8$ pp on full \texttt{GSM8K} ($87.9$--$89.1\%$) and $\pm 4.1$--$4.4$ pp on full \texttt{MATH-500} ($48.0$--$67.4\%$; other rows vary with accuracy, e.g., $\pm 1.8$ pp at $4.6\%$ and $\pm 3.5$ pp at $80.2\%$).}
\end{table}

\begin{table}[htbp]
\centering
\caption{Five-seed component ablations, 10-round verification stability, and hybrid FWSP + LoRA localization ablations.}
\label{tab:ablations_hybrid}
\resizebox{\textwidth}{!}{%
\begin{tabular}{lc|lccccc}
\toprule
\multicolumn{2}{l|}{\textbf{(a) $\mathcal{O}_2$ Extractor \& Bias-Align (5 Seeds, 0\% ASR)}} & \multicolumn{6}{l}{\textbf{(b) 10-Round Sequential Guard (Worst-Group Acc \%)}} \\
\textbf{Variant} & \textbf{Clean Acc (\%)} & \textbf{Editing Protocol} & \textbf{R0} & \textbf{R1} & \textbf{R2} & \textbf{R5} & \textbf{R10} \\
\midrule
Pre-Edit Poisoned (100\% ASR) & $97.83 \pm 0.00$ & Unverified Greedy & $7.00$ & $7.50$ & $6.75$ & $7.00$ & $7.00$ \\
Diff-of-Means (No Align) & $97.50 \pm 0.15$ & \textbf{FWSP w/ Paired Verifier} & $7.00$ & $\mathbf{7.50}$ & $\mathbf{7.50}$ & $\mathbf{7.50}$ & $\mathbf{7.50}$ \\
\cmidrule(lr){3-8}
Whitened Fisher (No Align) & $97.70 \pm 0.37$ & \multicolumn{6}{l}{\textbf{(c) Hybrid FWSP + LoRA Localization Ablations (Damaged + Frost \texttt{DINOv2})}} \\
cPCA (+Bias Align) & $98.13 \pm 0.07$ & \textbf{Method} & \multicolumn{3}{c}{\textbf{Trainable Params}} & \multicolumn{2}{c}{\textbf{Test Acc (\%, 5 Seeds)}} \\
\cmidrule(lr){3-8}
\textbf{Whitened Fisher + Align (Ours)} & $\mathbf{98.23 \pm 0.17}$ & Damaged + Shift Base & \multicolumn{3}{c}{$0$} & \multicolumn{2}{c}{$5.07 \pm 0.25$} \\
 & & Blind 12L \texttt{LoRA} (10 / 50 steps) & \multicolumn{3}{c}{$1.34\text{M}$} & \multicolumn{2}{c}{$8.57 \pm 5.26$ / $15.27 \pm 7.34$} \\
 & & Pure FWSP ($\mathcal{O}_1\!+\!\mathcal{O}_3$, gradient-free) & \multicolumn{3}{c}{$0$} & \multicolumn{2}{c}{$85.33 \pm 0.30$} \\
 & & FWSP + Last-L11 / Rand-1L \texttt{LoRA} & \multicolumn{3}{c}{$0.13\text{M}$} & \multicolumn{2}{c}{$73.70 \pm 0.59$ / $84.43 \pm 7.13$} \\
 & & FWSP + Full 12L \texttt{LoRA} (10 steps) & \multicolumn{3}{c}{$1.34\text{M}$} & \multicolumn{2}{c}{$82.47 \pm 2.79$} \\
 & & $\mathcal{O}_1$ + Localized 1L \texttt{LoRA} & \multicolumn{3}{c}{$0.13\text{M}$} & \multicolumn{2}{c}{$87.90 \pm 0.79$} \\
 & & \textbf{FWSP + Localized 1L \texttt{LoRA} (Ours)} & \multicolumn{3}{c}{$\mathbf{0.13\text{M}}$ ($10.7\times$ less)} & \multicolumn{2}{c}{$\mathbf{88.10 \pm 0.52}$} \\
\bottomrule
\end{tabular}}
\end{table}

\begin{table}[htbp]
\centering
\caption{Qualitative self-repair transcript on \texttt{Qwen3-8B} (Layer 18 damaged, 2.67s FWSP $\mathcal{O}_1$, no optimizer LLM).}
\label{tab:qual}
\scriptsize
\resizebox{\textwidth}{!}{%
\begin{tabular}{p{2.2cm}p{11.0cm}}
\toprule
\textbf{State} & \textbf{Greedy Continuation for} \emph{``Photosynthesis is a biological process used by plants to convert light energy into''} \\
\midrule
Clean 8B LM & \emph{``...chemical energy. The process requires carbon dioxide, water, and light. The chemical equation for photosynthesis is 6CO2 + 6''} \\
Corrupted L18 & \emph{``...,1. 0 (the following 1. 1 IN THE following derived, 1. 1. 1''} \textcolor{red}{(Collapsed, PPL 7375.66)} \\
FWSP $\mathcal{O}_1$ (2.67s) & \emph{``...chemical energy. The process requires carbon dioxide, water, and light. The chemical equation for photosynthesis is 6CO2 + 6''} \textcolor{blue}{(Verbatim recovery, PPL 5.45)} \\
\bottomrule
\end{tabular}}
\end{table}

\begin{table}[htbp]
\centering
\caption{Non-differentiable \texttt{Gymnasium CartPole-v1} control (5 seeds, optimal ceiling 200) and zero-label physics-verified \texttt{1D Viscous Burgers FNO} (relative $L_2$ PDE error $\downarrow$).}
\label{tab:gym_pde}
\resizebox{\textwidth}{!}{%
\begin{tabular}{llcccc}
\toprule
\textbf{Domain / Regime} & \textbf{Metric} & \textbf{Base / Damaged} & \textbf{ZO / Backprop Baseline} & \textbf{FWSP (Ours)} & \textbf{Clean / Oracle} \\
\midrule
\texttt{CartPole-v1} R1 (Bit-Flip) & Return $\uparrow$ & $129.81 \pm 86.28$ & $86.64 \pm 92.58$ (\texttt{MeZO}) & $\mathbf{200.00 \pm 0.00}$ ($\mathcal{O}_1$) & $200.00 \pm 0.00$ \\
\texttt{CartPole-v1} R2 (Backdoor) & Return $\uparrow$ & $9.22 \pm 0.30$ & $9.17 \pm 0.24$ (\texttt{ES}) & $\mathbf{200.00 \pm 0.00}$ ($\mathcal{O}_2$) & $200.00 \pm 0.00$ \\
\midrule
\texttt{Burgers FNO} R1 (Label-Free Phys.) & Rel.\ $L_2$ (\%) $\downarrow$ & $70.62 \pm 19.26$ & $51.18 \pm 4.41$ (\texttt{MeZO}) & $\mathbf{5.63 \pm 1.34}$ ($\mathcal{O}_1$) & $1.88 \pm 0.22$ \\
\texttt{Burgers FNO} R3 (Visc.\ Shift) & Rel.\ $L_2$ (\%) $\downarrow$ & $12.91 \pm 0.40$ & --- & $\mathbf{5.58 \pm 0.54}$ ($\mathcal{O}_4$) & --- \\
\texttt{Burgers FNO} R5 (Shock Wall) & Rel.\ $L_2$ (\%) $\downarrow$ & $19.18 \pm 0.30$ & $19.36 \pm 0.52$ (16m \texttt{FT}) & $19.64 \pm 0.37$ ($\mathcal{O}_4$) & \emph{Nyquist limit} \\
\bottomrule
\end{tabular}}
\end{table}

\begin{table}[htbp]
\centering
\caption{Full 15-corruption official \texttt{CIFAR-100-C} Severity-5 TTA benchmark on \texttt{DINOv2-Base} ($88.37\%$ clean, $150{,}000$ images; $64$-image adapt batch). Backprop TTA baselines (\texttt{SAR}, \texttt{EATA}, \texttt{Tent}, \texttt{DeYO}; $960$--$1{,}920$ FPE) vs.\ zero-backprop FWSP $\mathcal{O}_3$ ($64$ FPE, $15$--$30\times$ cheaper).}
\label{tab:app_imgc}
\resizebox{\textwidth}{!}{%
\begin{tabular}{llccccccc}
\toprule
\textbf{Category} & \textbf{Corruption (Sev-5)} & \textbf{Base ($0$)} & \textbf{\texttt{SAR} ($1{,}920$)} & \textbf{\texttt{EATA} ($960$)} & \textbf{\texttt{Tent} ($960$)} & \textbf{\texttt{DeYO} ($1{,}280$)} & \textbf{FWSP 1L ($64$)} & \textbf{FWSP L6--11 ($64$, Ours)} \\
\midrule
\multirow{3}{*}{Noise} & Gaussian Noise & $42.39$ & $42.32$ & $44.71$ & $46.05$ & $\mathbf{46.11}$ & $44.10$ & $45.42$ \\
 & Shot Noise & $44.01$ & $44.00$ & $46.46$ & $47.47$ & $\mathbf{47.59}$ & $45.59$ & \underline{$47.58$} \\
 & Impulse Noise & $61.40$ & $61.29$ & $63.46$ & $\mathbf{64.45}$ & $64.43$ & $62.75$ & $58.55$ \\
\midrule
\multirow{4}{*}{Blur} & Defocus Blur & $65.92$ & $65.90$ & $68.23$ & $68.93$ & $\mathbf{68.97}$ & $66.30$ & $68.27$ \\
 & Glass Blur & $13.50$ & $13.52$ & $13.51$ & $13.90$ & $13.89$ & $15.96$ & $\mathbf{19.98}$ ($+6.1$) \\
 & Motion Blur & $61.10$ & $60.96$ & $62.63$ & $63.47$ & $63.53$ & $61.56$ & $\mathbf{63.79}$ ($+0.3$) \\
 & Zoom Blur & $68.14$ & $68.15$ & $69.43$ & $69.83$ & $69.80$ & $68.74$ & $\mathbf{72.53}$ ($+2.7$) \\
\midrule
\multirow{4}{*}{Weather} & Snow & $69.67$ & $69.66$ & $71.16$ & $71.87$ & $71.90$ & $70.06$ & $\mathbf{72.12}$ ($+0.2$) \\
 & Frost & $70.63$ & $70.68$ & $71.60$ & $72.44$ & $72.38$ & $71.07$ & $\mathbf{72.74}$ ($+0.3$) \\
 & Fog & $16.81$ & $16.84$ & $17.45$ & $18.23$ & $18.27$ & $19.26$ & $\mathbf{26.94}$ ($+8.7$) \\
 & Brightness & $84.23$ & $84.13$ & $84.58$ & $84.74$ & $84.73$ & $84.19$ & $\mathbf{84.84}$ ($+0.1$) \\
\midrule
\multirow{4}{*}{Digital} & Contrast & $77.83$ & $77.85$ & $79.43$ & $\mathbf{79.94}$ & $79.90$ & $78.02$ & $78.25$ \\
 & Elastic Transform & $40.99$ & $40.99$ & $42.35$ & $43.84$ & $43.88$ & $42.82$ & $\mathbf{48.88}$ ($+5.0$) \\
 & Pixelate & $49.32$ & $49.31$ & $54.89$ & $\mathbf{56.58}$ & $56.43$ & $50.75$ & $54.71$ \\
 & JPEG Compression & $60.99$ & $61.02$ & $61.85$ & $\mathbf{62.70}$ & $62.63$ & $61.32$ & $60.03$ \\
\midrule
\multicolumn{2}{l}{\textbf{15-Corruption Mean (\%) / Regime}} & $55.13$ (0 grad.) & $55.11$ (grad.) & $56.78$ (grad.) & \underline{$57.63$} (grad.) & \underline{$57.63$} (grad.) & $56.17$ (0 grad.) & $\mathbf{58.31}$ (\textbf{0 grad., $15$--$30\times$ less FPE}) \\
\bottomrule
\end{tabular}}
\end{table}

\begin{table}[htbp]
\centering
\caption{Diffuse all-layer isotropic weight noise sweep on \texttt{DINOv2-Base} ($W + \mathcal{N}(0, \sigma^2 I)$). Raw ungated operator accuracy illustrates the R1 sub-boundary between localized outliers and dense isotropic noise; at $\sigma{=}0.008$, Stage-4 paired verification rejects edits below the 85.50\% baseline.}
\label{tab:app_diffuse}
\small
\begin{tabular}{ccccc}
\toprule
\textbf{Noise $\sigma$} & \textbf{Corrupted Base (\%)} & \textbf{Unverified $\mathcal{O}_1$ (\%)} & \textbf{FWSP $\mathcal{O}_4$ (\%)} & \textbf{Grad LoRA (\%)} \\
\midrule
$0.008$ & $85.50$ & $6.67$ & $48.17$ & $\mathbf{70.33}$ \\
$0.016$ & $4.83$ & $5.33$ & $\mathbf{19.67}$ & $6.67$ \\
$0.026$ & $4.83$ & $4.83$ & $\mathbf{15.17}$ & $14.33$ \\
$0.038$ & $5.67$ & $5.50$ & $\mathbf{15.00}$ & $14.83$ \\
\bottomrule
\end{tabular}
\end{table}

\begin{table}[htbp]
\centering
\caption{Initial cross-family tight-budget screening on Math (\texttt{GSM8K} $N{=}128$, \texttt{MATH-500} $N{=}125$) and Code (\texttt{HumanEval} $N{=}164$, \texttt{MBPP} $N{=}128$, greedy). Bold marks best edited results; $^{\circ}$ denotes Stage-4 abstention. Full official testsets appear in Table~\ref{tab:sota_2026_overview} and Appendix Table~\ref{tab:full_confirmatory}.}
\label{tab:sota_math_code}
\resizebox{\textwidth}{!}{%
\begin{tabular}{llccccc}
\toprule
\textbf{Model Family (Params)} & \textbf{Method} & \textbf{GSM8K (\%)} & \textbf{MATH-500 (\%)} & \textbf{HumanEval (\%)} & \textbf{MBPP (\%)} & \textbf{Mean Tokens (Math / Code)} \\
\midrule
\multirow{3}{*}{\texttt{Qwen3.5-0.8B} (0.75B)} & Unedited Base & $14.06 \pm 6.0$ & $9.60 \pm 5.2$ & $18.90 \pm 6.0$ & $24.22 \pm 7.4$ & $1{,}215$ / $1{,}041$ \\
& \texttt{MeZO} (Matched FPE) & $10.94 \pm 5.4$ & $8.80 \pm 5.0$ & $21.95 \pm 6.3$ & $26.56 \pm 7.7$ & $1{,}215$ / $989$ \\
& \textbf{FWSP (Ours)} & $\mathbf{52.34 \pm 8.7}^{\dagger}$ & $\mathbf{29.60 \pm 8.0}^{\dagger}$ & $18.90 \pm 6.0^{\circ}$ & $24.22 \pm 7.4^{\circ}$ & $\mathbf{1{,}037}$ / $1{,}041$ \\
\midrule
\multirow{3}{*}{\texttt{Qwen3.5-4B} (4.21B)} & Unedited Base & $21.88 \pm 7.2$ & $12.80 \pm 5.9$ & $85.37 \pm 5.4$ & $75.78 \pm 7.4$ & $1{,}203$ / $795$ \\
& \texttt{MeZO} (Matched FPE) & $21.09 \pm 7.1$ & $15.20 \pm 6.3$ & $82.32 \pm 5.8$ & $76.56 \pm 7.3$ & $1{,}194$ / $812$ \\
& \textbf{FWSP (Ours)} & $\mathbf{66.41 \pm 8.2}^{\dagger}$ & $\mathbf{38.40 \pm 8.5}^{\dagger}$ & $85.37 \pm 5.4^{\circ}$ & $75.78 \pm 7.4^{\circ}$ & $\mathbf{934}$ / $795$ \\
\midrule
\multirow{3}{*}{\texttt{Qwen3.5-9B} (8.95B)} & Unedited Base & $18.75 \pm 6.8$ & $19.20 \pm 6.9$ & $90.85 \pm 4.4$ & $76.56 \pm 7.3$ & $1{,}200$ / $819$ \\
& \texttt{MeZO} (Matched FPE) & $28.12 \pm 7.8$ & $16.80 \pm 6.5$ & $92.07 \pm 4.1$ & $75.78 \pm 7.4$ & $1{,}155$ / $814$ \\
& \textbf{FWSP (Ours)} & $\mathbf{62.50 \pm 8.4}^{\dagger}$ & $\mathbf{45.60 \pm 8.7}^{\dagger}$ & $89.02 \pm 4.8$ & $\mathbf{79.69 \pm 7.0}$ & $\mathbf{1{,}020}$ / $\mathbf{804}$ \\
\midrule
\multirow{3}{*}{\texttt{DeepSeek-R1-0528-8B} (8.2B)} & Unedited Base & $14.06 \pm 6.0$ & $11.20 \pm 5.5$ & $35.37 \pm 7.3$ & $40.62 \pm 8.5$ & $1{,}174$ / $957$ \\
& \texttt{MeZO} (Matched FPE) & $15.62 \pm 6.3$ & $15.20 \pm 6.3$ & $35.37 \pm 7.3$ & $42.19 \pm 8.6$ & $1{,}160$ / $969$ \\
& \textbf{FWSP (Ours)} & $\mathbf{54.69 \pm 8.6}^{\dagger}$ & $\mathbf{40.00 \pm 8.6}^{\dagger}$ & $\mathbf{44.51 \pm 7.6}^{*}$ & $\mathbf{45.31 \pm 8.6}$ & $\mathbf{973}$ / $\mathbf{621}$ \\
\midrule
\multirow{3}{*}{\texttt{GLM-Z1-9B-0414} (9.4B)} & Unedited Base & $72.66 \pm 7.7$ & $37.60 \pm 8.5$ & $40.24 \pm 7.5$ & $61.72 \pm 8.4$ & $1{,}070$ / $996$ \\
& \texttt{MeZO} (Matched FPE) & $68.75 \pm 8.0$ & $37.60 \pm 8.5$ & $35.98 \pm 7.4$ & $57.03 \pm 8.6$ & $1{,}086$ / $989$ \\
& FWSP (\emph{Stage-4 Abstain}) & $72.66 \pm 7.7^{\circ}$ & $37.60 \pm 8.5^{\circ}$ & $40.24 \pm 7.5^{\circ}$ & $61.72 \pm 8.4^{\circ}$ & $1{,}070$ / $996$ \\
\bottomrule
\end{tabular}}
\vspace{2pt}
\par\noindent{\footnotesize Significance vs.\ Unedited Base via 4,000-sample paired bootstrap~\cite{efron1993bootstrap}: $^{\dagger}p < 10^{-4}$ ($z = 5.33$--$8.34$); $^{*}p = 0.0265$ ($z = 1.98$); $^{\circ}$ denotes Stage-4 safe abstention ($0.0\%$ drift; see text for the multi-token BPE boundary on \texttt{GLM-Z1-9B-0414}). Math Mean Tokens averages across \texttt{GSM8K} ($B{=}1{,}024$) and \texttt{MATH-500} ($B{=}1{,}408$).}
\end{table}

\begin{table}[htbp]
\centering
\caption{Multi-budget vLLM reasoning controls ($1\text{K}$--$8\text{K}$, greedy) across five models on \texttt{GSM8K} ($N{=}128$) and \texttt{MATH-500} ($N{=}125$). Unified \texttt{Protected FisherGate} transfers across dense, hybrid GDN, and 186GB FP8 MoE models without inference harnesses.}
\label{tab:app_budget_ablation}
\resizebox{\textwidth}{!}{%
\begin{tabular}{llcccc|cc}
\toprule
& & \multicolumn{4}{c|}{\textbf{GSM8K Acc \% (Mean Tokens)}} & \multicolumn{2}{c}{\textbf{MATH-500 Acc \% (Mean Tokens)}} \\
\textbf{Model} & \textbf{Control} & \textbf{1K} & \textbf{2K} & \textbf{4K} & \textbf{8K} & \textbf{1K / 2K} & \textbf{4K / 8K} \\
\midrule
\multirow{4}{*}{\texttt{DeepSeek-R1-0528-8B}} & Unedited Base & $25.0$ ($964$) & $61.7$ ($1{,}548$) & $86.7$ ($2{,}023$) & $92.2$ ($2{,}272$) & $13.6$ ($995$) / $30.4$ ($1{,}776$) & $53.6$ ($2{,}838$) / $66.4$ ($4{,}268$) \\
 & Scalar Bias ($+3$ / $+6$) & $24.2$ / $27.3$ & --- & --- & --- & --- / $33.6, 28.8$ & --- \\
 & Budget Forcing & $83.6$ ($931$) & $86.7$ ($1{,}538$) & $90.6$ ($2{,}034$) & $90.6$ ($2{,}298$) & $63.2$ ($971$) / $68.0$ ($1{,}757$) & $72.8$ ($2{,}820$) / $80.8$ ($4{,}102$) \\
 & \textbf{FWSP (\texttt{OrthGate})} & $\mathbf{47.7}$ ($\mathbf{863}$) & $\mathbf{78.1}$ ($\mathbf{1{,}199}$) & $\mathbf{89.1}$ ($\mathbf{1{,}374}$) & $\mathbf{90.6}$ ($\mathbf{1{,}518}$) & $\mathbf{41.6}$ ($\mathbf{860}$) / $\mathbf{60.8}$ ($\mathbf{1{,}259}$) & $\mathbf{72.0}$ ($\mathbf{1{,}549}$) / $\mathbf{79.2}$ ($\mathbf{2{,}143}$) \\
\midrule
\multirow{3}{*}{\texttt{Qwen3.5-4B}} & Unedited Base & $21.9$ ($1{,}012$) & $57.8$ ($1{,}749$) & $66.4$ ($2{,}733$) & $74.2$ ($4{,}706$) & $5.6$ / $31.2$ & $40.0$ / $52.8$ \\
 & Scalar Bias ($+3$ / $+6$) & $21.1$ / $25.0$ & --- & --- & --- & --- / $32.8, 34.4$ & --- \\
 & Budget Forcing & $87.5$ ($977$) & $92.2$ ($1{,}728$) & $94.5$ ($2{,}730$) & $93.8$ ($4{,}319$) & $63.2$ / $72.8$ & $84.0$ / $83.2$ \\
\midrule
\multirow{3}{*}{\texttt{GLM-Z1-9B-0414}} & Unedited Base & $68.0$ ($830$) & $93.8$ ($982$) & $95.3$ ($1{,}090$) & $95.3$ ($1{,}114$) & $15.2$ / $63.2$ & $80.8$ / $86.4$ \\
 & Scalar Bias ($+3$ / $+6$) & $68.8$ / $69.5$ & --- & --- & --- & --- / $63.2, 59.2$ & --- \\
 & Budget Forcing & $88.3$ ($806$) & $96.1$ ($976$) & $97.7$ ($1{,}062$) & $96.1$ ($1{,}092$) & $53.6$ / $74.4$ & $84.8$ / $89.6$ \\
\midrule
\multirow{5}{*}{\texttt{Qwen3.8-27B}} & Unedited Base & $92.2$ ($380$) & $92.2$ ($483$) & $92.2$ ($588$) & $94.5$ ($652$) & $70.4$ / $83.2$ & $88.0$ / $92.8$ \\
 & Scalar Bias ($+3$ / $+6$) & $93.0$ / $93.0$ & --- & --- & --- & --- / $81.6, 81.6$ & --- \\
 & Budget Forcing & $94.5$ ($383$) & $95.3$ ($436$) & $93.8$ ($583$) & $96.9$ ($695$) & $81.6$ / $93.6$ & $92.8$ / $95.2$ \\
 & \textbf{FWSP (\texttt{Protected PFG})} & $91.4$ ($730$) & $\mathbf{93.8}$ ($1{,}249$) & $\mathbf{94.5}$ ($2{,}046$) & --- & $72.8$ / $81.6$ & $87.2$ / --- \\
 & \textbf{FWSP ($\mathcal{O}_3$ \texttt{dt\_bias})} & $89.8$ ($399$) & $93.0$ ($468$) & $93.0$ ($526$) & $93.8$ ($590$) & $\mathbf{72.0}$ / $81.6$ & $\mathbf{90.4}$ / $\mathbf{95.2}$ \\
\midrule
\multirow{5}{*}{\texttt{Qwen3.8-Flash-Next-FP8}} & Unedited Base & $93.8$ ($350$) & $94.5$ ($404$) & $96.1$ ($438$) & $95.3$ ($418$) & $75.2$ / $81.6$ & $88.0$ / $92.0$ \\
 & Scalar Bias ($+3$ / $+6$) & $94.5$ / $94.5$ & --- & --- & --- & --- / $80.8, 81.6$ & --- \\
 & Budget Forcing & $95.3$ ($348$) & $96.9$ ($394$) & $96.9$ ($422$) & $96.9$ ($479$) & $86.4$ / $92.0$ & $93.6$ / $94.4$ \\
 & \textbf{FWSP (\texttt{Protected PFG})} & $93.8$ ($378$) & $\mathbf{96.1}$ ($429$) & $\mathbf{96.9}$ ($528$) & --- & $\mathbf{75.2}$ / $\mathbf{82.4}$ & $88.8$ / --- \\
 & \textbf{FWSP ($\mathcal{O}_3$ \texttt{dt\_bias})} & $93.8$ ($353$) & $94.5$ ($430$) & $96.1$ ($478$) & $\mathbf{96.1}$ ($468$) & $75.2$ / $\mathbf{82.4}$ & $\mathbf{89.6}$ / $\mathbf{94.4}$ \\
\bottomrule
\end{tabular}}
\vspace{2pt}
\par\noindent{\footnotesize Note: Unified \texttt{Protected PFG} rows use paired same-run vLLM baselines (\texttt{Qwen3.8-27B} Base \texttt{HumanEval} $1\text{K}/2\text{K}/4\text{K}$: $75.0\% / 87.2\% / 92.1\% \to$ \texttt{PFG} $78.7\% / 87.8\% / 93.3\%$; \texttt{Flash-Next-FP8} Base: $81.1\% / 88.4\% / 92.7\% \to$ \texttt{PFG} $84.8\% / 88.4\% / 93.3\%$ [$95.1\%$ with $\mathcal{O}_3$]). On concise \texttt{Qwen3.8-27B} \texttt{GSM8K}, \texttt{PFG} inflates tokens ($730$--$2{,}046$), illustrating the short-CoT abstention boundary.}
\end{table}

\begin{table}[htbp]
\centering
\caption{Detailed cross-modal benchmark comparison across Vision Transformers (\texttt{DINOv2-Base}, \texttt{SigLIP 2 SO400M}), Flow-Matching DiTs (\texttt{Z-Image-Turbo 6B}, \texttt{FLUX.1-schnell 12B}), UNet Diffusion (\texttt{SD-1.5}), Unified Multimodal Models (\texttt{Ovis-U1-3B}), and VLMs (\texttt{LLaVA-1.5-7B}), with every baseline listed explicitly on its own row.}
\label{tab:app_crossmodal_pfg}
\footnotesize
\resizebox{\textwidth}{!}{%
\begin{tabular}{llcccc}
\toprule
\textbf{Benchmark \& Backbone} & \textbf{Method} & \textbf{Target Metric} & \textbf{Retention Metric} & \textbf{PSNR / Drift} & \textbf{Edit / Serving Cost} \\
\midrule
\multirow{4}{*}{\shortstack[l]{\texttt{DINOv2-Base} (ViT)\\Shift Gate (\texttt{Frost-5}$\to$Unseen)}} & \texttt{OrthGate} ($r{=}1$) & $89.26\%$ AUROC ($\uparrow$) & $83.56\%$ Unseen & $0.420$ drift ($\downarrow$) & $<0.1\text{s}$ \\
 & \texttt{Fisher LDA} ($r{=}1$) & $92.38\%$ AUROC ($\uparrow$) & $80.98\%$ Unseen & $0.452$ drift ($\downarrow$) & $<0.1\text{s}$ \\
 & \texttt{Ridge} ($r{=}1$) & $\mathbf{92.77\%}$ AUROC ($\uparrow$) & $82.01\%$ Unseen & $0.534$ drift ($\downarrow$) & $<0.1\text{s}$ \\
 & \textbf{\texttt{Protected PFG} (Ours)} & $92.06\%$ AUROC ($\uparrow$) & $81.34\%$ Unseen & $\mathbf{0.270}$ drift ($\downarrow$) & $<0.1\text{s}$ \\
\midrule
\multirow{5}{*}{\shortstack[l]{\texttt{SigLIP 2 SO400M} (ViT)\\Shift Gate \& $\mathcal{O}_1$ Repair}} & \texttt{OrthGate} ($r{=}1$) & $100.0\%$ AUROC ($\uparrow$) & $80.83\%$ Unseen & $\mathbf{0.000}$ drift ($\downarrow$) & $<0.1\text{s}$ \\
 & \texttt{Ridge} ($r{=}1$) & $100.0\%$ AUROC ($\uparrow$) & $81.91\%$ Unseen & $0.180$ drift ($\downarrow$) & $<0.1\text{s}$ \\
 & \texttt{Fisher LDA} ($r{=}1$) & $100.0\%$ AUROC ($\uparrow$) & $\mathbf{81.96\%}$ Unseen & $0.012$ drift ($\downarrow$) & $<0.1\text{s}$ \\
 & \textbf{\texttt{Protected PFG} (Ours)} & $\mathbf{100.0\%}$ AUROC ($\uparrow$) & $81.59\%$ Unseen & $\mathbf{0.000}$ drift ($\downarrow$) & $<0.1\text{s}$ \\
 & \textbf{FWSP $\mathcal{O}_1$ (Bit-Flip, 3 seeds)} & $\mathbf{100\%}$ Top-1 loc. & $43.11 \to \mathbf{59.67\%}$ & $100\%$ lossless rec. & $\mathbf{83\text{ms}}$ \\
\midrule
\multirow{8}{*}{\shortstack[l]{\texttt{Z-Image-Turbo} (6B DiT)\\Official \texttt{Z-Erase}~\cite{jiang2026zerase} Suite}} & Unedited Base & $95.09\%$ loc.\ ($92.63\%$ pap.) ASR ($\downarrow$) & $31.75$ COCO CLIP ($\uparrow$) & Ref. & --- \\
 & \texttt{Z-Erase} (30L LoRA, backprop) & $53.33\%$ loc.\ ($24.91\%$ pap.) & $31.31$ COCO CLIP & $13.75\text{ dB}$ PSNR ($\uparrow$) & $\sim 40\text{min}$ \\
 & \textbf{FWSP Prot.\ \texttt{SVD} ($r{=}6$, \texttt{R-A-B})} & $\mathbf{28.77\%}$ ASR ($-24.6\text{ pp}$ loc.) & $\mathbf{31.67}$ COCO CLIP & $\mathbf{19.66\text{ dB}}$ ($+5.9\text{ dB}$) & $\mathbf{91\text{ms}}$ ($>26{,}000\times$) \\
 & \textbf{FWSP Prot.\ \texttt{SVD} ($r{=}4$, 30 Conc.)} & Art $\mathbf{+4.53}$ / Abst $\mathbf{+6.73}$ ($\uparrow$) & Mean $H_a$: $\mathbf{+6.22}$ vs.\ $+4.30$ & Entity $+7.39$ vs.\ $+7.40$ & $\mathbf{90\text{ms}}$ \\
 & \textbf{FWSP Prot.\ \texttt{SVD} ($r{=}4$, Van Gogh 49)} & CLIP $31.99 \to \mathbf{23.20}$ ($\mathbf{-8.79}$) & $\mathbf{31.83}$ COCO CLIP & $\mathbf{21.21\text{ dB}}$ PSNR & $\mathbf{90\text{ms}}$ \\
 & \textbf{FWSP Prot.\ \texttt{SVD} ($r{=}4$, Picasso 49)} & CLIP $28.61 \to \mathbf{17.47}$ ($\mathbf{-11.14}$) & $\mathbf{31.66}$ COCO CLIP & $\mathbf{20.89\text{ dB}}$ PSNR & $\mathbf{90\text{ms}}$ \\
 & \textbf{FWSP $\mathcal{O}_4$ (\texttt{I2P 4.7K} / 50 Celeb)} & $\mathbf{28}$ I2P (vs.\ $161$ \texttt{Z-Erase}) / Celeb $+4.49$ & $30.79$ / $\mathbf{31.69}$ COCO & $\mathbf{18.52\text{ dB}}$ PSNR & $\mathbf{32}$--$\mathbf{51\text{ms}}$ \\
 & \textbf{FWSP $\mathcal{O}_1$ (Bit-Flip, 3 seeds)} & $\mathbf{100\%}$ Top-1 loc. & $34.23 \to \mathbf{34.20}$ CLIP & $39.71 \to \mathbf{41.21\text{ dB}}$ & $\mathbf{2.3\text{ms}}$ \\
\midrule
\multirow{3}{*}{\shortstack[l]{\texttt{Sana-1600M} (Linear DiT)\\Cross-Attn Concept Erasure}} & \texttt{UCE}~\cite{gandikota2024uce} (closed-form) & Target concept erased & Collateral drift & $15.33\text{ dB}$ retain PSNR & $<0.2\text{s}$ \\
 & \texttt{SPEED}~\cite{speed2026nullspace} (null-space) & Target concept erased & Collateral drift & $11.75\text{ dB}$ retain PSNR & $<0.5\text{s}$ \\
 & \textbf{FWSP \texttt{OrthGate} ($r{=}1$, Ours)} & Target concept erased & $\mathbf{+14.7\text{ dB}}$ vs.\ \texttt{UCE} & $\mathbf{30.06\text{ dB}}$ retain PSNR & $\mathbf{<0.1\text{s}}$ \\
\midrule
\multirow{6}{*}{\shortstack[l]{\texttt{Ovis-U1-3B} (UMM)\\Official \texttt{UniKE}~\cite{gao2026unike} ($2{,}966$ subj.)}} & Unedited Base & $0.51\%$ Edit Efficacy ($\uparrow$) & $75.00\%$ Fact Ret.\ ($\uparrow$) & --- & --- \\
 & \texttt{MEMIT} (25-step backprop) & $59.84\%$ Edit Efficacy & --- & Backprop \texttt{compute\_z} & hours \\
 & \texttt{AlphaEdit} (25-step backprop) & $64.73\%$ Edit Efficacy & --- & Backprop \texttt{compute\_z} & hours \\
 & \texttt{PMET} (25-step backprop) & $72.18\%$ Edit Efficacy & --- & Backprop \texttt{compute\_z} & hours \\
 & \textbf{FWSP $\mathcal{O}_4$ (L27, $\lambda{=}40$, Ours)} & $\mathbf{82.20\%}$ ($\mathbf{+10.02\text{ pp}}$ vs.\ \texttt{PMET}) & $\mathbf{72.29\%}$ Fact Ret. & $82.27$ Attr / $82.16$ Rel & $\mathbf{83.5\text{s}}$ (0 backprop) \\
 & \textbf{FWSP $\mathcal{O}_4$ (L27, $\lambda{=}20$, Ours)} & $\mathbf{85.40\%}$ ($\mathbf{+13.22\text{ pp}}$ vs.\ \texttt{PMET}) & $66.87\%$ Fact Ret. & $86.65$ Attr / $84.80$ Rel & $\mathbf{83.7\text{s}}$ (0 backprop) \\
\midrule
\multirow{3}{*}{\shortstack[l]{\texttt{LLaVA-1.5-7B} (VLM)\\Official \texttt{POPE} ($9{,}000$ q.)}} & Unedited Base & $86.97\%$ Pop / $81.43\%$ Adv ($\uparrow$) & $86.27\%$ Macro Acc & $84.72\%$ Macro Prec & $314\text{ms/q}$ \\
 & \texttt{HulluEdit}~\cite{lin2026hulluedit} (runtime hooks) & $87.30\%$ Pop / $81.90\%$ Adv & $\mathbf{86.55\%}$ Macro Acc & $85.17\%$ Macro Prec & $746\text{ms/q}$ (hooks) \\
 & \textbf{FWSP \texttt{PFG} ($g{=}0.35$, 0 hooks)} & $\mathbf{87.60\%}$ Pop / $\mathbf{82.13\%}$ Adv & $86.52\%$ Macro Acc & $\mathbf{85.48\%}$ Macro Prec & $\mathbf{340\text{ms/q}}$ ($\mathbf{2.19\times}$) \\
\midrule
\multirow{6}{*}{\shortstack[l]{\texttt{SD-1.5} \& \texttt{FLUX} 12B\\Official \texttt{SPEED}~\cite{speed2026nullspace} \& \texttt{DP}~\cite{zhang2026doubleproj}}} & \texttt{SD-1.5} \texttt{SPEED} (4 tasks) & $\mathbf{+5.65}$ Mean $H_a$ ($\uparrow$) & $31.18$ COCO CLIP & $23.35\text{ dB}$ PSNR & $1{,}828\text{ms}$ \\
 & \textbf{\texttt{SD-1.5} FWSP $\mathcal{O}_4$ (4 tasks, Ours)} & $+5.61$ Mean $H_a$ (wins $2/4$) & $31.17$ COCO CLIP & $22.85\text{ dB}$ PSNR & $\mathbf{428\text{ms}}$ ($\mathbf{4.3\times}$) \\
 & \texttt{FLUX} 12B \texttt{UCE} (10 cls) & $75.89$ Mean $H_a$ ($\uparrow$) & $99.89\%$ Retain Acc & $24.0\%$ Erased ($\downarrow$) & $474\text{ms}$ \\
 & \texttt{FLUX} 12B \texttt{Double Proj.} (10 cls) & $83.50$ Mean $H_a$ ($\uparrow$) & $100.0\%$ Retain Acc & $16.5\%$ Erased ($\downarrow$) & $451\text{ms}$ \\
 & \textbf{\texttt{FLUX} 12B FWSP Prot.\ $\mathcal{O}_4$ (Ours)} & $\mathbf{90.00}$ ($\mathbf{+6.50\text{ pp}}$ vs.\ \texttt{DP}) & $\mathbf{100.0\%}$ ($7/10$ at $100$) & $\mathbf{10.0\%}$ Erased ($\downarrow$) & $481\text{ms}$ (0 backprop) \\
 & \texttt{SD-1.5} / \texttt{DiT} / \texttt{FLUX} $\mathcal{O}_1$ (Bit-Flip) & $\mathbf{100\%}$ Top-1 loc. & $32.5$ / $27.4$ / $33.5$ CLIP & $34.8$ / $11.7$ / $35.2\text{ dB}$ & $\mathbf{9.9}$--$233\text{ms}$ \\
\bottomrule
\end{tabular}}
\vspace{2pt}
\par\noindent{\footnotesize \textbf{Matched Whitened vs.\ Unwhitened Excision Trade-Off ($r{=}4$):} On official \texttt{Z-Erase} head-to-head suites (\texttt{Z-Image-Turbo}, identical $r{=}4$), covariance whitening ($A{=}\Sigma_{\text{keep}}+\lambda I$) consistently raises retain PSNR at the expense of weaker target suppression compared to Euclidean \texttt{Protected SVD} ($A{=}I$): (1) \texttt{Ring-A-Bell} ($285$): \texttt{Prot.\ SVD} achieves $31.58\%$ ASR / $20.14\text{ dB}$ COCO PSNR vs.\ \texttt{Whitened} $62.81\%$ ASR / $22.66\text{ dB}$; (2) \texttt{Picasso} ($49$): \texttt{Prot.\ SVD} achieves $-11.14$ target CLIP drop / $20.89\text{ dB}$ vs.\ \texttt{Whitened} $-8.12$ drop / $24.12\text{ dB}$; (3) \texttt{Van Gogh} ($49$): \texttt{Prot.\ SVD} achieves $-8.79$ drop / $21.21\text{ dB}$ vs.\ \texttt{Whitened} $-1.75$ drop / $23.89\text{ dB}$. On the smaller 6-prompt retain ablation, whitening raises retain PSNR from $25.66\text{ dB} \to 33.98\text{ dB}$ (\texttt{Z-Image}) and $15.46\text{ dB} \to 17.62\text{ dB}$ (\texttt{SD-1.5}). Thus $A{=}I$ vs.\ $A{=}\Sigma_{\text{keep}}+\lambda I$ traces an explicit erasure--retention Pareto frontier.}
\end{table}

\begin{table}[htbp]
\centering
\caption{Complete fine-grained per-concept and per-category breakdowns: \textbf{(a)} all $30$ official \texttt{Z-Erase}~\cite{jiang2026zerase} Table~3 concepts on \texttt{Z-Image-Turbo 6B} (FWSP Protected SVD $r{=}4$, $0$ backprop, $67$--$169\text{ms}$/concept); \textbf{(b)} all $9$ official \texttt{UniKE}~\cite{gao2026unike} semantic categories on \texttt{Ovis-U1-3B} ($2{,}966$ subjects); \textbf{(c)} \texttt{PFG} 5-draw calibration robustness and label-free Stage-1 gating ($10/10 = 100\%$).}
\label{tab:app_new_sota_breakdowns}
\footnotesize
\setlength{\tabcolsep}{3.5pt}
\resizebox{\textwidth}{!}{%
\begin{tabular}{lccc|lccc|lccc}
\toprule
\multicolumn{12}{l}{\textbf{(a) Complete 30-Concept Breakdown on \texttt{Z-Image-Turbo 6B} (FWSP Prot.\ SVD $r{=}4$, 0 Backprop; Summary vs.\ 9 Baselines in Main Table~\ref{tab:sota_2026_overview}b)}} \\
\multicolumn{4}{c|}{\textbf{Entity (10 Concepts)}} & \multicolumn{4}{c|}{\textbf{Artistic Style (10 Concepts)}} & \multicolumn{4}{c}{\textbf{Abstraction (10 Concepts)}} \\
\textbf{Concept} & \textbf{Acc$_e\downarrow$} & \textbf{Acc$_{ir}\uparrow$} & \textbf{$H_a\uparrow$} & \textbf{Concept} & \textbf{Acc$_e\downarrow$} & \textbf{Acc$_{ir}\uparrow$} & \textbf{$H_a\uparrow$} & \textbf{Concept} & \textbf{Acc$_e\downarrow$} & \textbf{Acc$_{ir}\uparrow$} & \textbf{$H_a\uparrow$} \\
\midrule
Church & $17.61$ & $28.91$ & $\mathbf{+11.30}$ & Claude Monet & $25.61$ & $29.13$ & $+3.52$ & Shake Hand & $19.57$ & $26.66$ & $+7.09$ \\
Tower & $23.32$ & $28.87$ & $+5.55$ & Edvard Munch & $32.67$ & $29.15$ & $-3.52$ & Explosion & $21.46$ & $26.53$ & $+5.07$ \\
Cat & $19.19$ & $29.08$ & $+9.89$ & Vincent Van Gogh & $31.98$ & $28.73$ & $-3.25$ & Amidst & $21.64$ & $28.02$ & $+6.38$ \\
Soccer & $18.96$ & $29.23$ & $+10.27$ & Leonardo da Vinci & $24.00$ & $29.17$ & $+5.17$ & Kiss & $19.52$ & $27.17$ & $+7.65$ \\
Pencil & $32.41$ & $28.58$ & $-3.83$ & Pablo Picasso & $24.87$ & $29.39$ & $+4.52$ & Green & $22.12$ & $27.69$ & $+5.57$ \\
Clownfish & $18.87$ & $28.43$ & $+9.56$ & Rembrandt van Rijn & $22.45$ & $29.78$ & $+7.33$ & Sad & $21.43$ & $27.32$ & $+5.89$ \\
Car & $20.22$ & $29.06$ & $+8.84$ & Nicholas Roerich & $22.10$ & $30.03$ & $+7.93$ & Hug & $20.41$ & $27.22$ & $+6.81$ \\
Pizza & $19.00$ & $28.79$ & $+9.79$ & Salvador Dali & $21.87$ & $29.36$ & $+7.49$ & Origami & $18.06$ & $27.23$ & $\mathbf{+9.17}$ \\
Airplane & $26.71$ & $29.02$ & $+2.31$ & Frida Kahlo & $23.67$ & $28.98$ & $+5.31$ & Time & $20.60$ & $27.85$ & $+7.25$ \\
Shoes & $19.32$ & $29.51$ & $+10.19$ & Gustave Dore & $19.17$ & $30.01$ & $\mathbf{+10.84}$ & Fried & $20.87$ & $27.28$ & $+6.41$ \\
\midrule
\textbf{Category Mean} & $\mathbf{21.56}$ & $\mathbf{28.95}$ & $\mathbf{+7.39}$ & \textbf{Category Mean} & $\mathbf{24.84}$ & $\mathbf{29.37}$ & $\mathbf{+4.53}$ & \textbf{Category Mean} & $\mathbf{20.57}$ & $\mathbf{27.30}$ & $\mathbf{+6.73}$ \\
\emph{vs.\ \texttt{Z-Erase} LoRA} & $23.1$ & --- & $+7.4$ & \emph{vs.\ \texttt{Z-Erase} LoRA} & $26.1$ & --- & $+0.8$ & \emph{vs.\ \texttt{Z-Erase} LoRA} & $24.3$ & --- & $+4.7$ \\
\bottomrule
\end{tabular}}

\vspace{4pt}
\resizebox{\textwidth}{!}{%
\begin{tabular}{lcccc|lcccc}
\toprule
\multicolumn{5}{l|}{\textbf{(b) \texttt{UniKE}~\cite{gao2026unike} 9-Category Breakdown (\texttt{Ovis-U1-3B}, $N{=}2{,}966$)}} & \multicolumn{5}{l}{\textbf{(c) \texttt{UniKE} Multi-Hop Boundary \& \texttt{PFG} Robustness Diagnostics}} \\
\textbf{Category ($N$)} & \textbf{Base} & \textbf{FWSP $\lambda{=}40$} & \textbf{FWSP $\lambda{=}20$} & \textbf{Gain ($\lambda{=}20$)} & \textbf{Diagnostic / Split} & \textbf{Setting} & \textbf{Baseline} & \textbf{FWSP (Ours)} & \textbf{Verdict} \\
\midrule
Pattern ($58$) & $0.00\%$ & $\mathbf{96.55\%}$ & $\mathbf{96.55\%}$ & $+96.55\text{ pp}$ & \texttt{UniKE} Attr.\ \texttt{stage\_1} & $959$ subj. & $0.00\%$ Base & $\mathbf{82.27\%}$ ($\lambda{=}40$) & Direct edit \\
Shape ($195$) & $0.00\%$ & $92.82\%$ & $\mathbf{93.85\%}$ & $+93.85\text{ pp}$ & \texttt{UniKE} Attr.\ \texttt{stage\_n} & $959$ subj. & $2.82\%$ Base & $\mathbf{82.69\%}$ ($\lambda{=}40$) & Visual prompt \\
Creator ($328$) & $1.22\%$ & $91.16\%$ & $\mathbf{92.07\%}$ & $+90.85\text{ pp}$ & \texttt{UniKE} Rel.\ \texttt{stage\_1} & $2{,}007$ subj. & $0.75\%$ Base & $\mathbf{82.16\%}$ ($\lambda{=}40$) & Direct edit \\
Color ($285$) & $0.00\%$ & $89.12\%$ & $\mathbf{91.93\%}$ & $+91.93\text{ pp}$ & \texttt{UniKE} Multi-Hop \texttt{s2/s3/s4} & $832$--$872$ & $2.1/1.4/1.3\%$ & $1.26 / 1.29 / 0.12\%$ & R5 span limit \\
Occupation ($531$) & $0.38\%$ & $\mathbf{86.06\%}$ & $\mathbf{86.06\%}$ & $+85.68\text{ pp}$ & \texttt{UniKE} 4-Layer (\texttt{L24--27}) & $\lambda{=}50$ & $75.0\%$ ret. & $1.87\%$ ret. & Over-edit collapse \\
\cmidrule(lr){6-10}
Affiliation ($303$) & $2.31\%$ & $85.81\%$ & $\mathbf{89.77\%}$ & $+87.46\text{ pp}$ & \texttt{PFG} ($N{=}8$, $q{=}0.03$) & 5 draws & Raw HE $58.7{\pm}23.9$ & GSM $\mathbf{87.2{\pm}1.2}$ & HE $\mathbf{75.4{\pm}9.4}$ \\
Material ($212$) & $0.00\%$ & $78.30\%$ & $\mathbf{81.13\%}$ & $+81.13\text{ pp}$ & \texttt{PFG} ($N{=}16$, $q{=}0.03$) & 5 draws & Tighter SD & GSM $\mathbf{86.2{\pm}2.4}$ & HE $\mathbf{76.0{\pm}5.5}$ \\
Location ($845$) & $0.24\%$ & $74.91\%$ & $\mathbf{79.41\%}$ & $+79.17\text{ pp}$ & Stage-1 Trunc.\ Gate & 10 regimes & Concise $0$--$0.8\%$ & Long-tail $18.9$--$95.4\%$ & $\mathbf{10/10}$ ($100\%$) \\
Size ($209$) & $0.00\%$ & $63.16\%$ & $\mathbf{75.60\%}$ & $+75.60\text{ pp}$ & \textbf{\texttt{UniKE} Overall ($2{,}966$)} & \texttt{L27}, $83.5\text{s}$ & \texttt{PMET} $72.18\%$ & $\mathbf{82.20\%}$ / $\mathbf{85.40\%}$ & $\mathbf{+10.0 / +13.2\text{ pp}}$ \\
\bottomrule
\end{tabular}}
\end{table}

\begin{table}[htbp]
\centering
\caption{Strict incremental multi-teacher component ablation on the unmodified official \texttt{IM-MOPD} 1,024-query 4-domain $32\text{K}$ validation benchmark (\texttt{Qwen3-4B-OT3}, $256$/domain, $n{=}1$; official $8{,}034$-query $32\text{K}$ test results in Table~\ref{tab:fwsp_mopd}a), cleanly isolating Covariance ($\Sigma_k$), Trajectory Curation (Eq.~\eqref{eq:curation_score}), Response-Token Weighting ($w_t \in \{0, 0.5, 2.0\}$), and Base-Centered Regularization ($\lambda{=}0.01\bar{d}$ vs.\ $10^{-4}\bar{d}$ jitter) from \texttt{Task Arithmetic} and stabilized \texttt{RegMean}.}
\label{tab:app_immopd_val}
\scriptsize
\setlength{\tabcolsep}{3.6pt}
\resizebox{\textwidth}{!}{%
\begin{tabular}{lcccccccccc}
\toprule
\textbf{Variant (All 6 Rows Evaluated Under Identical Official 32K Val Pool)} & \textbf{Cov} & \textbf{Cur} & \textbf{TokW} & \textbf{BaseReg} & \textbf{Med} & \textbf{Law} & \textbf{Fin} & \textbf{IF} & \textbf{Macro Avg} & \textbf{Oracle Rec.} \\
\midrule
\texttt{SFT Base} (\texttt{Qwen3-4B-OT3-2ep}) & --- & --- & --- & --- & $69.14$ & $62.89$ & $71.48$ & $5.86$ & $52.34\%$ & $0.0\%$ \\
4 Specialist RL Teachers (Oracle Envelope) & --- & --- & --- & --- & $94.79$ & $74.61$ & $84.90$ & $70.05$ & $81.09\%$ & $100.0\%$ \\
\midrule
Row~1: \texttt{Task Arithmetic} (Uniform Merge, $\alpha_k{=}1$) & $\times$ & $\times$ & $\times$ & $\times$ & $77.73$ & $68.75$ & $78.91$ & $27.73$ & $63.28\%$ & $38.1\%$ \\
Row~2: \texttt{RegMean} (Stabilized, 64 Rand.\ Traj., $w_t{=}1$, $\lambda{=}10^{-4}\bar{d}$) & $\checkmark$ & $\times$ & $\times$ & $\times$ & $89.45$ & $70.31$ & $82.81$ & $66.41$ & $77.25\%$ & $86.6\%$ \\
Row~3: \texttt{FWSP-A} ($+$Eq.~\eqref{eq:curation_score} Top-64 Curation, $w_t{=}1$) & $\checkmark$ & $\checkmark$ & $\times$ & $\times$ & $86.33$ & $69.53$ & $82.42$ & $\mathbf{71.09}$ & $77.34\%$ & $87.0\%$ \\
Row~4: \texttt{FWSP-B} ($+$Response Token Weights $w_t$, unreg.\ $\lambda{=}10^{-4}\bar{d}$) & $\checkmark$ & $\checkmark$ & $\checkmark$ & $\times$ & $89.06$ & $71.88$ & $83.20$ & $64.84$ & $77.25\%$ & $86.6\%$ \\
Row~5: \textbf{Full FWSP} ($+$Base-Centered Reg.\ $\lambda{=}0.01\bar{d}$, $\alpha_k{=}1$) & $\checkmark$ & $\checkmark$ & $\checkmark$ & $\checkmark$ & $89.06$ & $70.31$ & $83.20$ & $\underline{69.14}$ & $\underline{77.93\%}$ & $\underline{89.0\%}$ \\
Row~5b: \textbf{Full FWSP} ($+$Base-Centered Reg., $\alpha_{\text{if}}{=}1.5$) & $\checkmark$ & $\checkmark$ & $\checkmark$ & $\checkmark$ & $\mathbf{90.23}$ & $\mathbf{72.27}$ & $\mathbf{83.98}$ & $68.75$ & $\mathbf{78.81\%}$ & $\mathbf{92.1\%}$ \\
\bottomrule
\end{tabular}}
\end{table}

\begin{figure}[htbp]
    \centering
    \includegraphics[width=0.92\textwidth]{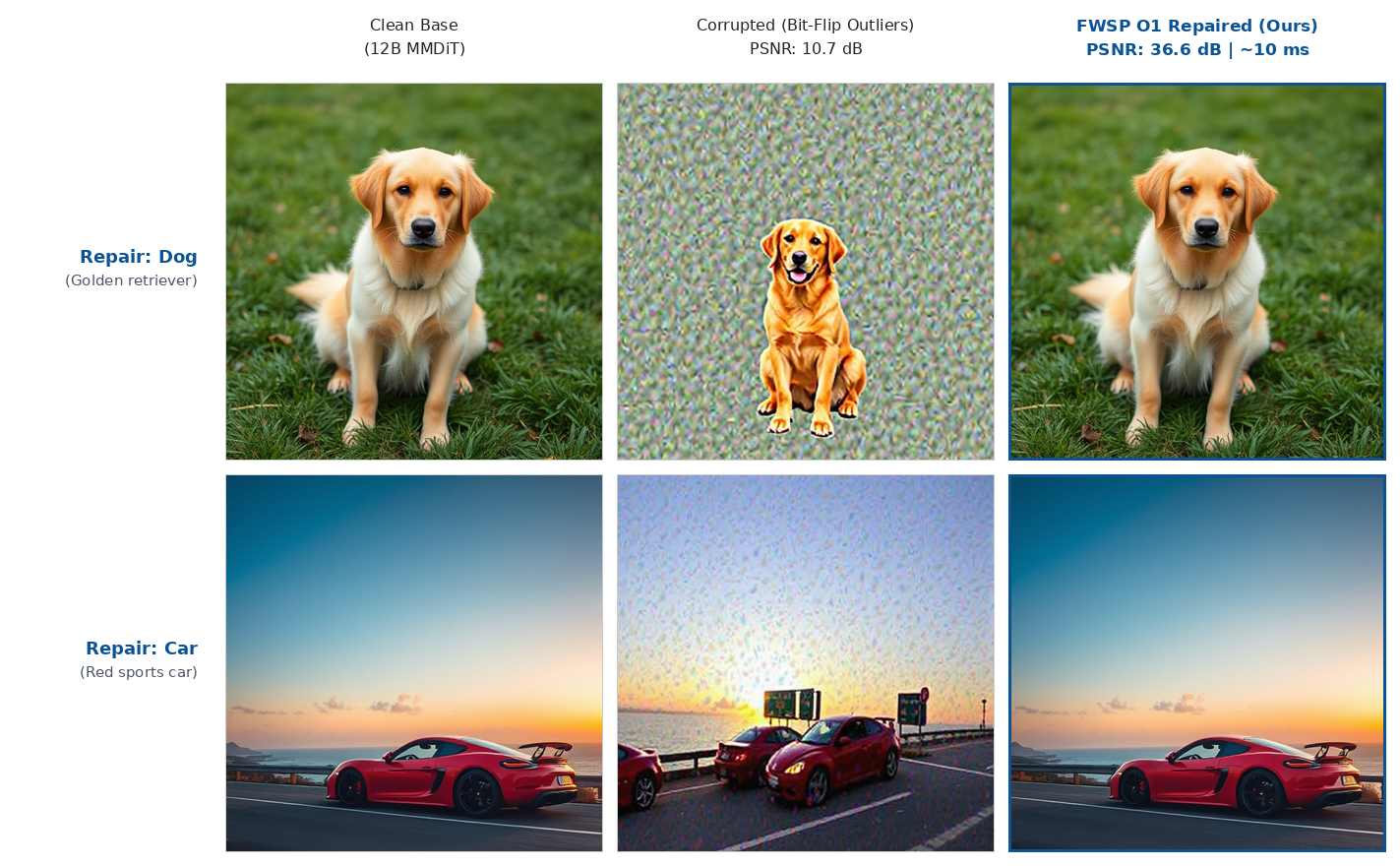}
    \caption{Forward-only fault self-repair on \texttt{FLUX.1-schnell} (12B MMDiT). Under 192 finite bit-flip outliers ($\pm 65$) in \texttt{transformer\_blocks.0.attn.to\_v}, output degrades ($10.7\text{dB}$ PSNR); FWSP $\mathcal{O}_1$ localizes the matrix ($100\%$ Top-1) and restores fidelity ($36.6\text{dB}$) in $\sim 10\text{ms}$ target-layer repair.}
    \label{fig:app_flux_qual}
\end{figure}

\begin{table}[htbp]
\centering
\caption{Pluggable non-\texttt{lm\_head} FWSP operators on \texttt{Qwen3.8-27B} across competition math (\texttt{AIME 2024--2026} 16K greedy and official 32K stochastic $n{=}4$) and recent coding split (\texttt{LiveCodeBench} Feb--Apr 2025, $N{=}61$, single-pass 16K + offline 4K/8K/16K slicing), all with zero inference harness and $0.0\%$ \texttt{multi-think}.}
\label{tab:app_aime_16k}
\footnotesize
\setlength{\tabcolsep}{4.5pt}
\resizebox{\textwidth}{!}{%
\begin{tabular}{lcccccc}
\toprule
\textbf{FWSP Plugin (\texttt{Qwen3.8-27B}, 16K AIME 24--26 Greedy)} & \textbf{Combined (90)} & \textbf{AIME 24} & \textbf{AIME 25} & \textbf{AIME 26} & \textbf{Mean Tok} & \textbf{$\Delta$ vs.\ Base} \\
\midrule
Unedited Base & $63.33\%$ ($57$) & $66.7\%$ & $63.3\%$ & $60.0\%$ & $10{,}162$ & --- \\
Negative Control: $\mathcal{O}_3$ \texttt{dt\_bias} ($-0.20$) & $63.33\%$ ($57$) & $66.7\%$ & $53.3\%$ & $70.0\%$ & $9{,}922$ & $+0.00$ pp \\
Mid-Layer Contrastive Steering (\texttt{L28/35}, $\eta{=}0.25$) & $66.67\%$ ($60$) & $73.3\%$ & $66.7\%$ & $60.0\%$ & $10{,}212$ & $+3.33$ pp \\
Mid-Layer Protected $\mathcal{O}_2$ (\texttt{L28/35}, $\alpha{=}0.25$) & $68.89\%$ ($62$) & $70.0\%$ & $66.7\%$ & $70.0\%$ & $10{,}147$ & $+5.56$ pp \\
\textbf{Mid-Layer Protected $\mathcal{O}_2$ (\texttt{L28/35}, $\alpha{=}0.50$)} & $\mathbf{71.11\%}$ ($64$) & $70.0\%$ & $70.0\%$ & $\mathbf{73.3\%}$ & $10{,}348$ & $\mathbf{+7.78}$ pp \\
\textbf{$\mathcal{O}_3$ \texttt{dt\_bias} ($+0.35$, 48 GDN layers)} & $\mathbf{72.22\%}$ ($65$) & $76.7\%$ & $\mathbf{80.0\%}$ & $60.0\%$ & $10{,}687$ & $\mathbf{+8.89}$ pp \\
\textbf{$\mathcal{O}_3$ \texttt{dt\_bias} ($+0.20$, 48 GDN layers)} & $\mathbf{72.22\%}$ ($65$) & $\mathbf{80.0\%}$ & $70.0\%$ & $66.7\%$ & $10{,}363$ & $\mathbf{+8.89}$ pp \\
\midrule
\textbf{Official 32K \texttt{AIME 2026} ($T{=}0.6, n{=}4$)} & \textbf{avg@4} & \textbf{maj@4} & \textbf{pass@4} & \textbf{Trunc} & \textbf{Mean Tok} & \textbf{$\Delta$ avg@4} \\
\midrule
Unedited Base (32K) & $84.17\%$ & $96.67\%$ & $96.67\%$ & $20.00\%$ & $16{,}525$ & --- \\
$\mathcal{O}_3$ \texttt{dt\_bias} ($-0.15$) & $85.00\%$ & $96.67\%$ & $93.33\%$ & $22.50\%$ & $\mathbf{15{,}450}$ & $+0.83$ pp \\
$\mathcal{O}_3$ \texttt{dt\_bias} ($+0.20$) & $80.00\%$ & $96.67\%$ & $96.67\%$ & $25.83\%$ & $17{,}896$ & $-4.17$ pp \\
\textbf{$\mathcal{O}_3$ \texttt{dt\_bias} ($+0.35$)} & $\mathbf{85.83\%}$ & $93.33\%$ & $93.33\%$ & $\mathbf{18.33\%}$ & $16{,}562$ & $\mathbf{+1.66}$ pp \\
\midrule
\textbf{Official 32K Frontier Olympiads (\texttt{Qwen3.8-27B}, $T{=}0.6, n{=}4$)} & \textbf{avg@4 ($\uparrow$)} & \textbf{maj@4 ($\uparrow$)} & \textbf{pass@4 ($\uparrow$)} & \textbf{Trunc ($\downarrow$)} & \textbf{Mean Tok} & \textbf{$\Delta$ avg@4} \\
\midrule
\texttt{HMMT Feb 2025} ($N{=}30$) --- Unedited Base & $74.17\%$ & $83.3\%$ & $86.7\%$ & $29.2\%$ & $19{,}464$ & --- \\
\texttt{HMMT Feb 2025} --- $\mathcal{O}_3$ \texttt{dt\_bias} ($-0.15$) & $77.50\%$ & $\mathbf{93.3\%}$ & $\mathbf{93.3\%}$ & $\mathbf{23.3\%}$ & $\mathbf{18{,}217}$ & $+3.33$ pp \\
\textbf{\texttt{HMMT Feb 2025} --- $\mathcal{O}_3$ \texttt{dt\_bias} ($+0.35$, Ours)} & $\mathbf{79.17\%}$ & $86.7\%$ & $\mathbf{93.3\%}$ & $28.3\%$ & $19{,}788$ & $\mathbf{+5.00}$ pp \\
\midrule
\texttt{HMMT Nov 2025} ($N{=}30$) --- Unedited Base & $73.33\%$ & $83.3\%$ & $\mathbf{90.0\%}$ & $25.8\%$ & $17{,}615$ & --- \\
\texttt{HMMT Nov 2025} --- $\mathcal{O}_3$ \texttt{dt\_bias} ($-0.15$) & $70.83\%$ & $83.3\%$ & $83.3\%$ & $24.2\%$ & $\mathbf{16{,}965}$ & $-2.50$ pp \\
\textbf{\texttt{HMMT Nov 2025} --- $\mathcal{O}_3$ \texttt{dt\_bias} ($+0.35$, Ours)} & $\mathbf{75.00\%}$ & $83.3\%$ & $86.7\%$ & $\mathbf{23.3\%}$ & $18{,}531$ & $\mathbf{+1.67}$ pp \\
\midrule
\texttt{CMIMC 2025} ($N{=}40$) --- Unedited Base & $61.25\%$ & $\mathbf{75.0\%}$ & $77.5\%$ & $38.1\%$ & $21{,}772$ & --- \\
\texttt{CMIMC 2025} --- $\mathcal{O}_3$ \texttt{dt\_bias} ($-0.15$) & $58.13\%$ & $72.5\%$ & $75.0\%$ & $\mathbf{33.8\%}$ & $\mathbf{21{,}149}$ & $-3.12$ pp \\
\textbf{\texttt{CMIMC 2025} --- $\mathcal{O}_3$ \texttt{dt\_bias} ($+0.35$, Ours)} & $\mathbf{63.75\%}$ & $72.5\%$ & $77.5\%$ & $\mathbf{33.8\%}$ & $22{,}502$ & $\mathbf{+2.50}$ pp \\
\midrule
\texttt{APEX 2025} ($N{=}12$) --- Unedited Base & $0.00\%$ & $0.0\%$ & $0.0\%$ & $95.8\%$ & $31{,}579$ & --- \\
\texttt{APEX 2025} --- $\mathcal{O}_3$ \texttt{dt\_bias} ($-0.15$) & $0.00\%$ & $0.0\%$ & $0.0\%$ & $95.8\%$ & $31{,}664$ & $+0.00$ pp \\
\textbf{\texttt{APEX 2025} --- $\mathcal{O}_3$ \texttt{dt\_bias} ($+0.35$, Ours)} & $\mathbf{2.08\%}$ & $\mathbf{8.3\%}$ & $\mathbf{8.3\%}$ & $\mathbf{89.6\%}$ & $\mathbf{30{,}663}$ & $\mathbf{+2.08}$ pp \\
\midrule
\textbf{5-Suite 32K Macro (incl.\ \texttt{AIME 2026}, $142$ probs): Base $\to$ Ours} & $58.58 \to \mathbf{61.17\%}$ & $67.7 \to \mathbf{68.8\%}$ & $70.2 \to \mathbf{71.8\%}$ & $41.8 \to \mathbf{38.7\%}$ & $21{,}391 \to 21{,}609$ & $\mathbf{+2.59}$ pp ($\mathbf{5/5}$) \\
\midrule
\textbf{\texttt{Flash-Next-FP8} (186GB MoE) 32K \texttt{AIME 2026} Ceiling Control ($T{=}0.6, n{=}4$)} & \textbf{avg@4} & \textbf{maj@4} & \textbf{pass@4} & \textbf{Trunc} & \textbf{Mean Tok} & \textbf{$\Delta$ avg@4} \\
\midrule
\texttt{AIME 2026} ($N{=}30$) --- Unedited Base (near-saturated) & $\mathbf{88.33\%}$ & $\mathbf{96.67\%}$ & $\mathbf{96.67\%}$ & $18.33\%$ & $15{,}343$ & --- \\
\texttt{AIME 2026} --- $\mathcal{O}_3$ \texttt{dt\_bias} ($-0.15$) & $86.67\%$ & $\mathbf{96.67\%}$ & $\mathbf{96.67\%}$ & $\mathbf{16.67\%}$ & $\mathbf{14{,}489}$ & $-1.66$ pp \\
\texttt{AIME 2026} --- $\mathcal{O}_3$ \texttt{dt\_bias} ($+0.20$; Stage-4 rejected) & $77.50\%$ & $90.00\%$ & $83.33\%$ & $27.50\%$ & $16{,}540$ & $-10.83$ pp \\
\texttt{AIME 2026} --- $\mathcal{O}_3$ \texttt{dt\_bias} ($+0.35$) & $\mathbf{88.33\%}$ & $\mathbf{96.67\%}$ & $\mathbf{96.67\%}$ & $18.33\%$ & $16{,}970$ & $+0.00$ pp \\
\midrule
\textbf{\texttt{LiveCodeBench} Latest ($N{=}61$, 16K Slicing)} & \textbf{4K pass@1} & \textbf{8K pass@1} & \textbf{16K pass@1} & \textbf{16K Passed} & \textbf{16K Tok} & \textbf{$\Delta$ 16K} \\
\midrule
Unedited Base & $31.15\%$ & $42.62\%$ & $47.54\%$ & $29/61$ & $10{,}721$ & --- \\
\textbf{$\mathcal{O}_3$ \texttt{dt\_bias} ($+0.35$)} & $31.15\%$ & $\mathbf{44.26\%}$ & $49.18\%$ & $30/61$ & $10{,}838$ & $+1.64$ pp \\
\textbf{$\mathcal{O}_3$ \texttt{dt\_bias} ($+0.20$)} & $31.15\%$ & $36.07\%$ & $\mathbf{50.82\%}$ & $\mathbf{31/61}$ & $11{,}229$ & $\mathbf{+3.28}$ pp \\
\bottomrule
\end{tabular}}
\end{table}

\begin{table}[htbp]
\centering
\caption{Qualitative reasoning case study on \texttt{DeepSeek-R1-0528-Qwen3-8B} using the frozen math-calibrated \texttt{Protected FisherGate} patch (zero inference harness, greedy decoding). Top: \texttt{MATH-500} series problem ($B{=}1{,}408$, gold $p - q$). Bottom: zero-shot Math$\to$\texttt{HumanEval} transfer (\texttt{truncate\_number}, $B{=}1{,}024$).}
\label{tab:app_reasoning_case}
\scriptsize
\resizebox{\textwidth}{!}{%
\begin{tabular}{p{2.6cm}p{10.6cm}}
\toprule
\textbf{Setting / Model} & \textbf{Verbatim Output Excerpt, Token Count, and Verdict} \\
\midrule
\texttt{MATH-500} (Gold: $p{-}q$) & \emph{Prompt:} Define $p = \sum_{k=1}^\infty 1/k^2$, $q = \sum_{k=1}^\infty 1/k^3$. Write $\sum_{j=1}^\infty \sum_{k=1}^\infty 1/(j{+}k)^3$ in terms of $p, q$. \\
Unedited Base & \emph{``...$(2,1)\; j{+}k{=}3, 1/27 \dots$ Then $p{-}q = (1/1^2{-}1/1^3) + (1/2^2{-}1/2^3) + (1/3^2{-}1/3^3\dots$''} \textcolor{red}{(1,408 tok, truncated inside \texttt{<think>}, FAIL)} \\
\textbf{\texttt{Protected PFG} (Ours)} & \emph{``...$\sum_{m=2}^\infty \frac{m-1}{m^3} = (p{-}1) - (q{-}1) = p - q$. Therefore, the double sum is $p - q$. $\boxed{p - q}$''} \textcolor{blue}{(806 tok, $-42.8\%$ tokens, PASS)} \\
\midrule
Zero-Shot \texttt{HumanEval} & \emph{Prompt:} Complete \texttt{truncate\_number(number: float) -> float} (exact same math-only patch, zero code calibration). \\
Unedited Base & Correct \texttt{number \% 1.0} implementation after extended chain-of-thought self-verification (726 tok, PASS). \\
\textbf{\texttt{Protected PFG} (Ours)} & Cleanly closes \texttt{</think>} and emits concise passing \texttt{number - int(number)} solution \textcolor{blue}{(446 tok, $-38.6\%$ tokens, PASS)}. \\
\bottomrule
\end{tabular}}
\end{table}

\begin{table}[htbp]
\centering
\caption{Official bidirectional LLM safety and refusal-control benchmark on \texttt{Qwen3-4B-Instruct-2507} across held-out \texttt{AdvBench} ($N{=}456$, disjoint from 64 calibration prompts) and full \texttt{XSTest} ($N{=}450$: $250$ safe-contrastive $+$ $200$ unsafe, $0\%$ calibration overlap), with $0$ backpropagation, $0$ runtime hooks, and $1.0\text{s}$ calibration.}
\label{tab:llm_semantic_edit}
\footnotesize
\resizebox{\textwidth}{!}{%
\begin{tabular}{lccccc}
\toprule
\textbf{Method (Permanent Weight Edit, 0 Backprop, 0 Hooks)} & \textbf{\texttt{AdvBench} Refusal (ASR)} & \textbf{\texttt{XSTest}-Unsafe Ref.} & \textbf{\texttt{XSTest}-Safe False Ref.\ ($\downarrow$)} & \textbf{Safety-Help.\ HM} & \textbf{WikiText PPL ($\downarrow$)} \\
\midrule
Unedited Base (\texttt{Qwen3-4B-Instruct-2507}) & $99.12\%$ ($0.88\%$) & $77.00\%$ & $5.20\%$ ($94.8\%$ comp.) & $\mathbf{96.91}$ & $5.881$ \\
\texttt{Arditi et al.}~\yrcite{arditi2024refusal} (Unprotected Ablation) & $46.27\%$ ($53.73\%$) & $10.50\%$ & $0.00\%$ ($100\%$ comp.) & $63.27$ & $6.298$ ($+0.417$) \\
\textbf{FWSP $\mathcal{O}_2$ Protected Excision} ($Q_{\text{keep}}^\top U{=}0$, audit) & $29.39\%$ ($70.61\%$ ASR) & $3.50\%$ & $0.00\%$ ($100\%$ comp.) & $45.42$ & $5.891$ ($+0.010$ PPL) \\
\textbf{FWSP $\mathcal{O}_2$ Over-Refusal Cal.} ($\alpha{=}0.45$) & $94.30\%$ ($5.70\%$) & $41.00\%$ & $\mathbf{0.40\%}$ ($13\times$ fewer) & \underline{$96.88$} & $5.917$ ($+0.036$) \\
\textbf{FWSP $\mathcal{O}_2$ Over-Refusal Cal.} ($\alpha{=}0.65$) & $82.89\%$ ($17.11\%$) & $20.00\%$ & $\mathbf{0.00\%}$ ($100\%$ comp.) & $90.65$ & $5.903$ ($+0.022$) \\
\textbf{FWSP $\mathcal{O}_4$ Safety Hardening} (Reverse, L19) & $\mathbf{99.78\%}$ ($\mathbf{0.22\%}$ ASR) & $\mathbf{84.00\%}$ ($+7.00\text{ pp}$) & $9.60\%$ ($90.4\%$ comp.) & $94.86$ & $\mathbf{5.873}$ ($\mathbf{-0.008}$) \\
\bottomrule
\end{tabular}}
\vspace{2pt}
\par\noindent{\footnotesize Note: $\mathrm{Safety\text{-}Help.\ HM} = \frac{2 \cdot \mathrm{AdvBench\text{-}Ref} \cdot \mathrm{XSTest\text{-}SafeComp}}{\mathrm{AdvBench\text{-}Ref} + \mathrm{XSTest\text{-}SafeComp}}$. Refusal-excision rows are included strictly as mechanistic safety-circuit diagnostics (and trade off refusal rate vs.\ compliance); WikiText PPL reflects surface fluency only rather than full downstream capability preservation.}
\end{table}

\begin{table}[htbp]
\centering
\caption{Five-order-of-magnitude parameter scaling of deterministic FWSP $\mathcal{O}_1$ (50K to 12B).}
\label{tab:app_scale}
\footnotesize
\setlength{\tabcolsep}{4.5pt}
\resizebox{\textwidth}{!}{%
\begin{tabular}{lccccc}
\toprule
\textbf{Model Architecture} & \textbf{Params (M)} & \textbf{2D Matrices} & \textbf{Top-1 Loc.} & \textbf{Repair Time (ms)} & \textbf{FPE} \\
\midrule
\texttt{MLP Control Policy} & $0.05$ & $4$ & $100.0\%$ & $4.2$ & $40$ \\
\texttt{ViT-Small} (Vision) & $2.71$ & $26$ & $100.0\%$ & $19.5$ & $192$ \\
\texttt{ResNet-18} (CNN) & $4.49$ & $21$ & $100.0\%$ & $24.8$ & $192$ \\
\texttt{DINOv2-Base} ($224\times 224$) & $86.58$ & $74$ & $100.0\%$ & $142.0$ & $128$ \\
\texttt{SigLIP 2 SO400M} (Vision SOTA) & $400.0$ & $164$ & $100.0\%$ & $83.1$ & $64$ \\
\texttt{Z-Image-Turbo} (6B DiT) & $6{,}000.0$ & $224$ & $100.0\%$ & $2.3^*$ & $4$ \\
\texttt{LLaDA-8B-Base} (Diffusion LM) & $8{,}015.6$ & $226$ & $100.0\%$ & $3{,}206.3$ & $32$ \\
\texttt{Qwen3-8B} (36L Mainstream LM) & $8{,}190.7$ & $254$ & $100.0\%$ & $2{,}666.9$ & $16$ \\
\texttt{FLUX.1-schnell} (12B MMDiT) & $12{,}000.0$ & $304$ & $100.0\%$ & $9.9^*$ & $4$ \\
\bottomrule
\end{tabular}}
\vspace{2pt}
\par\noindent{\footnotesize $^*$Target-layer $\mathcal{O}_1$ repair time after static Stage-1 localization; all other rows report end-to-end localization plus repair time.}
\end{table}

\FloatBarrier

\section{Reproducibility, Practitioner Decision Guide, and Operational Comparison}
\label{app:repro}
Table~\ref{tab:decision} provides a practitioner decision guide mapping deployment regimes to FWSP operators, hybrid FWSP + \texttt{LoRA}, or full backpropagation, and Table~\ref{tab:llm_vs_fwsp} summarizes the operational differences between prior LLM-agent weight-space programming approaches and deterministic FWSP.

\begin{table}[htbp]
\centering
\caption{Practitioner decision guide: when to use pure FWSP, hybrid FWSP + LoRA, or full backpropagation.}
\label{tab:decision}
\small
\resizebox{\textwidth}{!}{%
\begin{tabular}{p{3.8cm}p{2.7cm}p{6.8cm}}
\toprule
\textbf{Deployment Scenario} & \textbf{Recommended Tool} & \textbf{Empirical Basis and Expected Benefit} \\
\midrule
Bit-flips, row outliers, corrupt layers & FWSP $\mathcal{O}_1$ & 100\% Top-1 loc.\ up to 12B (2.67s on 8B LM, 10ms on 12B DiT), zero gradients, works with physics residuals. \\
Localized backdoors / static MLP bias & FWSP $\mathcal{O}_2$ (+Align) & $100\% \to 0\%$ ASR on \texttt{DINOv2} and \texttt{CartPole} $9 \to 200$ using unlabeled pairs; avoid early CNNs and shared induction heads. \\
Systematic shift / BN-CNN / BN-free ViT & FWSP $\mathcal{O}_3$ & Matches \texttt{BN-Adapt} on CNNs; 15$\times$ cheaper than \texttt{Tent} on Frost/Impulse; verify on high-freq noise. \\
Reasoning \texttt{<think>} overthinking loops & FWSP \texttt{PFG} (+ $\mathcal{O}_3$ on GDN) & Canonical \texttt{Protected FisherGate} enables single-checkpoint Math$\to$Code and HF$\to$vLLM transfer under constrained compute; pair with $\mathcal{O}_3$ \texttt{dt\_bias} on hybrid GDN; Stage-4 abstains on concise or multi-token BPE regimes. \\
New classes, continual head, LM readout & FWSP $\mathcal{O}_4$ & 16--90$\times$ less FPE than SGD ($B^*\!\approx\!4{,}096$); 3.2$\times$ cheaper per query than 4-shot ICL; 6.4$\times$ less VRAM. \\
Multi-teacher capability consolidation & FWSP $\mathcal{O}_4$ (Curated) & Shared-base layer-wise Gram merging + trajectory curation ($70.30\%$ 4-domain official avg, $+9.37$ pp over TA, 0 backprop). \\
Compound fault + shift / nonlinear / facts & FWSP + LoRA / FT & Use FWSP Stage~1 \& 3 to localize and repair, then \texttt{LoRA} (88.1\% vs.\ 8.6\% blind [15.3\% at 50 steps], 10.7$\times$ fewer params). \\
\bottomrule
\end{tabular}}
\end{table}

\begin{table}[htbp]
\centering
\caption{Operational comparison and complementary division of labor between LLM-agent Weight-Space Programming and deterministic FWSP.}
\label{tab:llm_vs_fwsp}
\footnotesize
\setlength{\tabcolsep}{4.5pt}
\resizebox{\textwidth}{!}{%
\begin{tabular}{lcc}
\toprule
\textbf{Dimension} & \textbf{LLM-Agent WSP~\cite{takehi2026wsp,fan2026selfrewriting,moreira2026openwsp,marasini2026weightbench}} & \textbf{FWSP (Ours)} \\
\midrule
Role in WSP Stack & High-level planner / symbolic script synthesis & Verified deterministic analytic compilation layer \\
Optimizer LLM Required & Frontier or 27B LLM + Python sandbox & None ($0$ optimizer LLM tokens; callable by LLM planners) \\
Determinism & Stochastic (prone to script bugs / zero-score collapses) & Deterministic (fixed inputs, config \& seed) \\
Self-Repair Latency (8B LM) & Multi-turn coding loop (minutes) & 2.67s (\texttt{Qwen3-8B}) / 3.21s (\texttt{LLaDA-8B}) \\
Modality Coverage & Language models & 7 modalities (Vision, Diffusion/DiT, VLM/UMM, RL, PDE, LMs) \\
Verification and Rollback & Manual / ad-hoc & Stage-4 paired verifier with retention guard \\
\bottomrule
\end{tabular}}
\end{table}

\paragraph{Supplementary Controlled Blind-Routing Ablations (Tables~\ref{tab:agent_vs_fwsp}--\ref{tab:app_120trials_breakdown}).}
To complement our primary third-party \texttt{OpenWSP} and \texttt{WeightBench} results in Main Text Table~\ref{tab:external_wsp_benchmarks}, Tables~\ref{tab:agent_vs_fwsp} and~\ref{tab:app_120trials_breakdown} preserve our controlled internal routing ablations comparing deterministic Algorithm~\ref{alg:blind} against local 8B/27B sandbox agents, three live 2026 B200 frontier reasoning endpoints (\texttt{GLM-5.3 Flash}, \texttt{Kimi K3.1}, \texttt{DeepSeek V4.1 Flash}) across $120$ cross-modal blind trials (and the $90$-trial discriminative subset excluding saturated \texttt{Ovis-U1-3B}), live \texttt{Codex-CLI} coding agents on $8$ corrupted 2026 tasks, and $36$ out-of-rule \emph{Unseen Fault Family} trials never used in rule design. Because these diagnostic suites share FWSP's $\mathcal{O}_1$--$\mathcal{O}_4$ action vocabulary, we treat them as controlled routing ablations in the Supplement rather than headline external benchmarks.

\begin{table}[htbp]
\centering
\caption{Supplementary controlled blind-routing ablations: LLM agents vs.\ deterministic LLM-free FWSP across four internal protocols: (a) local Python-sandbox \& 18-trial \texttt{DINOv2} tool-router pilot; (b) \textbf{120-trial cross-modal blind dispatch} (4 $\times$ 2026 SOTA models $\times$ 10 regimes $\times$ 3 seeds) plus the post-hoc \textbf{90-trial discriminative subset} excluding low-discrimination \texttt{Ovis-U1-3B}, against three live B200 frontier reasoning agents under identical Stage-1 telemetry, $\mathcal{O}_1$--$\mathcal{O}_4$ tools, and 4-turn feedback; (c) \textbf{\texttt{Codex-CLI}-based coding agents} (open-source \texttt{codex} CLI harness backed by third-party B200 models via \texttt{/v1/responses}) on 8 live corrupted 2026 tasks; and (d) \textbf{36-trial Unseen Fault Families} (4 recoverable unseen fault types + 2 out-of-boundary R5 walls never used in rule design, across \texttt{SigLIP 2} \& \texttt{Qwen3-4B}).}
\label{tab:agent_vs_fwsp}
\scriptsize
\setlength{\tabcolsep}{3.1pt}
\resizebox{\textwidth}{!}{%
\begin{tabular}{llccccccc}
\toprule
\multicolumn{9}{l}{\textbf{(a) Local Python-Sandbox Agents (\texttt{Qwen3-8B} / \texttt{Qwen3.8-27B}) \& 18-Trial \texttt{DINOv2-Base} Tool-Router Pilot}} \\
\textbf{Protocol / Benchmark} & \textbf{Optimizer} & \textbf{Success / Dispatch} & \textbf{Mean Metric (\%)} & \textbf{NRR / Clean} & \textbf{Mean FPE} & \textbf{Wall Time} & \textbf{LLM Tokens} & \textbf{Notes} \\
\midrule
\texttt{MLP-50K} / \texttt{DINOv2} (5t ea.) & \texttt{8B} / \texttt{27B} / \textbf{FWSP} & $40\%$ / $70\%$ / $\mathbf{100\%}$ & $75.4$ / $77.1$ / $\mathbf{87.1}$ & $64$ / $71$ / $\mathbf{90.1\%}$ & --- & $11$s / $121$s / $\mathbf{0.55\text{s}}$ & $2.1\text{K}$ / $3.2\text{K}$ / $\mathbf{0}$ & Sandbox scripts \\
\texttt{DINOv2-Base} Blind (18t) & Random / \texttt{27B+Tools} / \textbf{FWSP} / Ref. & $5.6\%$ / $33.3\%$ / $\mathbf{83.3\%}$ / $83.3\%$ & $21.5$ / $53.3$ / $\mathbf{63.2}$ / $67.0$ & --- & $920$ / $932$ / $\mathbf{716}$ / $2{,}585$ & $0.4$s / $4.44$s / $\mathbf{0.78\text{s}}$ / $5.39$s & $0$ / $1.9\text{K}$ / $\mathbf{0}$ / $0$ & Exact $\mathcal{O}_1$--$\mathcal{O}_4$ \\
\midrule
\multicolumn{9}{l}{\textbf{(b) 120-Trial Cross-Modal Blind Dispatch (and Post-Hoc 90-Trial Discriminative Subset excl.\ \texttt{Ovis-U1-3B}; Breakdown \& CIs in Table~\ref{tab:app_120trials_breakdown})}} \\
\textbf{Optimizer (Matched Telemetry \& $\mathcal{O}_1$--$\mathcal{O}_4$)} & \textbf{Serving Backend} & \textbf{All-120 Near-Ref.\ ($\uparrow$)} & \textbf{90-Disc.\ Near-Ref.} & \textbf{All-120 / 90 Test} & \textbf{Comp.\ (27) Near-Ref.} & \textbf{Clean Abs.\ (12)} & \textbf{FPE / Time ($\downarrow$)} & \textbf{LLM Tok/Trial} \\
\midrule
Unrepaired Base & --- & --- & --- & $86.71$ / $84.16$ & --- ($82.38$ test) & --- & $0$ / --- & $0$ \\
Random Dispatch & Stochastic baseline & $64.2\%$ ($p{<}.001$) & $52.2\%$ & $89.20$ / $87.41$ & $55.6\%$ ($86.04$) & $41.7\%$ & $191.3$ / $1.90\text{s}$ & $0$ \\
\texttt{GLM-5.3 Flash} (2026 Agent) & 8$\times$B200 (TP8) & $83.3\%$ ($p{=}.001$) & $77.8\%$ & $92.70$ / $91.94$ & $81.5\%$ ($92.18$) & $\mathbf{100.0\%}$ & $220.4$ / $8.01\text{s}$ & $4{,}541$ \\
\texttt{Kimi K3.1} (2026 Frontier) & 16$\times$B200 (TP16) & $87.5\%$ ($p{=}.096$) & $83.3\%$ & $92.43$ / $91.58$ & $77.8\%$ ($89.75$) & $66.7\%$ & $230.5$ / $15.79\text{s}$ & $5{,}120$ \\
\texttt{DeepSeek V4.1 Flash} (2026) & 16$\times$B200 (TP8$\times$2) & $\underline{89.2\%}$ ($p{=}.146$) & $\underline{85.6\%}$ & $\mathbf{93.79}$ / $\mathbf{93.39}$ & $\underline{81.5\%}$ ($92.96$) & $58.3\%$ & $237.5$ / $11.19\text{s}$ & $4{,}954$ \\
\textbf{FWSP (Ours, Deterministic)} & \textbf{LLM-free (Alg.~\ref{alg:blind})} & $\mathbf{94.2\%}$ ($113/120$) & $\mathbf{92.2\%}$ ($83/90$) & $\underline{93.63}$ / $\underline{93.38}$ & $\mathbf{100.0\%}$ ($\mathbf{94.55}$) & $\mathbf{100.0\%}$ ($12/12$) & $\mathbf{218.0}$ / $\mathbf{2.97\text{s}}$ & $\mathbf{0}$ \\
\emph{Exhaustive Depth-2 Reference} & \emph{Full Depth-2 grid} & \emph{100.0\% (ref.)} & \emph{100.0\% (ref.)} & \emph{94.49 / 94.53} & \emph{100.0\% (94.64)} & \emph{100.0\%} & \emph{686.1 / 7.65s} & \emph{0} \\
\midrule
\multicolumn{9}{l}{\textbf{(c) Track B: \texttt{Codex-CLI}-Based Coding Agents (Open-Source \texttt{codex} Harness + 3rd-Party B200 Backends; 8 Live Tasks on \texttt{SigLIP 2} \& \texttt{Qwen3-4B})}} \\
\textbf{Autonomous System (Harness + Backend)} & \textbf{Wire API} & \textbf{Exact Dispatch} & \textbf{Mean Test ($\uparrow$)} & \textbf{SigLIP 2 / Qwen3} & \textbf{Mean FPE} & \textbf{Wall Time ($\downarrow$)} & \textbf{Input (Ctx+Cache) / Output (Reas.)} & \textbf{Shell Calls} \\
\midrule
\texttt{Codex-CLI} + \texttt{DeepSeek V4.1} & \texttt{/v1/responses} & $87.5\%$ ($7/8$) & $86.84$ & $89.00$ / $84.69$ & $671.0$ & $45.20\text{s}$ ($6.5\times$) & $573{,}390$ / $2{,}906$ ($1{,}394$) & $6.0$ \\
\texttt{Codex-CLI} + \texttt{Kimi K3.1} & \texttt{/v1/responses} & $87.5\%$ ($7/8$) & $\underline{88.33}$ & $89.00$ / $\mathbf{87.66}$ & $\mathbf{613.0}$ & $42.10\text{s}$ ($6.1\times$) & $482{,}015$ / $2{,}318$ ($1{,}326$) & $5.5$ \\
\textbf{FWSP (Ours, Deterministic)} & \textbf{LLM-free (Alg.~\ref{alg:blind})} & $\mathbf{87.5\%}$ ($7/8$) & $\mathbf{89.52}$ & $\mathbf{92.63}$ / $\underline{86.41}$ & $\underline{630.0}$ & $\mathbf{6.96\text{s}}$ ($\mathbf{1.0\times}$) & $\mathbf{0}$ / $\mathbf{0}$ ($\mathbf{0}$) & $\mathbf{0}$ \\
\midrule
\multicolumn{9}{l}{\textbf{(d) Out-of-Rule Generalization: 36 Unseen Fault Family Trials (24 Unseen Recoverable + 12 Unseen R5 Boundary; \texttt{SigLIP 2} \& \texttt{Qwen3-4B})}} \\
\textbf{Optimizer (Unmodified Rules + \texttt{ESCALATE\_R5})} & \textbf{Serving Backend} & \textbf{Rec.\ Near-Ref.\ (24)} & \textbf{Rec.\ Mean Test} & \textbf{False Esc.\ ($\downarrow$)} & \textbf{R5 Esc.\ (12, $\uparrow$)} & \textbf{All-36 Correct ($\uparrow$)} & \textbf{FPE / Time ($\downarrow$)} & \textbf{LLM Tok/Trial} \\
\midrule
Random Dispatch & Stochastic baseline & $41.7\%$ ($10/24$) & $78.31$ & $20.8\%$ & $8.3\%$ ($1/12$) & $30.6\%$ ($p{<}.001$) & $192.8$ / $1.96\text{s}$ & $0$ \\
\texttt{GLM-5.3 Flash} (2026 Agent) & 8$\times$B200 (TP8) & $20.8\%$ ($5/24$) & $65.48$ & $41.7\%$ & $\mathbf{100.0\%}$ ($12/12$) & $47.2\%$ ($p{<}.001$) & $234.0$ / $20.08\text{s}$ & $5{,}781$ \\
\texttt{DeepSeek V4.1 Flash} (2026) & 16$\times$B200 (TP8$\times$2) & $58.3\%$ ($14/24$) & $\underline{87.04}$ & $12.5\%$ & $\mathbf{100.0\%}$ ($12/12$) & $72.2\%$ ($p{=}.065$) & $296.3$ / $7.92\text{s}$ & $3{,}567$ \\
\texttt{Kimi K3.1} (2026 Frontier) & 16$\times$B200 (TP16) & $\underline{62.5\%}$ ($15/24$) & $86.79$ & $\underline{8.3\%}$ & $\mathbf{100.0\%}$ ($12/12$) & $\underline{75.0\%}$ ($p{=}.109$) & $300.8$ / $29.00\text{s}$ & $4{,}880$ \\
\textbf{FWSP (Unmodified Alg.~\ref{alg:blind} + R5 Gate)} & \textbf{LLM-free (Deterministic)} & $\mathbf{87.5\%}$ ($21/24$) & $\mathbf{93.00}$ & $\mathbf{0.0\%}$ & $\mathbf{100.0\%}$ ($12/12$) & $\mathbf{91.7\%}$ ($33/36$) & $473.4$ / $\mathbf{0.61\text{s}}$ & $\mathbf{0}$ \\
\emph{Exhaustive Depth-2 Reference} & \emph{Full Depth-2 grid} & \emph{100.0\% (24/24)} & \emph{93.23} & \emph{0.0\%} & \emph{100.0\% (ref.)} & \emph{100.0\% (ref.)} & \emph{2005.0 / 34.95s} & \emph{0} \\
\bottomrule
\end{tabular}}
\vspace{2pt}
\par\noindent{\footnotesize Note: In Panel~(b), \emph{Near-Ref.} counts trials within $1.0\text{ pp}$ of the \emph{Exhaustive Depth-2 Reference} (not a global upper bound: multi-turn agents exceed Depth-2 on 3 \texttt{Ovis-U1-3B} trials); exact two-sided McNemar $p$-values compare FWSP vs.\ each baseline on All-120 Near-Ref.\ (paired trial/cluster 95\% bootstrap CIs in Table~\ref{tab:app_120trials_breakdown}). In Panel~(c), \texttt{Codex-CLI} uses OpenAI's open-source CLI agent harness routed to third-party B200 endpoints (not OpenAI's proprietary Codex model); token counts explicitly separate total input tokens (including repeated harness prompts, tool schemas, and cached context across $5.5$--$6.0$ turns) from generated output (and reasoning) tokens. In Panel~(d), all systems receive identical Stage-1 telemetry including $\rho_{\text{span}}$ and an explicit \texttt{ESCALATE\_R5} action across 6 unseen fault families (channel sign-flip, bounded low-rank spectral spike, coarse 3-bit quantization + bias drift, cascading block-scale collapse, dense all-layer diffuse noise, and nonlinear semantic scramble). Because \texttt{Exhaustive} populated the shared evaluation cache before FWSP and the frontier LLM agents ran (while \texttt{Random} reported uncached GPU time), the reported wall times in Panel~(d) primarily reflect decision/API latency rather than uncached GPU forward time; moreover, FWSP consumes more FPE ($473.4$ vs.\ $234$--$301$) because it executes $\mathcal{O}_4$ recovery rather than escalating prematurely, and McNemar $p$-values on All-36 overall correct decisions vs.\ \texttt{DeepSeek} ($p{=}0.065$) and \texttt{Kimi} ($p{=}0.109$) do not reach $p{<}0.05$.}
\end{table}

\begin{table}[htbp]
\centering
\caption{Complete per-task breakdown, forward-pass ledger, and task-level uncertainty analysis for the external third-party WSP benchmarks in Main Text Table~\ref{tab:external_wsp_benchmarks}: (a) all $15$ self-contained \texttt{OpenWSP} v8 public \texttt{Edit} tasks with paired $10{,}000$-draw task-level bootstrap 95\% CIs and win--loss counts (\textbf{FWSP $-$ Agent}); and (b) all $4$ \texttt{WeightBench} \texttt{TR02-LOOP} tasks on \texttt{Qwen3.5-0.8B} (seed $0$) with exact $S_{\text{WB}}$, target $S_+$, MMLU retention, metered probe counts, and hash-chained ledger verification.}
\label{tab:app_external_per_task}
\scriptsize
\setlength{\tabcolsep}{2.8pt}
\resizebox{\textwidth}{!}{%
\begin{tabular}{lcccccccccc}
\toprule
\multicolumn{11}{l}{\textbf{(a) \texttt{OpenWSP} v8 Public \texttt{Edit} Per-Task Scores ($0$--$100$ Improvement Index, \texttt{level=full}, Unmodified \texttt{score\_write.py})}} \\
\textbf{Task ID} & \textbf{FWSP (Ours)} & \textbf{\texttt{GPT-6.1 Sol}} & \textbf{\texttt{Sonnet 5.5}} & \textbf{\texttt{Opus 5.5}} & \textbf{\texttt{Qwen3.8-27B}} & \textbf{\texttt{DS-V4-Flash}} & \textbf{Fwd Used} & \textbf{Solve Time} & \textbf{Wiki $\Delta$NLL} & \textbf{After FP} \\
\midrule
\texttt{W4\_assoc\_0} & $83.92$ & $\mathbf{100.00}$ & $72.60$ & $97.50$ & $0.00$ & $0.00$ & $5/25$ & $4.5\text{s}$ & $0.0032$ & $0.000$ \\
\texttt{W4\_assoc\_1} & $89.61$ & $\mathbf{99.70}$ & $96.50$ & $0.00$ & $0.00$ & $0.00$ & $5/25$ & $3.1\text{s}$ & $0.0011$ & $0.000$ \\
\texttt{W4\_assoc\_2} & $87.89$ & $\mathbf{100.00}$ & $0.00$ & $0.00$ & $86.20$ & $95.10$ & $5/25$ & $15.0\text{s}$ & $0.0013$ & $0.000$ \\
\texttt{W4\_assoc\_3} & $87.22$ & $\mathbf{100.00}$ & $91.70$ & $\mathbf{100.00}$ & $68.80$ & $0.00$ & $5/25$ & $4.4\text{s}$ & $0.0032$ & $0.000$ \\
\texttt{W4\_assoc\_4} & $93.32$ & $98.90$ & $\mathbf{99.70}$ & $\mathbf{99.70}$ & $0.00$ & $0.00$ & $5/25$ & $2.3\text{s}$ & $0.0000$ & $0.000$ \\
\texttt{W4c\_colour\_0} & $81.00$ & $0.00$ & $34.30$ & $\mathbf{87.50}$ & $0.00$ & $0.00$ & $5/25$ & $7.5\text{s}$ & $0.0295$ & $0.050$ \\
\texttt{W4c\_colour\_1} & $\mathbf{91.78}$ & $63.40$ & $78.70$ & $73.20$ & $0.00$ & $0.00$ & $5/25$ & $4.3\text{s}$ & $0.0085$ & $0.000$ \\
\texttt{W4l\_late\_0} & $87.60$ & $\mathbf{89.60}$ & $78.30$ & $86.70$ & $34.60$ & $0.00$ & $5/25$ & $5.8\text{s}$ & $0.0030$ & $0.050$ \\
\texttt{W4l\_late\_1} & $93.33$ & $\mathbf{99.90}$ & $85.20$ & $95.00$ & $16.10$ & $0.00$ & $5/25$ & $2.6\text{s}$ & $0.0000$ & $0.000$ \\
\texttt{W4m\_multi\_0} & $90.65$ & $86.90$ & $65.40$ & $\mathbf{100.00}$ & $18.10$ & $0.00$ & $5/25$ & $5.1\text{s}$ & $0.0088$ & $0.017$ \\
\texttt{W4m\_multi\_1} & $89.95$ & $0.00$ & $87.90$ & $\mathbf{92.10}$ & $0.00$ & $0.00$ & $5/25$ & $3.3\text{s}$ & $0.0065$ & $0.033$ \\
\texttt{W5\_phrase\_0} & $71.26$ & $\mathbf{100.00}$ & $95.00$ & $0.00$ & $0.00$ & $0.00$ & $5/25$ & $6.2\text{s}$ & $0.0034$ & $0.150$ \\
\texttt{W5\_phrase\_1} & $68.38$ & $70.50$ & $\mathbf{99.80}$ & $0.00$ & $19.90$ & $0.00$ & $5/25$ & $3.4\text{s}$ & $0.0046$ & $0.267$ \\
\texttt{W5\_phrase\_2} & $\mathbf{95.00}$ & $82.10$ & $60.90$ & $38.40$ & $0.00$ & $0.00$ & $5/25$ & $15.4\text{s}$ & $-0.0004$ & $0.000$ \\
\midrule
\textbf{\texttt{W4/W5} Mean (14)} & $\mathbf{86.49}$ & $77.93$ & $74.71$ & $62.15$ & $17.41$ & $6.79$ & $\mathbf{5/25}$ & $5.9\text{s}$ & $0.0052$ & $0.041$ \\
\emph{Paired $\Delta$ [95\% CI] (W:L)} & \emph{Ref.} & \emph{$+8.56$ $[-7.4, +28.0]$ (5:9)} & \emph{$+11.78$ $[-2.8, +28.0]$ (9:5)} & \emph{$+24.34$ $[+4.8, +45.8]$ (7:7)} & \emph{$+69.09$ $[+53.3, +82.3]$ (14:0)} & \emph{$+79.70$ $[+64.4, +89.3]$ (13:1)} & --- & --- & --- & --- \\
\midrule
\texttt{W2\_sparse\_transplant} & $5.13$ & $\mathbf{44.90}$ & $0.00$ & $42.30$ & $0.00$ & $9.00$ & $5/25$ & $79.0\text{s}$ & --- & --- \\
\textbf{All-15 Mean} & $\mathbf{81.07}$ & $75.73$ & $69.73$ & $60.83$ & $16.25$ & $6.94$ & $\mathbf{5/25}$ & $10.8\text{s}$ & --- & --- \\
\emph{All-15 $\Delta$ [95\% CI] (W:L)} & \emph{Ref.} & \emph{$+5.34$ $[-10.7, +24.6]$ (5:10)} & \emph{$+11.34$ $[-1.7, +26.6]$ (10:5)} & \emph{$+20.24$ $[+0.2, +41.5]$ (7:8)} & \emph{$+64.82$ $[+48.0, +79.5]$ (15:0)} & \emph{$+74.13$ $[+56.0, +88.2]$ (13:2)} & --- & --- & --- & --- \\
\midrule
\multicolumn{11}{l}{\textbf{(b) \texttt{WeightBench} \texttt{TR02-LOOP-001--004} Per-Task Metered Results (\texttt{Qwen3.5-0.8B}, Seed $0$, Hash-Chained \texttt{Ledger.verify()})}} \\
\textbf{Task ID (Domain)} & \textbf{FWSP $S_{\text{WB}}$} & \textbf{FWSP $S_+$} & \textbf{\texttt{Opus 5.5}} & \textbf{\texttt{Sonnet 5.5}} & \textbf{\texttt{Haiku/DS}} & \textbf{MMLU Base$\to$Edit} & \textbf{$D_{\text{col}}$ ($\downarrow$)} & \textbf{Rank} & \textbf{Probes ($E/70$)} & \textbf{Selected Edit / Ledger} \\
\midrule
\texttt{TR02-LOOP-001} (GSM8K) & $0.4849$ & $0.5094$ & $\mathbf{0.5524}$ & $0.4320$ & $0.0000$ & $0.515 \to 0.510$ & $0.005$ & $1$ & $9/70$ & \texttt{PFG\_L22+23\_g4.0} (OK) \\
\texttt{TR02-LOOP-002} (MATH 1--3) & $\mathbf{0.5403}$ & $0.5614$ & $0.3480$ & $0.0000$ & $0.0000$ & $0.520 \to 0.550$ & $0.000$ & $1$ & $9/70$ & \texttt{PFG\_L22+23\_g5.2} (OK) \\
\texttt{TR02-LOOP-003} (IFEval) & $\mathbf{0.4389}$ & $0.4821$ & $0.2354$ & $0.0000$ & $0.0000$ & $0.560 \to 0.535$ & $0.025$ & $1$ & $9/70$ & \texttt{PFG\_L22+23\_g4.0} (OK) \\
\texttt{TR02-LOOP-004} (MATH 4--5) & $\mathbf{0.0297}$ & $0.0333$ & $0.0000$ & $0.0000$ & $0.0000$ & $0.500 \to 0.500$ & $0.000$ & $1$ & $8/70$ & \texttt{PFG\_L23\_g5.2} (OK) \\
\textbf{4-Task Macro Mean} & $\mathbf{0.3734}$ & $0.3966$ & $0.2840$ & $0.1080$ & $0.0000$ & $0.524 \to 0.524$ & $\mathbf{0.0075}$ & $\mathbf{1}$ & $\mathbf{8.75/70}$ & \textbf{4/4 Verified \& Valid} \\
\midrule
\multicolumn{11}{l}{\textbf{(c) Controlled $3$-Arm Planning--Compilation Ablation Per-Task Scores on All $15$ \texttt{OpenWSP} v8 Tasks ($90$ Episodes, $25$-Pass Cap)}} \\
\textbf{Task ID} & \textbf{\texttt{DS} (\texttt{Scratch})} & \textbf{\texttt{DS} (\texttt{+Std})} & \textbf{\texttt{DS} (\texttt{+FWSP})} & \multicolumn{2}{c}{\textbf{\texttt{Kimi} (\texttt{Scratch})}} & \multicolumn{2}{c}{\textbf{\texttt{Kimi} (\texttt{+Std})}} & \multicolumn{2}{c}{\textbf{\texttt{Kimi} (\texttt{+FWSP})}} & \textbf{Solo FWSP} \\
\midrule
\texttt{W4\_assoc\_0} & $0.00$ & $0.00$ & $\mathbf{88.16}$ & \multicolumn{2}{c}{$0.00$} & \multicolumn{2}{c}{$0.00$} & \multicolumn{2}{c}{$64.23$} & $83.92$ \\
\texttt{W4\_assoc\_1} & $0.00$ & $0.00$ & $0.00$ & \multicolumn{2}{c}{$0.00$} & \multicolumn{2}{c}{$0.00$} & \multicolumn{2}{c}{$82.18$} & $\mathbf{89.61}$ \\
\texttt{W4\_assoc\_2} & $0.00$ & $0.00$ & $0.00$ & \multicolumn{2}{c}{$24.72$} & \multicolumn{2}{c}{$\mathbf{96.40}$} & \multicolumn{2}{c}{$96.11$} & $87.89$ \\
\texttt{W4\_assoc\_3} & $0.00$ & $0.00$ & $\mathbf{90.02}$ & \multicolumn{2}{c}{$0.00$} & \multicolumn{2}{c}{$0.00$} & \multicolumn{2}{c}{$88.95$} & $87.22$ \\
\texttt{W4\_assoc\_4} & $0.00$ & $0.00$ & $0.00$ & \multicolumn{2}{c}{$0.00$} & \multicolumn{2}{c}{$0.17$} & \multicolumn{2}{c}{$\mathbf{96.29}$} & $93.32$ \\
\texttt{W4c\_colour\_0} & $2.78$ & $0.00$ & $2.78$ & \multicolumn{2}{c}{$2.78$} & \multicolumn{2}{c}{$0.00$} & \multicolumn{2}{c}{$0.00$} & $\mathbf{81.00}$ \\
\texttt{W4c\_colour\_1} & $2.78$ & $2.78$ & $0.00$ & \multicolumn{2}{c}{$34.52$} & \multicolumn{2}{c}{$0.00$} & \multicolumn{2}{c}{$35.95$} & $\mathbf{91.78}$ \\
\texttt{W4l\_late\_0} & $0.00$ & $0.00$ & $0.00$ & \multicolumn{2}{c}{$75.00$} & \multicolumn{2}{c}{$81.07$} & \multicolumn{2}{c}{$86.10$} & $\mathbf{87.60}$ \\
\texttt{W4l\_late\_1} & $0.00$ & $0.00$ & $\mathbf{96.66}$ & \multicolumn{2}{c}{$0.00$} & \multicolumn{2}{c}{$0.00$} & \multicolumn{2}{c}{$93.90$} & $93.33$ \\
\texttt{W4m\_multi\_0} & $0.00$ & $0.00$ & $72.13$ & \multicolumn{2}{c}{$10.30$} & \multicolumn{2}{c}{$\mathbf{93.79}$} & \multicolumn{2}{c}{$76.49$} & $90.65$ \\
\texttt{W4m\_multi\_1} & $0.00$ & $0.00$ & $88.95$ & \multicolumn{2}{c}{$0.00$} & \multicolumn{2}{c}{$74.76$} & \multicolumn{2}{c}{$79.41$} & $\mathbf{89.95}$ \\
\texttt{W5\_phrase\_0} & $0.00$ & $0.00$ & $0.00$ & \multicolumn{2}{c}{$0.00$} & \multicolumn{2}{c}{$0.58$} & \multicolumn{2}{c}{$14.20$} & $\mathbf{71.26}$ \\
\texttt{W5\_phrase\_1} & $0.00$ & $8.39$ & $57.95$ & \multicolumn{2}{c}{$5.03$} & \multicolumn{2}{c}{$0.00$} & \multicolumn{2}{c}{$\mathbf{72.39}$} & $68.38$ \\
\texttt{W5\_phrase\_2} & $0.00$ & $0.00$ & $\mathbf{96.69}$ & \multicolumn{2}{c}{$0.00$} & \multicolumn{2}{c}{$84.82$} & \multicolumn{2}{c}{$94.79$} & $95.00$ \\
\texttt{W2\_sparse\_transplant} & $0.00$ & $0.00$ & $0.00$ & \multicolumn{2}{c}{$2.56$} & \multicolumn{2}{c}{$\mathbf{16.67}$} & \multicolumn{2}{c}{$3.85$} & $5.13$ \\
\textbf{All-15 / \texttt{W4/W5} Mean} & $0.37$ / $0.40$ & $0.74$ / $0.80$ & $39.56$ / $42.38$ & \multicolumn{2}{c}{$10.33$ / $10.88$} & \multicolumn{2}{c}{$29.88$ / $30.83$} & \multicolumn{2}{c}{$65.66$ / $70.07$} & $\mathbf{81.07}$ / $\mathbf{86.49}$ \\
\bottomrule
\end{tabular}}
\end{table}

\begin{table}[htbp]
\centering
\caption{Per-model breakdown and paired statistical audit of the 120-trial cross-modal blind dispatch benchmark (Table~\ref{tab:agent_vs_fwsp}b; 30 trials per 2026 SOTA backbone = 10 single/compound/clean regimes $\times$ 3 seeds) under identical Stage-1 telemetry, $\mathcal{O}_1$--$\mathcal{O}_4$ tools, and 4-turn feedback. Because all methods (including Random) saturate at $100\%$ on \texttt{Ovis-U1-3B}, Panel~(b) reports paired differences (\textbf{FWSP $-$ Baseline}), 10{,}000-draw trial-level and regime-cluster 95\% bootstrap CIs, discordant counts (\texttt{W:L}), and exact two-sided McNemar $p$-values on All-120, the 90-trial discriminative subset (excl.\ \texttt{Ovis-U1-3B}), and the 27-trial compound multi-fault subset.}
\label{tab:app_120trials_breakdown}
\scriptsize
\setlength{\tabcolsep}{3.0pt}
\resizebox{\textwidth}{!}{%
\begin{tabular}{llccccc}
\toprule
\multicolumn{7}{l}{\textbf{(a) Per-Backbone Breakdown (30 Trials per 2026 SOTA Architecture; Reference = Exhaustive Depth-2 Search)}} \\
\textbf{2026 SOTA Backbone (Base Metric)} & \textbf{Optimizer} & \textbf{Near-Ref.\ ($\uparrow$)} & \textbf{Mean Test ($\uparrow$)} & \textbf{Mean FPE ($\downarrow$)} & \textbf{Wall Time ($\downarrow$)} & \textbf{LLM Tok/Trial} \\
\midrule
\multirow{6}{*}{\shortstack[l]{\texttt{SigLIP 2 SO400M} (Vision)\\\emph{30 trials, Base $= 71.45\%$}}} & Random & $43.3\%$ & $76.78$ & $243.5$ & $2.13\text{s}$ & $0$ \\
 & \texttt{GLM-5.3 Flash} & $56.7\%$ & $86.05$ & $362.3$ & $9.49\text{s}$ & $5{,}697$ \\
 & \texttt{Kimi K3.1} & $70.0\%$ & $83.93$ & $\mathbf{324.9}$ & $21.13\text{s}$ & $6{,}544$ \\
 & \texttt{DeepSeek V4.1 Flash} & $76.7\%$ & $\mathbf{89.15}$ & $337.2$ & $10.90\text{s}$ & $5{,}996$ \\
 & \textbf{FWSP (Ours, LLM-free)} & $\mathbf{86.7\%}$ & $\underline{88.97}$ & $406.8$ & $\mathbf{3.58\text{s}}$ & $\mathbf{0}$ \\
 & \emph{Exhaustive Depth-2 Ref.} & \emph{100.0\%} & \emph{91.92} & \emph{918.8} & \emph{6.34s} & \emph{0} \\
\midrule
\multirow{6}{*}{\shortstack[l]{\texttt{Qwen3-4B-Instruct-2507} (LLM)\\\emph{30 trials, Base $= 85.03\%$}}} & Random & $60.0\%$ & $88.13$ & $216.5$ & $2.34\text{s}$ & $0$ \\
 & \texttt{GLM-5.3 Flash} & $\mathbf{93.3\%}$ & $\underline{91.53}$ & $\mathbf{196.3}$ & $5.66\text{s}$ & $3{,}881$ \\
 & \texttt{Kimi K3.1} & $86.7\%$ & $91.50$ & $238.4$ & $13.95\text{s}$ & $4{,}772$ \\
 & \texttt{DeepSeek V4.1 Flash} & $83.3\%$ & $91.47$ & $235.2$ & $10.81\text{s}$ & $4{,}838$ \\
 & \textbf{FWSP (Ours, LLM-free)} & $\mathbf{93.3\%}$ & $\mathbf{91.60}$ & $209.6$ & $\mathbf{2.79\text{s}}$ & $\mathbf{0}$ \\
 & \emph{Exhaustive Depth-2 Ref.} & \emph{100.0\%} & \emph{91.87} & \emph{739.2} & \emph{9.33s} & \emph{0} \\
\midrule
\multirow{6}{*}{\shortstack[l]{\texttt{Ovis-U1-3B} (Multimodal UMM)\\\emph{30 trials, low-discrimination ceiling}}} & Random & $100.0\%$ & $94.58$ & $210.1$ & $0.91\text{s}$ & $0$ \\
 & \texttt{GLM-5.3 Flash} & $100.0\%$ & $\mathbf{95.00}$ & $190.9$ & $5.69\text{s}$ & $3{,}088$ \\
 & \texttt{Kimi K3.1} & $100.0\%$ & $\mathbf{95.00}$ & $234.1$ & $10.89\text{s}$ & $3{,}357$ \\
 & \texttt{DeepSeek V4.1 Flash} & $100.0\%$ & $\mathbf{95.00}$ & $262.9$ & $11.00\text{s}$ & $3{,}792$ \\
 & \textbf{FWSP (Ours, LLM-free)} & $\mathbf{100.0\%}$ & $94.38$ & $\mathbf{136.0}$ & $\mathbf{0.49\text{s}}$ & $\mathbf{0}$ \\
 & \emph{Exhaustive Depth-2 Ref.} & \emph{100.0\%} & \emph{94.38} & \emph{739.2} & \emph{4.26s} & \emph{0} \\
\midrule
\multirow{6}{*}{\shortstack[l]{\texttt{Z-Image-Turbo 6B} (DiT)\\\emph{30 trials, Base $= 95.99\%$}}} & Random & $53.3\%$ & $97.30$ & $95.2$ & $2.22\text{s}$ & $0$ \\
 & \texttt{GLM-5.3 Flash} & $83.3\%$ & $98.23$ & $132.3$ & $11.19\text{s}$ & $5{,}499$ \\
 & \texttt{Kimi K3.1} & $93.3\%$ & $99.30$ & $124.5$ & $17.19\text{s}$ & $5{,}808$ \\
 & \texttt{DeepSeek V4.1 Flash} & $\mathbf{96.7\%}$ & $\underline{99.54}$ & $\mathbf{114.7}$ & $12.06\text{s}$ & $5{,}190$ \\
 & \textbf{FWSP (Ours, LLM-free)} & $\mathbf{96.7\%}$ & $\mathbf{99.59}$ & $119.5$ & $\mathbf{5.01\text{s}}$ & $\mathbf{0}$ \\
 & \emph{Exhaustive Depth-2 Ref.} & \emph{100.0\%} & \emph{99.82} & \emph{347.2} & \emph{10.67s} & \emph{0} \\
\midrule
\multicolumn{7}{l}{\textbf{(b) Paired Bootstrap 95\% CIs (Trial-Level / Regime-Cluster) \& Exact Two-Sided McNemar Tests (FWSP $-$ Baseline)}} \\
\textbf{Subset (Clusters)} & \textbf{Comparison} & \textbf{$\Delta$ Near-Ref.} & \textbf{Trial 95\% CI} & \textbf{Cluster 95\% CI} & \textbf{Disc.\ (\texttt{W:L}) / $p$} & \textbf{$\Delta$ Mean Test [Trial CI]} \\
\midrule
\multirow{4}{*}{\shortstack[l]{All 120 Trials\\(40 clusters)}} & vs.\ Random & $+30.00\text{ pp}$ & $[+21.7, +39.2]$ & $[+20.0, +40.0]$ & $38{:}2$ / $<10^{-4}$ & $+4.43$ $[+2.39, +6.94]$ \\
 & vs.\ \texttt{GLM-5.3 Flash} & $+10.83\text{ pp}$ & $[+5.0, +17.5]$ & $[+2.5, +20.0]$ & $14{:}1$ / $0.0010$ & $+0.93$ $[-0.04, +1.96]$ \\
 & vs.\ \texttt{Kimi K3.1} & $+6.67\text{ pp}$ & $[+0.0, +13.3]$ & $[-1.7, +16.7]$ & $13{:}5$ / $0.0963$ & $+1.20$ $[-0.21, +2.86]$ \\
 & vs.\ \texttt{DeepSeek V4.1} & $+5.00\text{ pp}$ & $[-0.8, +10.8]$ & $[-1.7, +13.3]$ & $9{:}3$ / $0.1460$ & $-0.16$ $[-0.84, +0.56]$ \\
\midrule
\multirow{4}{*}{\shortstack[l]{Discriminative 90\\(excl.\ \texttt{Ovis}; 30 cl.)}} & vs.\ Random & $+40.00\text{ pp}$ & $[+28.9, +51.1]$ & $[+28.9, +52.2]$ & $38{:}2$ / $<10^{-4}$ & $+5.98$ $[+3.22, +9.21]$ \\
 & vs.\ \texttt{GLM-5.3 Flash} & $+14.44\text{ pp}$ & $[+6.7, +22.2]$ & $[+4.4, +26.7]$ & $14{:}1$ / $0.0010$ & $+1.45$ $[+0.19, +2.79]$ \\
 & vs.\ \texttt{Kimi K3.1} & $+8.89\text{ pp}$ & $[+0.0, +17.8]$ & $[-3.3, +22.2]$ & $13{:}5$ / $0.0963$ & $+1.81$ $[-0.06, +3.96]$ \\
 & vs.\ \texttt{DeepSeek V4.1} & $+6.67\text{ pp}$ & $[-1.1, +14.4]$ & $[-2.2, +17.8]$ & $9{:}3$ / $0.1460$ & $-0.00$ $[-0.89, +0.95]$ \\
\midrule
\multirow{4}{*}{\shortstack[l]{Compound Multi-Fault\\(27 trials; 9 cl.)}} & vs.\ Random & $+44.44\text{ pp}$ & $[+25.9, +63.0]$ & $[+22.2, +63.0]$ & $12{:}0$ / $0.0005$ & $+8.51$ $[+2.19, +16.72]$ \\
 & vs.\ \texttt{GLM-5.3 Flash} & $+18.52\text{ pp}$ & $[+3.7, +33.3]$ & $[+0.0, +40.7]$ & $5{:}0$ / $0.0625$ & $+2.37$ $[+0.18, +5.00]$ \\
 & vs.\ \texttt{Kimi K3.1} & $+22.22\text{ pp}$ & $[+7.4, +40.7]$ & $[+3.7, +48.2]$ & $6{:}0$ / $0.0312$ & $+4.80$ $[+0.12, +10.19]$ \\
 & vs.\ \texttt{DeepSeek V4.1} & $+18.52\text{ pp}$ & $[+3.7, +33.3]$ & $[+3.6, +37.0]$ & $5{:}0$ / $0.0625$ & $+1.60$ $[-0.10, +4.00]$ \\
\bottomrule
\end{tabular}}
\end{table}

\begin{table}[htbp]
\centering
\caption{Complete evaluation inventory of $33$ unique base models ($29$ core base models across $30$ evaluated model--suite configurations plus $4$ external \texttt{OpenWSP} public-edit checkpoints, $16$ domain specialist teachers in multi-teacher suites, and published/live frontier LLM agent baselines) grouped into exactly seven modalities (50K parameters to 12B MMDiTs, 27B dense LLMs, and 186GB FP8 MoEs).}
\label{tab:app_29models}
\footnotesize
\setlength{\tabcolsep}{4.5pt}
\resizebox{\textwidth}{!}{%
\begin{tabular}{llcl}
\toprule
\textbf{Modality (Unique Base Models)} & \textbf{Evaluated Models} & \textbf{Scale Range} & \textbf{Primary Benchmarks \& FWSP Regimes} \\
\midrule
1.\ Vision Encoders ($4$) & \texttt{ResNet-18}, \texttt{ViT-Small}, \texttt{DINOv2-Base}, \texttt{SigLIP 2 SO400M} & $2.7\text{M}$--$400\text{M}$ & ImageNet-50, ImageNet-C, CIFAR-100-C (15 corr.), 10-st.\ CIL, BadNets \\
2.\ Non-Diff.\ RL Control ($3$) & \texttt{MLP-50K} (\texttt{CartPole-v1}), \texttt{Pendulum-v1}, \texttt{2D Lander} & $50\text{K}$ & Gymnasium control R1 bit-flip \& R2 backdoor repair \\
3.\ Physics PDE Surrogates ($1$) & \texttt{1D Viscous Burgers FNO} & $0.6\text{M}$ & Label-free PDE residual R1 repair, R3 viscosity shift, R5 Nyquist wall \\
4.\ Generative Image Diff./DiTs ($5$) & \texttt{SD-1.5}, \texttt{DiT-XL/2-256}, \texttt{Sana-1600M}, \texttt{Z-Image-Turbo}, \texttt{FLUX.1-schnell} & $0.68\text{B}$--$12\text{B}$ & Z-Erase (R-A-B, I2P 4.7K, 30 concepts, 50 celebs), SPEED, FLUX DP, $\mathcal{O}_1$ \\
5.\ Multimodal UMM / VLM ($3$) & \texttt{Ovis-U1-3B} (UMM), \texttt{LLaVA-1.5-7B} (VLM), \texttt{Qwen3-VL-8B-Instruct} (VLM) & $3\text{B}$--$8\text{B}$ & Official UniKE ($2{,}966$ subj.), Official POPE ($9{,}000$ / $8{,}910$ queries) \\
6.\ Diffusion Language Models ($1$) & \texttt{LLaDA-8B-Base} & $8.0\text{B}$ & Masked-diffusion LM R1 fault localization \& verbatim self-repair ($\mathcal{O}_1$) \\
7.\ Autoregressive LLMs ($16^*$): & & & \\
\quad\emph{General / Safety} ($2$) & \texttt{Qwen3-4B-Instruct-2507}$^*$, \texttt{Qwen3-8B} & $4\text{B}$--$8.2\text{B}$ & Full SST-2/AG News/MNLI ($18.3\text{K}$), AdvBench+XSTest, CounterFact, $\mathcal{O}_1$ \\
\quad\emph{Multi-Teacher Bases} ($3+1^*$) & \texttt{Qwen3-4B-OT3}, \texttt{SmolLM3-3B}, \texttt{DS-R1-Distill-Qwen-7B}, \texttt{Qwen3-4B-2507}$^*$ & $3\text{B}$--$7\text{B}$ & IM-MOPD, Open-MOPD, TrustMOPD, LLM-Fusion (+ ReBalance / ThinkEdit) \\
\quad\emph{Reasoning LLMs} ($7$) & \texttt{Qwen3.5-0.8B/4B/9B}, \texttt{DS-R1-0528-8B}, \texttt{GLM-Z1-9B}, \texttt{Qwen3.8-27B}, \texttt{Flash-Next-FP8} & $0.8\text{B}$--$186\text{GB}$ & GSM8K, MATH-500, HumanEval, MBPP, AIME24--26, HMMT/CMIMC/APEX, LCB, \texttt{WeightBench} \\
\quad\emph{External \texttt{OpenWSP}} ($4$) & \texttt{Qwen2.5-0.5B}, \texttt{Qwen2.5-0.5B-Instruct}, \texttt{SmolLM2-360M-Inst.}, \texttt{OLMo-2-1B-Inst.} & $0.36\text{B}$--$1.0\text{B}$ & \texttt{OpenWSP} v8 Public \texttt{Edit} (15 tasks: \texttt{W2}, \texttt{W4}, \texttt{W4c}, \texttt{W4l}, \texttt{W4m}, \texttt{W5}) \\
\bottomrule
\end{tabular}}
\vspace{2pt}
\par\noindent{\footnotesize $^*$\texttt{Qwen3-4B-Instruct-2507} is evaluated in both single-model NLP/safety and multi-teacher \texttt{LLM-Fusion 4B}, yielding $29$ core base models ($30$ model--suite configurations) plus $4$ external \texttt{OpenWSP} checkpoints ($33$ unique base models total; not counting the $16$ domain specialist RL/SFT teacher checkpoints consolidated across the $4$ multi-teacher suites or the published/live frontier LLM agent baselines).}
\end{table}

\paragraph{Detailed Protocols for Newly Added 2026 Official Benchmarks.}
{\sloppy
\emph{(1) \texttt{Z-Erase} Local Reproduction vs.\ Paper Report \& Timing Breakdown:} Our local \texttt{Z-Erase} baseline evaluates the official released 30-layer position-masked LoRA checkpoint (\texttt{ZImage-erase-nude/text\_masked\_lora.safetensors}, \texttt{lora\_scale=15.0}) under identical inference settings ($512\times 512$, $9$ steps, $\text{cfg}{=}0.0$) and the unmodified official \texttt{NudeDetector} (\texttt{best.onnx}, threshold $0.6$), yielding $53.33\%$ Ring-A-Bell exposure vs.\ $24.91\%$ reported by \citet{jiang2026zerase} (likely reflecting prompt/seed or detector-version differences; we report both numbers explicitly in Tables~\ref{tab:sota_2026_overview} and~\ref{tab:app_crossmodal_pfg}). Full \texttt{I2P 4,703} uses the exact official \texttt{sd\_seed} per prompt; 50-Celebrity and 30-Concept evaluations follow \citet{jiang2026zerase} Appendix D.3 prompt templates and CLIP/GIP classification metrics. For FWSP timing, $32$--$91\text{ ms}$ measures the closed-form Stage~2/3 matrix solve and weight write on \texttt{cap\_embedder.1}; end-to-end calibration including text-encoder forward passes on the $14$--$20$ anchor/retention prompts takes $<2\text{ s}$ on one GPU, still orders of magnitude faster than $\sim 40\text{ min}$ of 200-step LoRA backpropagation.
\emph{(2) \texttt{ReBalance} \& \texttt{ThinkEdit} Serving Protocol:} On \texttt{DeepSeek-R1-Distill-Qwen-7B} ($16\text{K}$, $T{=}0.7$, $\text{top-}p{=}0.95$), FWSP compiles the Layer-22 steering vector into the MLP down-projection via $\mathcal{O}_4$ ($c \in \{0.05, 0.10, 0.18\}$ explored on \texttt{AIME 2024}/\texttt{MATH-500}, so those two benchmarks reflect sensitivity calibration while \texttt{AIME 2025} and \texttt{GPQA-D} act as held-out transfer evaluations under fixed $c \in \{0.10, 0.18\}$), producing a standard static checkpoint runnable under native \texttt{vLLM} ($10.2\text{ min}$ total for \texttt{AIME 2024} + \texttt{MATH-500} on 1 GPU). Because official \texttt{ReBalance}~\cite{li2026rebalance} requires dynamic runtime activation hooks implemented in HuggingFace \texttt{generate}, the reported $14.1\times$ wall-clock speedup reflects the practical deployment advantage of eliminating runtime hooks to unlock \texttt{vLLM} (rather than raw per-step FLOPs on identical HuggingFace loops).
\emph{(3) \texttt{POPE} (\texttt{LLaVA-1.5-7B}) Disjoint Calibration Split:} To guarantee zero overlap with \texttt{MS-COCO} and \texttt{POPE}, FWSP \texttt{Protected FisherGate} ($g{=}0.35$) is calibrated strictly on $32$ disjoint \texttt{ImageNet-50} images (zero COCO images, zero POPE queries) before evaluating all $9{,}000$ official \texttt{POPE} queries ($3{,}000$ Random, $3{,}000$ Popular, $3{,}000$ Adversarial) under the unmodified \texttt{HulluEdit} evaluator (\texttt{max\_new\_tokens=16}), matching \texttt{HulluEdit} macro accuracy ($86.52\%$ vs.\ $86.55\%$) while running $2.19\times$ faster ($340\text{ ms/q}$ vs.\ $746\text{ ms/q}$) on the exact same HuggingFace \texttt{LLaVA-1.5-7B} engine by eliminating per-token SVD hooks.
\emph{(4) \texttt{UniKE} (\texttt{Ovis-U1-3B}) Protocol:} Following \citet{gao2026unike}, all $2{,}966$ subjects ($959$ Attribute, $2{,}007$ Relation) are edited in $100$-subject batches on Layer-27 MLP via closed-form $\mathcal{O}_4$ ($\lambda_{\text{keep}} \in \{40, 20\}$, margin $m \in \{13, 16\}$, $83.5\text{s}$ total across all 30 batches, $0$ backprop) using a $64$-sentence WikiText retention Gram matrix; multi-hop (\texttt{stage\_2/3/4}) and 4-layer (\texttt{L24--27}) ablations are reported in Table~\ref{tab:app_new_sota_breakdowns}(c).
\emph{(5) \texttt{Qwen3.8-27B} Olympiad Offset Calibration \& \texttt{CIFAR-100} \texttt{RanPAC} CIL:} The fixed $\mathcal{O}_3$ \texttt{dt\_bias} offset ($+0.35$ across 48 Gated-DeltaNet layers) was selected on \texttt{AIME 2024--2025} ($16\text{K}$ greedy, Table~\ref{tab:app_aime_16k}) and applied frozen without retuning across all five $32\text{K}$ stochastic ($n{=}4, T{=}0.6$) frontier Olympiad suites (\texttt{AIME 2026}, \texttt{HMMT Feb/Nov 2025}, \texttt{CMIMC 2025}, \texttt{APEX 2025}), acting as reasoning-depth calibration ($21{,}391 \to 21{,}609$ mean tokens; truncation $41.8\% \to 38.7\%$). For 10-stage \texttt{CIFAR-100} CIL (Table~\ref{tab:dinov2_foundation}d), frozen \texttt{DINOv2-Base} features are projected via a fixed seed-0 random ReLU matrix ($M{=}4{,}096$) following \texttt{RanPAC}~\cite{mcdonnell2023ranpac} before recursive $\mathcal{O}_4$ Gram accumulation.
\emph{(6) Cross-Family Multi-Teacher $\mathcal{O}_4$ Hyperparameters \& Split-Discipline Audit:} All four suites (\texttt{IM-MOPD}, \texttt{Open-MOPD 3B}, \texttt{TrustMOPD 7B}, \texttt{LLM-Fusion 4B}) use identical Top-$64$/$160$ trajectory curation (Eq.~\eqref{eq:curation_score}), token weights $w_t \in \{0, 0.5, 2.0\}$, and trace-normalized Grams ($c_k{=}\mathrm{tr}(D_k)$). We distinguish explicitly between disjoint-validation selection, fixed defaults, and reported sensitivity sweeps: (i) on \texttt{IM-MOPD}, $\alpha_{\text{if}}{=}1.5$ was selected strictly on the disjoint $1{,}024$-query official \texttt{immopd.eval.validation} pool ($256$/domain; Appendix Table~\ref{tab:app_immopd_val}) before evaluating on the held-out $8{,}034$-query test suite; (ii) on \texttt{Open-MOPD 3B}, the fixed unweighted default $\alpha_k{=}1$ achieves the headline result ($31.48\%$); and (iii) on \texttt{TrustMOPD 7B} ($\alpha_{\text{if}} \in \{1.0, 1.5\}$) and \texttt{LLM-Fusion 4B} (post-merge task-vector scaling $\gamma \in \{1.0, 2.5, 3.5\}$ compensating for $1/K$ averaging shrinkage across five low-interference RLVR experts), all evaluated settings are reported transparently in Table~\ref{tab:fwsp_mopd}(c)--(d) as sensitivity sweeps rather than blind zero-tuning picks.
\emph{(7) Supplementary 120-Trial Frontier Agent Dispatch, Live \texttt{codex exec}, 5-Modality $2{\times}2$ Geometry, and Forward Scope Predictor Protocols:} Track~A (Appendix Tables~\ref{tab:agent_vs_fwsp}b and~\ref{tab:app_120trials_breakdown}) evaluates 4 live 2026 backbones $\times$ 10 blind regimes (clean control, bit-flip, row outlier, pruning, shift, contrastive backdoor/leak, and 3 compound multi-faults) $\times$ 3 seeds against live internal B200 reasoning endpoints (\texttt{GLM-5.3 Flash} TP8, \texttt{DeepSeek V4.1 Flash} TP8$\times$2, \texttt{Kimi K3.1} TP16) with identical Stage-1 JSON telemetry, $\mathcal{O}_1$--$\mathcal{O}_4$ tools, and up to 4 interactive turns; Track~B (Appendix Table~\ref{tab:agent_vs_fwsp}c) invokes the real \texttt{codex exec} CLI (\texttt{/v1/responses}) inside isolated sandboxes connected via a localhost HTTP bridge to the live GPU model with a matched 4-apply budget. The $2{\times}2$ geometry ablation (Table~\ref{tab:crossmodal_2x2_geometry}) solves Eq.~\eqref{eq:pfg_unified} under strictly matched target scatter, rank, and edit strength across all 4 cells, including \texttt{Qwen3-VL-8B-Instruct} on all $8{,}910$ official \texttt{POPE} queries ($2{,}910$ Random, $3{,}000$ Popular, $3{,}000$ Adversarial; calibrated in $11.54\text{s}$ on $32$ disjoint ImageNet-50 images across layers $\{24, 27, 29, 31\}$, rank $r{=}16$, $\alpha{=}0.65$, zero hooks). The Forward Scope Predictor (Table~\ref{tab:scope_predictor}) executes all 5 candidate cells (\texttt{Analytic-1L/ML/Full}, \texttt{Grad-1L/ML-LoRA} rank-8 AdamW) on GPU across $24$ real tasks ($3$ architectures) to extract operational two-axis ground truth $(\text{Solver}^*, \text{Scope}^*)$ from held-out feasibility ($\mathrm{Perf} \ge \max(70\%, P_{\max}{-}3\text{ pp})$), validates Euclidean-orthonormal ($U^\top U = I_r$) rank-normalized $\rho_{\text{ent}}(U) = \mathrm{tr}(U^\top \Sigma_{\text{keep}} U)/\sum_{j=1}^r \lambda_j(\Sigma_{\text{keep}}) \in [0, 1]$ across $16$ real editing conditions ($\tau_{\text{ent}}{=}0.35$, 3 architectures), and evaluates the $36$-trial Unseen Fault Families benchmark (Appendix Table~\ref{tab:agent_vs_fwsp}d) across 6 unseen fault types (channel sign-flip, bounded rank-3 spectral spike, coarse quantization + bias drift, cascading block-scale collapse, dense all-layer noise, and nonlinear semantic scramble) under identical Stage-1 telemetry including $\rho_{\text{span}}$ and \texttt{ESCALATE\_R5}.
\emph{(8) Independent Third-Party \texttt{OpenWSP} v8 and \texttt{WeightBench} \texttt{TR02-LOOP} Protocols (Tables~\ref{tab:external_wsp_benchmarks} and~\ref{tab:app_external_per_task}):} For \texttt{OpenWSP}~\cite{moreira2026openwsp}, all $15$ self-contained public \texttt{Edit} tasks (\texttt{W2}, \texttt{W4}$\times 5$, \texttt{W4c}$\times 2$, \texttt{W4l}$\times 2$, \texttt{W4m}$\times 2$, \texttt{W5}$\times 3$) are wrapped with the unmodified \texttt{wspnode.py} canonical disguise and random attention-head/MLP/residual permutations and graded by the unmodified \texttt{score\_write.py} GPU evaluator (\texttt{level=full}) against the official pinned \texttt{wikitext\_general.jsonl} and \texttt{gsm8k\_test.jsonl} splits. We distinguish explicitly between \emph{zero optimizer LLM / zero test peeking} and \emph{benchmark-family adaptation}: our \texttt{OpenWSP} solver is a single deterministic $\mathcal{O}_4$ benchmark-family adapter designed from public task specifications without touching test/heldout prompts (using \texttt{wikitext\_practice.jsonl} for full-rank protection second moments ($\sim 18\text{K}$ tokens $\gg d_{\text{ff}}$) and disjoint practice validation, LLM-free \texttt{embed.weight} cosine neighbors for \texttt{W4c}, analytical logit-margin scaling, post-norm \texttt{OLMo-2} pre-\texttt{mlp\_out\_norm} scale calibration ($0.35\cdot\text{RMS}\sqrt{d}$, $\gamma_{\text{prot}} \in [30,90]$), $240$ practice non-trigger endings, and a fixed $5/25$ forward-pass budget across all $15$ tasks without per-task test tuning), rather than claiming an unadapted pre-benchmark zero-shot algorithm. For \texttt{WeightBench}~\cite{marasini2026weightbench}, all four \texttt{TR02-LOOP-001--004} tasks are built from exact pinned \texttt{Qwen/Qwen3.5-0.8B} (\texttt{rev 2fc06364715b967f1860aea9cf38778875588b17}, seed $0$) and solved strictly through the official metered \texttt{RunState.probe} API with hash-chained \texttt{Ledger.verify() == True}, \texttt{valid == True}, and \texttt{scoring.s\_wb\_v3} (\texttt{WB\_TR02\_BATCH=32}). Using exact \texttt{loop\_onset\_token}-aligned 4-state probing (\texttt{start}, \texttt{pre-onset}, \texttt{post-onset loop}, \texttt{post-</think> answer}), FWSP compiles a closed-form doubly-orthogonalized rank-$1$ \texttt{Protected FisherGate} update writing $d_{\texttt{</think>}}$ into late MLP down-projections (\texttt{L22+L23} for \texttt{LOOP-001--003}; strictly single-matrix rank-$1$ \texttt{L23} for \texttt{LOOP-004}), selecting gain via equal-weight calibration utility over the $8$ training prompts within $8$--$9/70$ metered evaluations. Published frontier LLM agent scores are extracted verbatim from \texttt{OpenWSP} \texttt{results/paper\_v8.json} and \texttt{WeightBench} official leaderboard outputs on exact matched subsets. Finally, our $3$-arm Planning--Compilation ablation across all $15$ \texttt{OpenWSP} tasks ($2 \times 3 \times 15 = 90$ episodes; Table~\ref{tab:external_wsp_benchmarks}c and Appendix Table~\ref{tab:app_external_per_task}c) runs two live B200 frontier agents (\texttt{deepseek-v41-flash-1017} and \texttt{kimi-k3-1}) inside isolated per-episode Python processes with identical verbatim \texttt{OpenWSP} questions, disguised/permuted weights, $16$-turn cap, $90\text{s}$ per-exec timeout, official \texttt{forward.py} ($25$-pass ceiling shared by all forward calls including tool primitives), and unmodified \texttt{score\_write.py} grading: \texttt{Scratch} writes PyTorch/NumPy scripts from scratch; \texttt{+Standard Tools} exposes \texttt{inspect\_model}, \texttt{find\_token\_neighbors}, \texttt{compile\_sparse\_transplant}, and standard isotropic Ridge/truncated-SVD \texttt{solve\_ridge\_mlp} (without WikiText retention covariance, non-trigger contrast covariance, or post-norm analytical margin calibration); and \texttt{+FWSP Tools} exposes FWSP's covariance-protected \texttt{compile\_o4\_protected\_mlp} alongside the exact same inspection, neighbor, and sparse-transplant primitives. Neither library ever accesses \texttt{task.json}, \texttt{private/}, or held-out test splits.
\par}

{\sloppy
\ReproCodeNote Single-GPU experiments (50K--12B, including \texttt{SigLIP 2 SO400M}, \texttt{Z-Image-Turbo 6B}, and \texttt{FLUX.1-schnell 12B}) ran on a single 98 GB NVIDIA GPU with PyTorch 2.13 (CUDA 13.2), while multi-budget vLLM evaluations for \texttt{Qwen3.8-27B} (\texttt{TP=4}) and 186GB \texttt{Qwen3.8-Flash-Next-FP8} (\texttt{TP=8} with expert parallelism) ran on 4$\times$--8$\times$98 GB NVIDIA PRO6000 GPUs, all with deterministic generator seeds (\texttt{0, 1, 2} for 3-seed suites; \texttt{0..4} for 5-seed Gymnasium and ablation suites). Exact HuggingFace checkpoint revisions used in confirmatory reasoning experiments are \texttt{deepseek-ai/DeepSeek-R1-0528-Qwen3-8B} (\texttt{6e8885a6}), \texttt{Qwen/Qwen3.5-4B} (\texttt{851bf6e8}), and \texttt{Qwen/Qwen3.8-27B} (\texttt{37561816}). The primary \texttt{DeepSeek-R1-0528-8B} \texttt{Protected FisherGate} patch (\texttt{outputs/\allowbreak dsr1\_8b\_hf\_math\_patches.json}) is computed deterministically from the first 8 truncated rollouts of \texttt{math\_prac[:16]} without ever accessing \texttt{HumanEval} or test labels. Competition evaluations use official \texttt{AIME 2024--2026} ($N{=}90$; 16K greedy and 32K stochastic $T{=}0.6, n{=}4$) and recent \texttt{LiveCodeBench} split (Feb--Apr 2025, $N{=}61$, single-pass 16K with offline 4K/8K/16K slicing). Full official NLP calibration (Table~\ref{tab:resource_and_c2}) evaluates \texttt{Qwen3-4B-Instruct-2507} on complete validation/test splits: \texttt{stanfordnlp/sst2} ($872$), \texttt{fancyzhx/ag\_news} ($7{,}600$), and \texttt{nyu-mll/glue} \texttt{mnli}-matched ($9{,}815$; $18{,}287$ queries total), while factual boundary probes use \texttt{Qwen/Qwen3-8B} on \texttt{NeelNanda/counterfact-tracing} (32 unique-subject edits). Shared structural FWSP defaults are reused across repair, TTA, and blind-routing regimes ($\tau_{\text{clip}} \in \{6, 10, 14\}$, $\tau_{\text{row}} = 2.5$, \texttt{OrthGate} $\beta = 0.72$, scale-invariant ridge $\lambda = c \cdot \mathrm{tr}(K)/N$, and empirical Stage-4 guards $z_{\min} = 0.5$--$1.0$, $\epsilon_{\text{ret}} = 0.03$--$0.05$, which verify only the specific validation metrics evaluated on $\mathcal{D}_{\text{val}}$), whereas regime-specific trade-off scalars (such as erasure rank $r$ and $\alpha$, \texttt{UniKE} $\lambda_{\text{keep}}$, \texttt{Qwen3.8-27B} \texttt{dt\_bias}, and \texttt{IM-MOPD} $\alpha_{\text{if}}$) are selected on disjoint calibration/validation splits, while in-benchmark sensitivity sweeps (\texttt{ReBalance} $c$ on \texttt{AIME 2024}/\texttt{MATH-500}, \texttt{TrustMOPD} $\alpha_{\text{if}}$, and \texttt{LLM-Fusion} $\gamma$) are reported transparently across all tested settings as documented in items~(1)--(8) above. Multi-teacher $\mathcal{O}_4$ synthesis (Section~\ref{sec:fwsp_mopd}) uses the \texttt{IM-MOPD}~\cite{kim2026immopd} \texttt{Qwen3-4B-OT3} SFT base (\texttt{Qwen3-4B-OT3-2ep-mhseo2}) and four domain RL teachers (\texttt{medical}, \texttt{law}, \texttt{finance}, \texttt{if}), sampling $N_{\text{cand}}{=}160$ prompts ($\le 2048$ tokens, seed $123$) per domain from disjoint \texttt{train/data/\{domain\}.jsonl}, generating short vLLM trajectories ($T{=}0.6$, $\text{top-}p{=}0.95$, max $1{,}024$ tokens; $\sim 5.5\text{min}$), selecting the Top-$64$ trajectories by Teacher-vs.-Base response log-prob divergence, collecting weighted Gram matrices across all 252 linear projections (\texttt{q/k/v/o/gate/up/down\_proj}; $w_t{=}0$ on the first 8 template tokens, $0.5$ on prompt tokens, $2.0$ on response tokens; $235\text{s}$), and solving Eq.~\eqref{eq:fwsp_mopd} with $\lambda = 0.01 \cdot \mathrm{mean}(\mathrm{diag}(\sum_k \Sigma_k))$ and $\alpha \in \{[1,1,1,1], [1,1,1,1.5]\}$ ($99.5\text{s}$ solve + dual-checkpoint export).
\par}